\documentclass[3p, times]{elsarticle}
\usepackage{amssymb}
\usepackage{lineno}
\usepackage{amssymb}
\usepackage{babel}
\usepackage{paralist, tabularx}
\usepackage{lipsum}
\usepackage{amsmath,amssymb,amsfonts}
\usepackage{algorithmic}
\usepackage{graphicx}
\usepackage{textcomp}
\usepackage{xcolor}
\usepackage{graphicx}

\usepackage{colortbl}
\usepackage{pifont}
\usepackage{multirow}
\usepackage{enumitem}
\usepackage{amsmath, amsfonts}
\usepackage{listings}
\usepackage{xcolor}
\usepackage{framed}
\usepackage{fancyhdr}
\usepackage{url}
\usepackage{subfigure}
\usepackage{xspace}
\usepackage{booktabs}

\begin{document}

\begin{frontmatter}

%% Title, authors and addresses

%% use the tnoteref command within \title for footnotes;
%% use the tnotetext command for the associated footnote;
%% use the fnref command within \author or \address for footnotes;
%% use the fntext command for the associated footnote;
%% use the corref command within \author for corresponding author footnotes;
%% use the cortext command for the associated footnote;
%% use the ead command for the email address,
%% and the form \ead[url] for the home page:
%%
%% \title{Title\tnoteref{label1}}
%% \tnotetext[label1]{}
%% \author{Name\corref{cor1}\fnref{label2}}
%% \ead{email address}
%% \ead[url]{home page}
%% \fntext[label2]{}
%% \cortext[cor1]{}
%% \address{Address\fnref{label3}}
%% \fntext[label3]{}

%\dochead{}
%% Use \dochead if there is an article header, e.g. \dochead{Short communication}
%% \dochead can also be used to include a conference title, if directed by the editors
%% e.g. \dochead{17th International Conference on Dynamical Processes in Excited States of Solids}

%\title{Enhancing Mixup-Based Graph Learning for Language Processing via Hybrid Pooling}
\title{On the Effectiveness of Hybrid Pooling in \\Mixup-Based Graph Learning for Language Processing}
%\title{An Empirical Study On the Effectiveness of Hybrid Pooling in Mixup-Based Graph Learning}

%% use optional labels to link authors explicitly to addresses:
%% \author[label1,label2]{<author name>}
%% \address[label1]{<address>}
%% \address[label2]{<address>}

\author[1]{Zeming Dong}
\author[2]{Qiang Hu\corref{correspondingauthor}}

\cortext[correspondingauthor]{Corresponding author}

\author[1]{Zhenya Zhang}
\author[3]{Yuejun Guo}
\author[2]{Maxime Cordy}
\author[2]{Mike Papadakis}
\author[2]{Yves Le Traon}
\author[1]{Jianjun Zhao}

\address[1]{Kyushu University}
\address[2]{University of Luxembourg}
\address[3]{Luxembourg Institute of Science and Technology}

 \begin{abstract}

\emph{Graph neural network (GNN)}-based graph learning has been popular in natural language and programming language processing, particularly in text and source code classification. Typically, GNNs are constructed by incorporating alternating layers which learn transformations of graph node features, along with graph pooling layers that use graph pooling operators (e.g., Max-pooling) to effectively reduce the number of nodes while preserving the semantic information of the graph. %In particular, the graph pooling operators~(e.g., Max-pooling) in the graph pooling layer, which process node representations into the entire graph representations, play a vital role in condensing the graph representation, ensuring computational efficiency while preserving crucial graph information. 
Recently, to enhance GNNs in graph learning tasks, \emph{Manifold-Mixup}, a data augmentation technique that produces synthetic graph data by linearly mixing a pair of graph data and their labels, has been widely adopted. However, the performance of \emph{Manifold-Mixup} can be highly affected by graph pooling operators, and there have not been many studies that are dedicated to uncovering such affection. 
%the potential affection of graph pooling operators to the performance of graph learning with \emph{Manifold-Mixup} remains an open question.
%graph pooling operators, which could potentially affect the performance of \emph{Manifold-Mixup}, have not been well studied. 
To bridge this gap, we take an early step to explore how graph pooling operators affect the performance of Mixup-based graph learning. To that end, we conduct a comprehensive empirical study by applying \emph{Manifold-Mixup} to a formal characterization of graph pooling based on 11 graph pooling operations (9 hybrid pooling operators, 2 non-hybrid pooling operators). The experimental results on both natural language datasets (Gossipcop, Politifact) and programming language datasets (JAVA250, Python800) demonstrate that hybrid pooling operators are more effective for \emph{Manifold-Mixup} than the standard Max-pooling and the state-of-the-art graph multiset transformer (GMT) pooling, in terms of producing more accurate and robust GNN models.

\end{abstract}

\begin{keyword} Hybrid Pooling \sep Data Augmentation \sep Graph Learning \sep Manifold-Mixup \sep Language Processing

%% keywords here, in the form: keyword \sep keyword

%% PACS codes here, in the form: \PACS code \sep code

%% MSC codes here, in the form: \MSC code \sep code
%% or \MSC[2008] code \sep code (2000 is the default)

\end{keyword}
\end{frontmatter}
%%
%% Start line numbering here if you want
%%
%\linenumbers

%% main text

%The alternating layers serve as a mechanism for information exchange among neighboring nodes, fostering a holistic understanding of the graph structure. In parallel, the pooling layers play a vital role in condensing the graph representation, ensuring computational efficiency while preserving crucial information.

\section{Introduction}
\label{sec:intro}
Since texts, as well as source code, can be represented as graph-structured data~\cite{huang2019, allamanis2018survey}, \emph{graph neural network (GNN)}-based \emph{graph learning} has been increasingly applied for both \emph{natural language processing (NLP)} \cite{wu2020comprehensive}, and \emph{programming language (PL)} understanding~\cite{dinella2020hoppity, wang2020detecting}. The application of GNNs has achieved remarkable results, e.g., Allamanis et al.~\cite{allamanis2018learning} utilize GNNs to learn the syntax tree and data flow representations of source code, by which they manage to accomplish several software engineering tasks, such as code completion and defect detection. %This hot trend makes GNNs increasingly important in the PL understanding field.

Typically, high-quality training data, including features and their corresponding labels, are necessary to train GNN models with competitive performance. However, preparing the labeled data is often not easy, especially in the context of source code labeling that requires advanced expertise~\cite{zhou2019devign}. %due to the huge need for human resources with domain knowledge. 
%In the context of source code labeling, this becomes even more difficult and costly, since it requires sophisticated expertise in code understanding. Indeed, 
%Especially for the PL datasets, understanding code is the premise of labeling code. For example, 
%labeling only four libraries of code can take up to 600 man-hours~\cite{zhou2019devign}. 
To alleviate the data labeling issue, \emph{data augmentation} has been proposed to enhance training data by modifying original data.
%during the training process. %As an advanced data augmentation technique,  
As the state-of-the-art in data augmentation, 
\emph{Mixup}~\cite{zhang2018mixup} achieves impressive results in different tasks. Take image classification for an example: \emph{Mixup}
 synthesizes new images and labels as additional training data by first selecting two raw images at random from the original training data and then linearly mixing their features and labels. Recent research~\cite{wang2021mixup, dong2023mixcode} demonstrates that \emph{Manifold-Mixup}, the specialized application of \emph{Mixup} on graph-structured data~\cite{dong2023mixcode}, focusing on interpolating graph-level embeddings, can also achieve great performance for graph-structured data. With the success of \emph{Manifold-Mixup}, utilizing Mixup-based data augmentation in graph learning has emerged as a mainstream paradigm.

%Mixup \cite{zhang2018mixup} is one simple yet efficient data augmentation method that linearly mixes two data and their labels to create new cases.

%Although \emph{Mixup} is initially proposed for image data, recent research~\cite{wang2021mixup, dong2023mixcode} demonstrates that it can also achieve good performance for graph-structured data. 
%With the success of \emph{Mixup}, utilizing Mixup-based data augmentation has emerged as a mainstream paradigm to enhance graph learning.

As indicated by existing studies~\cite{dong2023mixcode,verma2019manifold}, Mixup-based graph learning is mainly influenced by two factors, namely, 1) the hyperparameters in \emph{Mixup} itself, such as the \emph{Mixup ratio} that balances the proportion of the source data, and 2) the \emph{Mixup} strategies that are associated with representation generation. Over these factors, the hyperparameter issue is a common one across several different Mixup-applied fields, and has been extensively studied in fields such as image classification~\cite{zhang2021and,zhang2020does}. However, the second issue about Mixup strategies is highly correlated to the context in which Mixup is applied, and for graph-structured data, the influence of such an issue has not been well studied.

%For image data, 
%However, different from other data types (e.g., image) where 
%the factors of \emph{Mixup} that could affect the performance of the trained model have been extensively studied \cite{zhang2021and,zhang2020does}, and accordingly, multiple variants of \emph{Mixup} have been proposed \cite{verma2019manifold,summers2019improved,guo2019mixup,takahashi2018ricap,yun2019cutmix,kim2020puzzle,wu2020dual,mao2019virtual}. 
%However, for graph-structured data, especially for code data, such analysis about \emph{Manifold-Mixup} remains at an early stage. Existing studies~\cite{dong2023mixcode,verma2019manifold} have indicated that Mixup-based graph learning can be influenced by two factors: 1) the \emph{Mixup} ratio, and 2) the \emph{Mixup} strategies that are associated with representation generation. 

In the context of Mixup-based graph learning, 
%Given a Mixup-based graph learning architecture, 
as shown in Figure~\ref{fig:overview},
\emph{Manifold-Mixup} is fed with the inputs from the \emph{graph pooling layer},  
which use graph pooling operators (e.g., Max-pooling) to produce coarsened representations of the given graph while preserving its semantic information.
%are used to produce coarsened representations of the given graph. The output value of the graph pooling layer serves as the input value for the \emph{Manifold-Mixup} operation. 
Namely, this layer is the key to representation generation of \emph{Manifold-Mixup}, in that \emph{Manifold-Mixup} generates augmented training data by interpolating these representations. Therefore, the performance of \emph{Manifold-Mixup} can be highly affected by the graph pooling operators. 
Recent works~\cite{Knyazev2019, Mesquita2020,9836996} have attempted to systematically analyze the importance of graph pooling in representation generation; however, the following question, namely, \textit{how different graph pooling operators affect the effectiveness of Mixup-based graph learning},  still remains open.

%Generally, a GNN model is composed of alternating layers and graph pooling layers. The alternating layers are responsible for learning the transformation of each node feature, and the graph pooling layers are used to produce coarsened representations of the given graph. As shown in Figure~\ref{fig:overview}, the output value of the pooling layer serves as the input value for the \emph{Mixup} operation. It is a remarkable fact that \emph{Mixup} then leverages interpolating these values as augmented training data. There might be various potential factors that could affect the training of GNNs when using \emph{Mixup}.  Recent works~\cite{Knyazev2019, Mesquita2020,9836996} have attempted to systematically analyze the importance of graph pooling in representation, and, notably, the effect of graph pooling in data augmentation is still missing. This raises the question: \textit{how do different graph pooling operators affect the effectiveness of Mixup-based graph learning?} 

In this paper, we tackle this problem by empirically analyzing the difference when \emph{Mixup} is applied in different graph representations generated by different graph pooling operators. Specifically, we focus on two types of graph pooling methods, namely, standard pooling methods and a unifying formulation of hybrid (mixture) pooling operators. For the standard pooling, the Max-pooling, which is the most widely used one~\cite{Mesquita2020}, and the state-of-the-art \emph{graph multiset transformer pooling} (GMT)~\cite{baek2021accurate} which is a global pooling layer based on multi-head attention and capturing node interactions based on structural dependencies, are considered. For the hybrid pooling, we extend the prior work \cite{9836996, nguyen2022regvd} and design 9 types of hybrid pooling strategies, and more details are introduced in Table~\ref{hybrid_pooling_layer}.  Here, GMT and hybrid pooling operators are considered more advanced strategies. We conduct empirical experiments to evaluate the effectiveness of graph learning using \emph{Manifold-Mixup}~\cite{verma2019manifold}, under different hybrid pooling operators. In total, our experiments cover diverse types of datasets, including two programming languages (JAVA and Python) and one natural language (English), and consider different tasks, two widely-studied graph-level classification tasks (program classification and Fake news detection), and six GNN model architectures. Based on that, we answer the following research questions:

\smallskip
\noindent\textbf{RQ1: How effective are hybrid pooling operators for enhancing the accuracy of Mixup-based graph learning?}
The results on NLP datasets (Gossipcop and Politifact used for fake news detection) show that the hybrid pooling operator Type 1~($\mathcal{M}_{sum}(\mathcal{P}_{att},\mathcal{P}_{max})$) outperforms GMT by up to 4.38\% accuracy. On PL datasets (JAVA250 and Python800 used for problem classification), also the hybrid pooling operator Type 1~($\mathcal{M}_{sum}(\mathcal{P}_{att},\mathcal{P}_{max})$) surpasses GMT by up to 2.36\% accuracy.

\smallskip
\noindent\textbf{RQ2: How effective are hybrid pooling operators for enhancing the robustness of Mixup-based graph learning?}
The results demonstrate that in terms of robustness, the hybrid pooling operator Type 6~($\mathcal{M}_{concat}(\mathcal{P}_{att},\mathcal{P}_{sum})$) surpasses GMT by up to 23.23\% in fake news detection, while the hybrid pooling operator Type 1~($\mathcal{M}_{sum}(\mathcal{P}_{att},\mathcal{P}_{max})$) outperforms GMT by up to 10.23\% in program classification.

\smallskip
\noindent\textbf{RQ3: How does the hyperparameter setting affect the effectiveness of Manifold-Mixup when hybrid pooling operators are applied?} According to~\cite{zhang2018mixup}, the hyperparameter $\lambda$ denotes the interpolation ratio, and it is sampled from a \emph{Beta} distribution with a shape parameter $\alpha\left(\lambda\sim\textit{Beta}\left(\alpha,\alpha\right)\right)$. Existing works~\cite{dong2023mixcode,verma2019manifold} show that the hyperparameter setting $\lambda$ affects the performance of Mixup. Therefore, we study the effectiveness of \emph{Manifold-Mixup} when using hybrid pooling operators under different hyperparameters of \emph{Mixup}. Experimental results indicate that a smaller value of the hyperparameter leads to better robustness and accuracy.

In summary, the contributions of this paper are as follows:
\begin{itemize}
\item This is the first work that explores the potential influence of graph pooling operators on Mixup-based graph-structured data augmentation. To facilitate reproducibility, our code and data are available online~\footnote{\url{https://github.com/zemingd/HybridPool4Mixup}\label{site}}.

\item We discuss and further extend the hybrid pooling operators from existing works.

\item The comprehensive empirical analysis demonstrates that hybrid pooling is a better way for Mixup-based graph-structured data augmentation. 
\end{itemize}

\section{Background and Related Work}

\subsection{Graph Data Classification }
Researchers have proposed multiple approaches for the text classification task that analyze the data based on its graph structure. In which Yao \emph{et al.}~\cite{yao2019graph} constructed text graph data by using the words and documents as nodes. To further enhance text classification performance, Zhang \emph{et al.}~\cite{zhang2020every} proposed the graph-based word interaction to capture the contextual word relationships. Similar to text data, by capturing the relationship of different components (e.g., variables and operators) in the code, source code data can also be represented structurally as the graph data~\cite{wang2020detecting, dong2023boosting}. Zhou \emph{et al.}~\cite{zhou2019devign} mainly integrated four separate subgraph representations of source code into one joint graph data. Furthermore, to advance the generalization, Allamanis \emph{et al.}~\cite{allamanis2021self} offered four code rewrite rules, such as variable renaming, comment deletion, etc., as a data augmentation for graph-level program classification. 

Different from the above works, our study focuses on enhancing the performance of graph classification with Mixup-based data augmentation.

\subsection{Graph Pooling Operators}
\label{sec:bgpoolingop}
Graph pooling~\cite{ying2018hierarchical,9836996} plays a crucial role in capturing relevant structure information of the entire graph. Existing works~\cite{atwood2016diffusion,simonovsky2017dynamic,xu2018powerful} have proposed the basic graph pooling methods, such as summing or averaging all of the node features. However, such pooling methods treat every node information identically, which could lose the structural information. To solve this problem, researchers~\cite{zhang2018end,gao2019graph,lee2019self,cangea2018towards} dropped nodes with lower scores using a learnable scoring function, which can compress the graph and alleviate the impact of irrelevant nodes to save the important structural information. Additionally, to locate the tightly related communities on a graph, recent works~\cite{ma2019graph,wang2019haarpooling,bianchi2020spectral,yuan2020structpool} have considered the graph pooling as the node clustering problem, where nodes were specifically aggregated to the same cluster. To combine these advantages, Ranjan \emph{et al.}~\cite{ranjan2020asap} first clustered nearby nodes locally and dropped clusters with lower scores. In addition, there existed a kind of attention-based pooling methods~\cite{li2015gated,li2019semi,bahdanau2014neural,baek2021accurate} that scored nodes with an attention mechanism to weight the relevance of nodes to the current graph-level task. Besides, different from the above standard pooling methods, Nguyen \emph{et al.}~\cite{nguyen2022regvd} leveraged a mixture of Sum-pooling and Max-pooling methods for graph classification. 

In our study, we examine the application of \emph{Manifold-Mixup} in graph-level classification, incorporating both hybrid pooling and standard pooling techniques. Moreover, different from existing research~\cite{nguyen2022regvd}, we specifically include attention pooling because it has been proven effective for GNN model training~\cite{lee2019self, li2015gated}. We extended from the original three different types of hybrid pooling operators~\cite{nguyen2022regvd} to the current nine.

\subsection{Mixup}
Due to its effectiveness in graph-structured data processing and the promising performance on graph-specific downstream tasks, e.g., graph classification, GNN has recently received considerable attention. Meanwhile, as a sophisticated data augmentation method, \emph{Mixup}~\cite{zhang2018mixup} was initially proposed and implemented within the domain of image classification~\cite{shorten2019survey}. Specifically, Mixup randomly selects a pair of images from the training data, linearly combines their features and labels, and synthesizes a new image and label, which are treated as augmented training data. By combining two different data points linearly, Mixup effectively smooths the data distribution in the feature space. This smoothing process helps mitigate the sharpness of decision boundaries between different classes, reducing susceptibility to overfitting~\cite{dong2024effectiveness}. Due to its remarkable efficacy in classification tasks, it has subsequently been gradually applied to NLP~\cite{guo2019augmenting} and source code learning~\cite{dong2023mixcode}. 

In the NLP, Guo \emph{et al.}~\cite{guo2019augmenting} proposed two basic strategies of Mixup for augmenting data. One was wording embedding-based, and another was sentence embedding-based. After these two different kinds of embedding, the feature of input data can be mixed to synthesize the new data in vector space. To solve the difficulty in mixing text data in the raw format, Sun \emph{et al.}~\cite{sun2020mixup} mixed text data from transformer-based pre-trained architecture. Chen \emph{et al.}~\cite{chen2020mixtext} increased the size of augmented samples by interpolating text data in hidden space. Zhang \emph{et al.}~\cite{zhang2020seqmix} generated extra labeled sequences in each iteration to augment the scale of training data. Unlike previous work, some researchers considered the raw text itself to augment the input data. Yoon \emph{et al.}~\cite{yoon2021ssmix} synthesized the new text data from two raw input data by span-based mixing to replace the hidden vectors. In source code learning, Dong \emph{et al.}~\cite{dong2023mixcode} proposed a mixup-based data augmentation method that linearly interpolates the features of a pair of programs as well as their labels for the GNN model training.

In our work, we do not simply employ \emph{Mixup} for graph-structured data classification. Instead, we explore how different graph pooling operators affect the effectiveness of \emph{Mixup}.

\subsection{Empirical Study on Data Augmentation}
Recently, there has been a surge of empirical studies exploring the topic of data augmentation. In the NLP, Konno \emph{et al.}~\cite{konno-etal-2020-empirical} presented an empirical analysis to evaluate the effectiveness of contextual data augmentation (CDA) in improving the quality of the augmented training data, in comparison to \emph{[MASK]-based augmentation} and \emph{linguistically-controlled masking}. To assist practitioners in selecting suitable augmentation strategies, Chen \emph{et al.}~\cite{chen2023empirical} conducted a comprehensive empirical study on 11 datasets, encompassing topics such as news classification, inference tasks, paraphrasing tasks, and single-sentence tasks. In source code learning, Yu \emph{et al.}~\cite{YU2022111304} conducted an empirical study on three program-related downstream tasks, namely method naming, code commenting, and clone detection. The study aimed to validate the effectiveness of data augmentation that is designed by 18 program transformation methods that preserve both semantics and syntax-naturalness. Dong \emph{et al.}~\cite{dong2023boosting} presented a meticulous empirical study that aimed to evaluate the effectiveness of data augmentation methods that were adapted from the domains of NLP and graph learning for source code learning. The study rigorously examined the impact of these augmentation techniques on enhancing the performance of various tasks related to source code analysis and understanding.

Different from existing empirical studies, our work focuses primarily on examining the influence of graph pooling operators on Mixup-based data augmentation that has received limited attention thus far.

\section{Mixup-Based Graph Learning via Hybrid Pooling}
\subsection{Overview}
\label{mixup}

\begin{figure*}[!tb]
	\centering
	\includegraphics[width=1.0\linewidth]{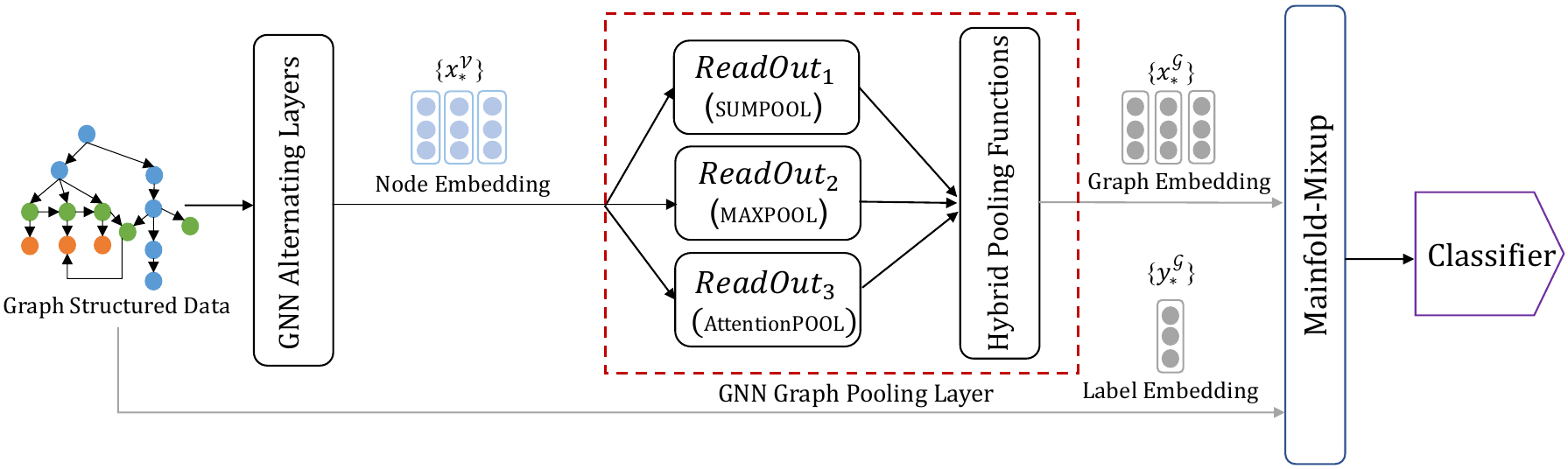}
	\caption{Overview of Mixup-based graph learning via hybrid pooling.}
	\label{fig:overview}
\end{figure*}

Figure \ref{fig:overview} provides an overview of GNNs for graph classification, demonstrating the application of \emph{Manifold-Mixup} after the hybrid pooling layer. Concretely, first, GNNs process the input graph-structured data and transform them into node attributes $\left\{x_i^{\mathcal{V}}\right\}_{i=1}^n$, where $\mathcal{V}$ is the vertex set. Then, after learning the features of each node, the hybrid pooling layer produces the entire graph embedding by utilizing three fundamental types of readout functions: Sum-pooling (SUMPOOL), Max-pooling (MAXPOOL), and Attention-pooling (AttentionPOOL). Finally, as shown in Eq.~(\ref{eqn:mixup}), \emph{Manifold-Mixup} is applied to randomly mix  two selected graph embeddings $\widetilde{{x}_{i}^{\mathcal{G}}}$, $\widetilde{{x_{j}^{\mathcal{G}}}}$ and their ground truth labels ${y_{i}^{\mathcal{G}}}$, ${y_{j}^{\mathcal{G}}}$ with one-hot values as the new training set. This augmented training set is then used for training the classifier. Especially, the utilization of \emph{Manifold-Mixup} employing the hybrid pooling layer is depicted as follows:
\begin{equation}\label{eqn:mixup}
\begin{aligned}
\widetilde{x_{mix}^{\mathcal{G}}} = \lambda \widetilde{x_{i}^{\mathcal{G}}} + (1 - \lambda) \widetilde{x_{j}^{\mathcal{G}}}\\
 y_{mix}^{\mathcal{G}} = \lambda y_{i}^{\mathcal{G}} + (1 - \lambda) y_{j}^{\mathcal{G}} %\nonumber
\end{aligned}
\end{equation}
\noindent where, $\widetilde{x^{\mathcal{G}}}$ is the graph embedding and $y^{\mathcal{G}}$ refers to its label embedding with one-hot value, wherein $\lambda$ means the \emph{Mixup} ratio. According to \cite{zhang2018mixup, guo2019augmenting}, $\lambda$ is sampled from a Beta distribution with a shape parameter $\alpha\left(\lambda\sim\textit{Beta}\left(\alpha,\alpha\right)\right)$.

\subsection{Graph-level hybrid pooling}
\label{poolingOP}

In recent research~\cite{wang2021mixup}, several methods have been proposed to represent graphs at different levels of granularity. These include node-level representation (e.g., node embeddings) and graph-level representation (e.g., graph embeddings), each serving different purposes: node embeddings can be utilized for tasks such as node classification and link prediction, while graph embeddings can be used for graph classification and graph generation~\cite{wu2020comprehensive}. Our work specifically focuses on graph-level representation and its related downstream tasks, e.g., graph classification.

Generally, GNNs learn the graph representation by exploiting the graph structure as an inductive bias, and the GNN architecture proposed by Gilmer \emph{et al.}~\cite{gilmer2017neural} is the most popular and widely used in the applications. Briefly speaking, first, given a node with its initialized information, GNN computes the representation by iteratively aggregating its adjacent nodes (\textbf{Aggregate}). Then, it combines this aggregated representation with the existing node representation (\textbf{Update}). After that, a final representation of the complete graph is created by pooling learned features of nodes (\textbf{Pooling}). The primary areas where the various models differ are how they handle aggregation, update, and pooling.

%Mathematically, 
Formally, given a graph-structured data $\mathcal{G}\left(\mathcal{V},\varepsilon \right)$, where $\mathcal{V}$ is the vertex set, and $\varepsilon$ represents the edge set, we simply formulate the entire graph  representation $x^{\mathcal{G}}$ as follows:

\begin{equation}\label{eqn:gnn1}
\small
\begin{aligned}
    \mathbf{H}_{u}^{k} &= \mathbf{GNN}\left(\mathbf{H}_{u}^{k-1}, W, \sum_{v\in \mathcal{N}\left(u\right)}\mathbf{H}_{v}^{k-1}\right) \\
    \mathbf{H}^{\mathcal{G}} &= \mathcal{P}\left(%\mathbf{H}_{u,u\in \mathcal{V}}^{k},
    H,
    \mathcal{V}\right) \quad\text{s.t.}\quad H = \mathbf{H}_{u}^{k} \quad(u\in \mathcal{V})
\end{aligned}
\end{equation}

\noindent where $\mathbf{H}_{u}^{k-1}$ and $\mathbf{H}_{v}^{k-1}$ denote the matrix representation of node $u$ and $v$ ($u,v\in\mathcal{V}$, $uv\in \varepsilon$) at the $(k-1)$-\emph{th} iteration, $W$ represents the trainable parameter matrix, $\mathcal{N}\left(u\right)$ is the set of neighbor nodes of $u$, and $\mathcal{P}$ is a graph pooling operator that produces $\mathbf{H}^{\mathcal{G}}$, the latent vector of the entire graph $x^{\mathcal{G}}$. In Eq.~\ref{eqn:gnn1},   firstly, neural network $\mathbf{GNN}\left(\cdot\right)$ is used to iteratively update the latent vector of each node via the message aggregated from the neighborhood  $\sum\nolimits_{v\in \mathcal{V}\left(u\right)} \mathbf{H}_{v}^{k-1}$. After that, the pooling operator $\mathcal{P}\left(\cdot\right)$ is used to construct the vector representation of the entire graph, which captures the global information, after $k$ steps of iteration. 

In our study, we consider three standard pooling operators $\mathcal{P}$, which are as follows:

\begin{equation}\label{eqn:gnn2}
\small
\begin{aligned}
    \mathcal{P}_{att}\left(H,\mathcal{V}\right) &= \sum_{u\in\mathcal{V}}
    %\limits_{i =1}^{Num\left(\mathcal{V}\right)} 
    \sigma\left(W\mathbf{H}^{k}_{u} +b\right)\odot\phi\left(W\mathbf{H}_{u}^{k}+b\right) \\
    \mathcal{P}_{sum}\left(H,\mathcal{V}\right) &= \sum_{u\in\mathcal{V}} 
    %\limits_{i =1}^{Num\left(\mathcal{V}\right)}
    \mathbf{H}_{u}^{k}\\
    \mathcal{P}_{max}\left(H,\mathcal{V}\right) &= \max_{u\in\mathcal{V}}%\nolimits_{i =1}^{Num\left(\mathcal{V}\right)}
    \mathbf{H}_{u}^{k} 
\end{aligned}
\end{equation}

\noindent where $\phi$ represents the nonlinear activation function, $\odot$ means the Hadamard Product and $\mathbf{H}_{i}^{k}$ refers to the final vector representation of the $i$-th node. Here, $\sigma\left(W\mathbf{H}^{k}_{i} +b\right)$ acts as a soft attention mechanism. We leverage the global attention pooling $\mathcal{P}_{att}$ as it better captures relevant global features for graph-level tasks~\cite{li2015gated}. 

Finally, we examine three hybrid pooling functions $\mathcal{M}_{sum}$, $\mathcal{M}_{mul}$ and $\mathcal{M}_{concat}$, defined as follows:  
%\begin{equation}\label{eqn:gnn2}
%\begin{aligned}
%    \mathcal{M}_{sum}(\mathcal{P},h) &= \textbf{h}_{\mathcal{P}_i}^{k} +  \textbf{h}_{\mathcal{P}_j}^{k}  \\
%    \mathcal{M}_{mul}(\mathcal{P},h) &= \textbf{h}_{\mathcal{P}_i}^{k} \odot   \textbf{h}_{\mathcal{P}_j}^{k}\\
%    \mathcal{M}_{concat}(\mathcal{P},h) &= W(\big[ \textbf{h}_{\mathcal{P}_i}^{k} \parallel  %\textbf{h}_{\mathcal{P}_j}^{k} \big])+b\nonumber
%\end{aligned}
%\end{equation}

\begin{equation}\label{eqn:gnn3}
\begin{aligned}
    \mathcal{M}_{sum}\left(\mathcal{P}_i, \mathcal{P}_j\right) &= \mathbf{H}_{\mathcal{P}_i}^\mathcal{G} + \mathbf{H}_{\mathcal{P}_j}^\mathcal{G} \\
    \mathcal{M}_{mul}\left(\mathcal{P}_i, \mathcal{P}_j\right) &= \mathbf{H}_{\mathcal{P}_i}^\mathcal{G} \odot \mathbf{H}_{\mathcal{P}_j}^\mathcal{G}\\
    \mathcal{M}_{concat}\left(\mathcal{P}_i, \mathcal{P}_j\right)  &= T\left(\left[ \mathbf{H}_{\mathcal{P}_i}^\mathcal{G} \parallel \mathbf{H}_{\mathcal{P}_j}^\mathcal{G}  \right]\right)+b
\end{aligned}
\end{equation}

\noindent where $\mathbf{H}_{\mathcal{P}_i}^\mathcal{G}$, $\mathbf{H}_{\mathcal{P}_j}^\mathcal{G}$ means the embedding vectors of the entire graph, respectively produced by two different pooling operators $\mathcal{P}_i, \mathcal{P}_j$ 
 $(\mathcal{P}_i \neq \mathcal{P}_j)$, as defined in Eq.~(\ref{eqn:gnn2}). Here, $\parallel$ denotes the vector concatenation operation, and $T$ is a linear transformation matrix to reduce the dimension $\left(\mathbb{R}^{2d}\rightarrow {\mathbb{R}^{d}}\right)$.

To conclude, the hybrid pooling layer, which consists of the pooling function and pooling operator, can be expressed as follows: 
\begin{equation}\label{eqn:gnn3}
\begin{aligned}
%\rho(p_{1},p_{2}, m\{x_i^{\mathcal{V}}\}_{i=1}^n) =  p_{1}(\{x_i^{\mathcal{V}}\}_{i=1}^n)), p_{1},p_{2} \in P{3 \choose 2}, m \in M{3 \choose 1}\\
 \widetilde{{H}^{\mathcal{G}}}=\sum_{u\in\mathcal{V}} \mathcal{M}_{n}(\mathcal{P}, H_{u}^{k}) %\nonumber
\end{aligned}
\end{equation}
\noindent where $\mathcal{M}_{n} \in \left\{\mathcal{M}_{sum}, \mathcal{M}_{mul},\mathcal{M}_{concat}\right\}$ and $\widetilde{{H^{\mathcal{G}}}}$ is the matrix representation of the graph data $\mathcal{G}$ produced by the hybrid pooling layer. To more clearly illustrate the design, Table~\ref{hybrid_pooling_layer} lists the 9 different types of hybrid pooling operators, which are considered in our study.
\begin{table}[!tb]
    \caption{Nine Types of Hybrid Pooling }
    \label{hybrid_pooling_layer}
\centering
\resizebox{0.95\textwidth}{!}{

    \begin{tabular}{llll}
        \toprule
        \multirow{2}*{} & \multicolumn{3}{c}{Hybrid Pooling Layer}   \\
        \cmidrule{2-4}
                                & \multicolumn{1}{l}{Pooling Function} & Pooling Operator & \multicolumn{1}{c}{Description} \\
        \midrule
        Type1  &$\mathcal{M}_{sum}$       & $\mathcal{P}_{att}, \mathcal{P}_{max}$    & Perform matrix addition on the vectors generated by $\mathcal{P}_{att}$ and $\mathcal{P}_{max}$\\
        Type2  & $\mathcal{M}_{mul}$      & $\mathcal{P}_{att}, \mathcal{P}_{max}$    & Perform Hadamard product on the vectors generated by $\mathcal{P}_{att}$ and $\mathcal{P}_{max}$\\
        Type3  & $\mathcal{M}_{concat}$   & $\mathcal{P}_{att}, \mathcal{P}_{max}$     & Concatenate the vectors that are generated by $\mathcal{P}_{att}$ and $\mathcal{P}_{max}$\\
        Type4  & $\mathcal{M}_{sum}$      & $\mathcal{P}_{att}, \mathcal{P}_{sum} $     & Perform matrix addition on the vectors generated by $\mathcal{P}_{att}$ and $\mathcal{P}_{sum}$\\
        Type5  & $\mathcal{M}_{mul}$      & $\mathcal{P}_{att}, \mathcal{P}_{sum}$     & Perform Hadamard product on the vectors generated by $\mathcal{P}_{att}$ and $\mathcal{P}_{sum}$\\
        Type6  & $\mathcal{M}_{concat}$   & $\mathcal{P}_{att}, \mathcal{P}_{sum}$  & Concatenate the vectors that are generated by $\mathcal{P}_{att}$ and $\mathcal{P}_{sum}$\\
        Type7  & $\mathcal{M}_{sum}$      & $\mathcal{P}_{sum}, \mathcal{P}_{max} $  & Perform matrix addition on the vectors generated by $\mathcal{P}_{sum}$ and $\mathcal{P}_{max}$  \\
        Type8  & $\mathcal{M}_{mul}$      & $\mathcal{P}_{sum},\mathcal{P}_{max} $   & Perform Hadamard product on the vectors generated by $\mathcal{P}_{sum}$ and $\mathcal{P}_{max}$ \\
        Type9  & $\mathcal{M}_{concat}$   & $\mathcal{P}_{sum},\mathcal{P}_{max}$    & Concatenate the vectors that are generated by $\mathcal{P}_{sum}$ and $\mathcal{P}_{max}$\\
        \bottomrule
    \end{tabular}}
\end{table}

%\subsection{Mixup via hybrid pooling}
%\label{mixup}
%The utilization of \emph{Manifold-Mixup} employing the hybrid pooling layer is depicted as follows:
%\begin{equation}\label{eqn:mixup}
%\begin{aligned}
%\widetilde{x_{mix}^{\mathcal{G}}} = \lambda \widetilde{x_{i}^{\mathcal{G}}} + (1 - \lambda) \widetilde{x_{j}^{\mathcal{G}}}\\
% y_{mix}^{\mathcal{G}} = \lambda y_{i}^{\mathcal{G}} + (1 - \lambda) y_{j}^{\mathcal{G}} %\nonumber
%\end{aligned}
%\end{equation}
%\noindent where, $\widetilde{x^{\mathcal{G}}}$ is the graph embedding and $y^{\mathcal{G}}$ refers to its label embedding with one-hot value, wherein $\lambda$ means the \emph{Mixup} ratio. According to \cite{zhang2018mixup, guo2019augmenting}, $\lambda$ is sampled from a Beta distribution with a shape parameter $\alpha\left(\lambda\sim\textit{Beta}\left(\alpha,\alpha\right)\right)$.

\section{Experimental Setup}
\subsection{Case Study}

\begin{figure}[!tb]
    \centering       
    \subfigure[Effectiveness of Accuracy]{
    \includegraphics[scale=0.45]{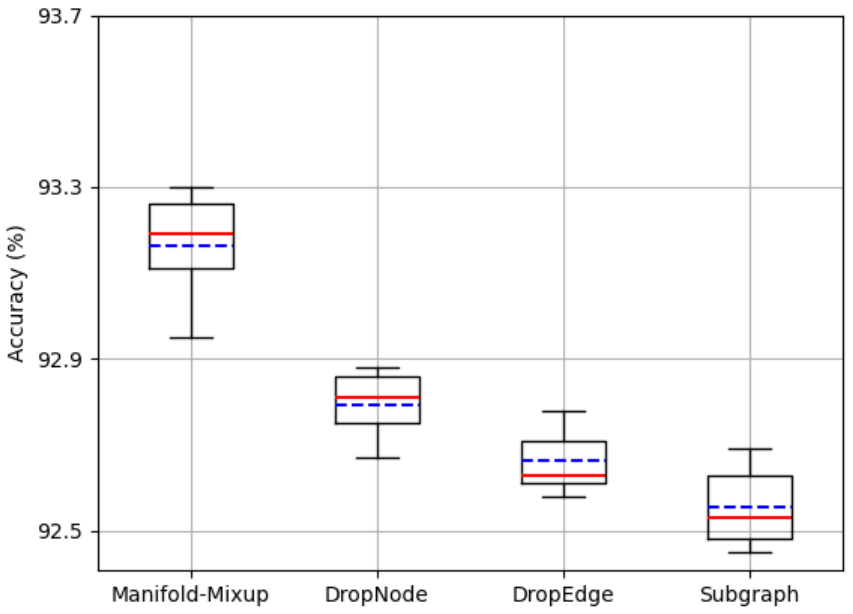}
    } 
    \subfigure[Effectiveness of Robustness]{
    \includegraphics[scale=0.45]{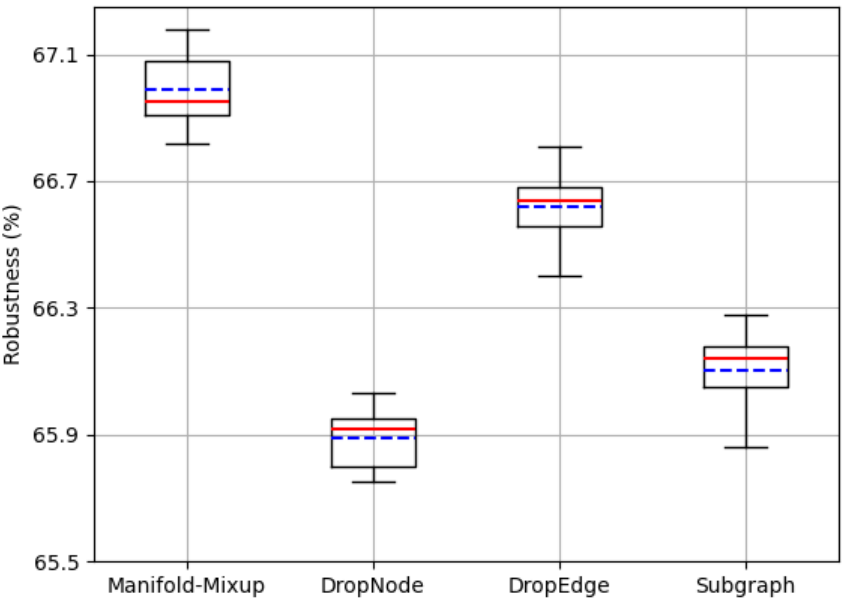}
    }
    \caption{A case study of graph data augmentation methods (Model: GCN, Dataset: Gossipcop-Profile).}
    \label{fig:case_study}
\end{figure}

We conduct a case study to evaluate the effectiveness of existing graph data augmentation methods, in terms of accuracy and robustness. Specifically, 1) we perform a survey to collect graph data augmentation methods from existing studies~\cite{zhao2022graph, dong2024effectiveness}. As a result, graph data augmentation methods including \emph{Manifold-Mixup}~\cite{verma2019manifold}, \emph{DropNode}~\cite{feng2020graph}, \emph{DropEdge}~\cite{rong2020dropedge}, and \emph{Subgraph}~\cite{wang2020graphcrop}, which are adopted in our experiment. 2) We subsequently apply these methods to train GCN using the NLP dataset Gossipcop-Profile. In particular, the model training is repeated five times to mitigate the influence of randomness. 3) Finally, we record the results of each model training.  Figure~\ref{fig:case_study} represents our experimental results. From the left part of Figure~\ref{fig:case_study}, we can see that Manifold-Mixup shows the best performance compared to all considered graph data augmentation methods in the evaluation of accuracy. Moving to robustness, we observe a similar finding where Manifold-Mixup can consistently produce a more robust GCN model compared to other methods. Interestingly, although Subgraph showing lower accuracy improvement performance compared to DropNode, it exhibits better robustness improvement. 

Manifold-Mixup has demonstrated its effectiveness on both graph data~\cite{wang2021mixup,zhao2022graph} and PL data~\cite{dong2024effectiveness}. The findings from our experiments are consistent with those of existing studies. By interpolating two different data points in the hidden space, Manifold-Mixup can help the GNN model learn more about the underlying data structure, thereby enhancing the accuracy performance. Additionally, the features generated by Manifold-Mixup result in more evenly distributed class-specific representations compared to other data augmentation methods such as Subgraph and DropNode. The smoothing effect on the decision boundary of classification induced by mixup can reduce the impact of these noises on the model's behavior, thereby generating more robust GNN models~\cite{verma2019manifold, dong2024effectiveness}.

As mentioned in Section~\ref{sec:intro}, the performance of Manifold-Mixup is mainly influenced by two factors including the Mixup ratio and Mixup strategies that are associated with the graph representation. While the former has been extensively studied, the impact of graph representations on Manifold-Mixup is rarely discussed. Therefore,  considering its superior performance, we select Manifold-Mixup from existing graph data augmentation in our study and explore the potential of graph representations to further improve the performance of Manifold-Mixup.

\noindent\colorbox{gray!20}{\framebox{\parbox{0.96\linewidth}{
\textbf{Finding}:  Manifold-Mixup shows the best performance compared to other existing graph data augmentation methods including Subgraph, DropNode, and DropEdge in terms of accuracy and robustness improvement.} }}

\subsection{Study Design}
As discussed in Section~\ref{sec:intro}, the success of Mixup, especially Manifold-Mixup, has led to the widespread adoption of Mixup-based data augmentation in graph learning, establishing it as a mainstream paradigm~\cite{zhao2022graph}. Therefore, in our study, our main focus is on exploring the potential of hybrid pooling operators for Mixup-based graph learning to enhance both accuracy and robustness performance. To assess the effectiveness of hybrid pooling operators, as introduced in Section~\ref{poolingOP}, in Mixup-based graph learning, we design three research questions, as follows:

\begin{compactitem}[$\bullet$]
\item \textbf{RQ1: How effective are hybrid pooling operators for enhancing the accuracy of Mixup-based graph learning?} 
We first explore whether hybrid pooling operators are able to help graph learning with \emph{Manifold-Mixup} improve the accuracy of GNN models or not. Accuracy which calculates the percentage of correctly identified data among the total number of data in the given test data is the basic metric for evaluating the performance of trained models. Therefore, in this RQ we first evaluate the accuracy performance of hybrid pooling operators (as introduced in Section~\ref{poolingOP}) on the NLP binary classification dataset and PL multi-classification dataset, compared to baseline approaches including Max-pooling and the state-of-the-art GMT pooling operators (as introduced in~\ref{sec:bgpoolingop}). 

\smallskip
\item \textbf{RQ2: How effective are hybrid pooling operators for enhancing the robustness of Mixup-based graph learning?}
Robustness is another important metric to reflect the generalization ability of the trained GNN models in practice~\cite{xu2012robustness,tu2000robust,neyshabur2017exploring}. In this RQ, we first generate robust test sets by the method~\cite{papp2021dropgnn} that randomly drops the edge-connectivity (i.e., the local connections between a node and its neighboring nodes) of each graph data from the original sets and then calculate the percentage of correctly classified data from this new robust set, to evaluate the robustness of hybrid pooling operators when applied to Mixup-based graph learning~\cite{fabian2015topological,oehlers2021graph}. 

%In this RQ, to evaluate the robustness of hybrid pooling operators in
%Mixup-based graph learning, we create a new robust test set following previous works~\cite{fabian2015topological,oehlers2021graph}, which focus on evaluating the robustness of GNNs. In detail, given structured data, we employ a technique called local connectivity dropout which involves randomly dropping the local connections between a node and its neighboring nodes, resulting in modified topological features and the generation of new data samples.

\smallskip
\item \textbf{RQ3: How does the hyperparameter setting affect the effectiveness of Manifold-Mixup when hybrid pooling operators are applied?} 
The hyperparameter $\lambda$ (as introduced in Section~\ref{mixup}) affects the performance including both accuracy and robustness of Mixup-based graph learning~\cite{zhang2018mixup,dong2023mixcode,verma2019manifold}. Therefore, in this RQ we investigate whether this conclusion also holds for Mixup-based graph learning when hybrid pooling operators are applied. Concretely, we evaluate the accuracy and robustness of nine hybrid pooling operators under the setting of the \emph{Mixup} ratio $\alpha$ from 0.05 to 0.5 in 0.05 intervals.

\end{compactitem}

\subsection{Task, Dataset, and Model}
\begin{figure}[!tb]
    \centering       
    \subfigure[Fake news detection]{
    \includegraphics[scale=0.40]{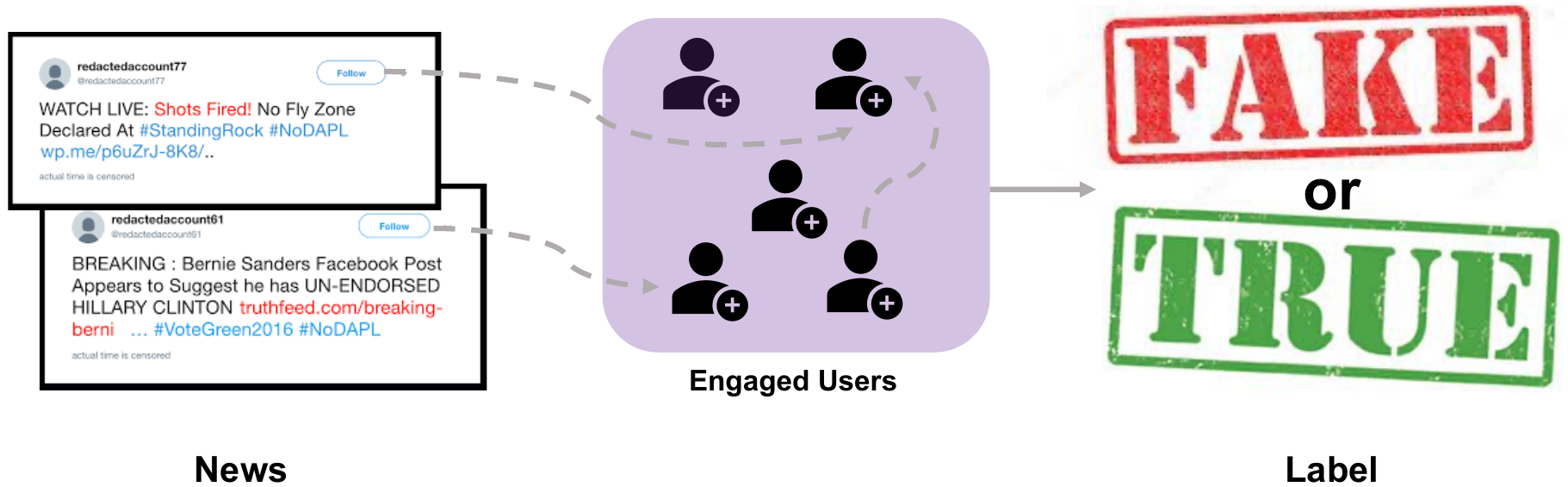}
    }
    \subfigure[Program classification]{
    \includegraphics[scale=0.40]{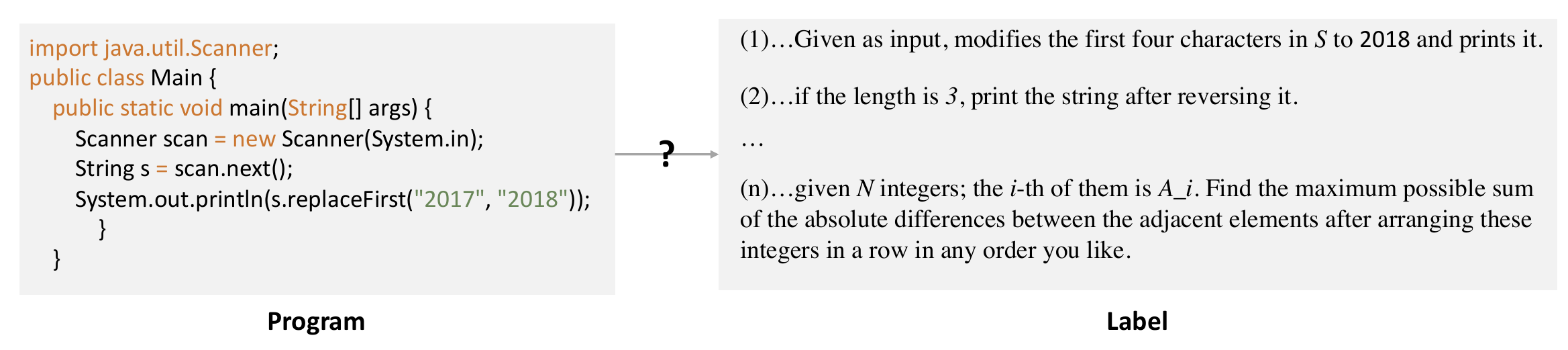}
    }
    
    \caption{Examples of Fake news detection task and Program classification task.}
    \label{fig:examples_NLP_PL}
\end{figure}

\noindent\textbf{Task.} In our study, we focus on two widely studied tasks including fake news detection and program classification. We provide a simple example to understand these two tasks, as depicted in Figure~\ref{fig:examples_NLP_PL}. The proliferation of fake news on the internet presents a pressing social concern. To address this issue, researchers generally conceptualize Twitter users and news articles as nodes within a social network graph, with (re)tweeting actions represented as edges. Positioned at the intersection of graph learning and NLP, GNNs are applied to propagate text embeddings of nodes and aggregate them for binary classification (i.e., Fake or True)~\cite{dou2021user}. 
Since the problem classification task is relatively new in the PL field, we also use one example to better explain it. When provided with a series of problems accompanied by detailed descriptions and their corresponding candidate source code, trained models can effectively identify the problem that the program solves. Moreover, representing the program as a graph allows for capturing semantic content~\cite{allamanis2018survey,dong2024effectiveness}. Therefore, GNNs can be applied to learn the graph information of programs. Program classification task involves servers that automatically estimate the functionality of programs, which is crucial for software reuse~\cite{puri2021codenet}.

\noindent\textbf{Dataset.} We conduct our study on both traditional NLP and PL tasks. For NLP, we consider User Preference-aware Fake News Detection (UPFD)~\cite{dou2021user}, which is a text-level binary classification problem. For PL, we consider function-level problem multi-classification task provided by Project\_CodeNet~\cite{puri2021codenet}. Two popular programming languages, Java and Python, are included in our study. 

UPFD contains two sets of tree-structured fake and real news propagation graphs that are derived from Twitter. Given a single graph, the source news is represented by the root node, and the leaf nodes represent Twitter users who retweeted the root news. Edges reflect the 1) connection between users and the 2) connection between the user and root news that have been retweeted. Moreover, UPFD also includes 4 different node feature types. These are \ding{202} \emph{Profile} feature, a 10-dimensional vector derived from ten user profile attributes; \ding{203} \emph{Spacy} feature, a 300-dimensional vector encoded using spacy Word2Vec; \ding{204} \emph{Bert} feature, a 768-dimensional vector encoded by BERT (Bidirectional Encoder Representations from Transformers); \ding{205} \emph{Content} feature, a 310-dimensional vector obtained by summing the \emph{Spacy} and \emph{Profile} features. Fake news detection is commonly approached as a binary classification task, where GNNs provide a negative or positive prediction given a graph of user preferences.
Project\_CodeNet provides two datasets, JAVA250 and Python800. JAVA250 consists of 250 classification problems, and each problem has 300 Java programs. Python800 contains 800 classification problems, and each problem has 300 Python programs. In addition, we transformed the raw source code data from both datasets into a graph representation by utilizing a simplified parse tree. We follow the same process as the original paper to divide the datasets into training, validation, and test sets.

\smallskip
\noindent\textbf{Model.} We build 4 different GNN architectures for each dataset. In the text classification, we follow the recommendation of \cite{dou2021user} and build Graph Convolution Network (GCN) \cite{welling2016semi}, Graph Isomorphism Network (GIN) \cite{xu2018powerful}, Graph Attention Network (GAT) \cite{velickovic2017graph}, and GraphSAGE \cite{hamilton2017inductive} models. GCN is a specialized variant of a convolution neural network that is dedicated to operating graph-structured data. GAT mainly uses the attention mechanism for graph message passing. GIN is designed to generalize the Weisfeiler-Lehman (WL) test. GraphSAGE effectively generates node embeddings for previously undiscovered data by utilizing node feature information. In program classification, we use GNN models (GCN, GCN-Virtual, GIN, and GIN-Virtual) provided by \cite{puri2021codenet}. Especially to enhance the aggregation phase of GNNs, virtual nodes, which involve adding an artificial node to each network and connecting it in both directions to all other graph nodes, are offered. To sum up, Table\ref{tab:data_model} shows the details of used datasets and models in our experiment.

% Please add the following required packages to your document preamble:
\begin{table}[!tb]
\centering
\caption{Details of datasets and DNNs. \#Training, \#Ori Test, and \#Robust Test represent the number of training data, original test data, and test data for generalization evaluation, respectively.}
\label{tab:data_model}
%\resizebox{.9\columnwidth}{!}{
{\small
\begin{tabular}{llcccc}
\toprule
\textbf{Dataset} & \textbf{Task} & \textbf{\#Training} & \textbf{\#Clean Test} & \textbf{\#Robust Test} & \textbf{Model} \\ \hline
\multirow{2}{*}{\textbf{Politifact}} & \multirow{2}{*}{Fake news detection} & \multirow{2}{*}{62} & \multirow{2}{*}{221} & \multirow{2}{*}{314} & GCN \\
 &  &  &  &  & GIN \\ \cline{1-5}
\multirow{2}{*}{\textbf{Gossipcop}} & \multirow{2}{*}{Fake news detection} & \multirow{2}{*}{1,092} & \multirow{2}{*}{3,826} & \multirow{2}{*}{5,464} & GAT \\
 &  &  &  &  & GraphSAGE \\ \hline
\multirow{2}{*}{\textbf{JAVA250}} & \multirow{2}{*}{Problem classification} & \multirow{2}{*}{48,000} & \multirow{2}{*}{15,000} & \multirow{2}{*}{75,000} & GCN \\
 &  &  &  &  & GCN-Virtual  \\ \cline{1-5}
\multirow{2}{*}{\textbf{Python800}} & \multirow{2}{*}{Problem classification} & \multirow{2}{*}{153,600} & \multirow{2}{*}{48,000} & \multirow{2}{*}{240,000} & GIN \\
 &  &  &  &  & GIN-Virtual \\ \hline
\end{tabular}
}%}
\end{table}

\subsection{Experiment Settings}
We implement the hybrid pooling layer based on the open-source library \textbf{PyG} (Pytorch Geometric) \cite{Fey/Lenssen/2019}. The training epochs we set for Project\_CodeNet and UPFD are 100 and 200, respectively. The batch size is set to 80 for Project\_CodeNet and 128 for UPFD. For the optimizer, we use Adam \cite{kingma2014adam} with the learning rate $10^{-3}$ for all the models. For the Mixup ratio, $\alpha = 0.1$ is our default setting. To alleviate overfitting, we adopt early stopping with patience 20. To reduce the impact of randomness, following the convention in existing works~\cite{dong2023mixcode,puri2021codenet}, we train each GNN model five times by using different random seeds. We present the average experimental results with standard deviation in the later Section. All the experiments were conducted on a server with 2 GPUs of NVIDIA RTX A6000.  

\section{Results Analysis}
We first study the influence of graph pooling operators on the GNN training with \emph{Manifold-Mixup} in RQ1 and RQ2, then we go deeper into the hyperparameter settings of \emph{Manifold-Mixup} in RQ3. 
\subsection{RQ1: How effective are hybrid pooling operators for enhancing the accuracy of Mixup-based graph learning? }

\textbf{Accuracy analysis.} The left part in Table \ref{table:results1} presents the detailed test accuracy results of the GCN model on NLP datasets. From the results, first, we can see that \emph{Manifold-Mixup} is a powerful technique for augmenting graph-structured data. Compared to the model trained without \emph{Manifold-Mixup} (\textit{No Aug} in the tables), GCN model training with data augmentation is more accurate and robust. Then, concerning different pooling operators, we can see that the basic and widely used Max-pooling method (MAXPOOL) and the state-of-the-art GMT pooling operators (GMT) are not the best choices for GCN training with \emph{Manifold-Mixup}. There are always some hybrid pooling operators that achieve better results than them. More specifically, 8 out of 9 hybrid pooling operators outperform MAXPOOL on the Gossipcop-Profile. Similarly, on the Politifact-Bert, 5 out of 9 hybrid pooling operators achieve higher performance than MAXPOOL. Among these results, the Type 1 operator outperforms GMT by up to 2.14\% on Gossipcop and 4.38\% on Politifact, respectively. On average, for training with the GCN model, the Type 1 operator demonstrates the best performance under the NLP dataset. Moreover, maybe surprisingly, the gap between using the hybrid pooling operator and MAXPOOL can be up to 2.55\% of clean accuracy (Gossipcop, Type1, Profile). Moving to the average improvement, we primarily focus on Type 1, Type 3, and Type 9, as they exhibit superior accuracy improvement compared to other hybrid pooling operators, as indicated in the Average column. In GCN model training on the NLP dataset, Type 1, Type 3, and Type 9 exhibit average accuracy improvements of up to 1.08\%, 0.64\%, and 0.35\%, respectively, compared to MAXPOOL. On average, Type 1, Type 3, and Type 9 outperform  GMT by up to 1.02\%, 0.68\%, and 0.45\%, respectively. Overall, across both the maximum and average improvement values, Type 1 consistently outperforms other hybrid pooling operators.

The left part in Table \ref{table:Pool_gat} presents the results of the GAT model. Firstly, in both the Gossipcop-Profile and Politifact-Bert datasets, 4 out of the 9 hybrid pooling operators achieve higher test accuracy than MAXPOOL. Moreover, based on the results, the hybrid pooling operator (Type 3) brings the accuracy improvement by up to 2.31\% than GMT on the Gossipcop-Spacy, and the operator Type 1 outperforms GMT by up to 2.16\% on the Politifact-Content. On average, when training with the GAT model, the hybrid operator (Type 3) proves to be the optimal choice for the NLP dataset. It achieves the highest test accuracy on both the Gossipcop and Politifact datasets. We observe three different hybrid pooling operators, namely Type 1, Type 2, and Type 3, as they exhibit better accuracy improvement compared to others, as indicated in the Average column. Compared to MAXPOOL, Type 1, Type 2, and Type 3 exhibit average accuracy improvements of up to 0.59\%, 0.30\%, and 0.74\%.  Moving to GMT, Type 3 achieves the highest average accuracy improvement of 1.30\%, followed by Type 1 at 1.26\% and Type 2 at 0.91\%. Overall, Type 3 shows the best performance in both maximum and average improvement analyses in GAT training.

The left part in Table \ref{table:Pool_sage} presents the results of the NLP dataset when training with the GraphSAGE model. Hybrid pooling operators can always help \emph{Manifold-mixup} in improving test accuracy. For example, in evaluating the performance of hybrid pooling operators, we observe that 4 out of the 9 operators achieve better performance compared with MAXPOOL on both Gossipcop-Profile and Politifact-Bert. Interestingly, while operator Type 1 demonstrates the best performance in most cases, the results from the column Average indicate that operator Type 3 is the optimal choice for the Gossipcop dataset. Type 1 and Type 3 demonstrate the best performance, as indicated by the Average column. Compared to MAXPOOL, Type 1 performs an average accuracy improvement of 0.32\%, which is slightly better than Type 3, at 0.30\%. The same conclusion holds in the comparison with GMT, where Type 1 (1.50\% on average) still outperforms Type 3 (1.32\% on average).

\begin{table*}[!tb]
\caption{Effectiveness of nine types of hybrid pooling w.r.t. test accuracy $\uparrow$ (average $\pm$ standard deviation, \%) on original test data and robustness $\uparrow$ (average $\pm$ standard deviation, \%) on robust test data of trained GCN models on NLP datasets. \textbf{No Aug}: without Manifold-Mixup data augmentation. A gray background highlights results that are better than MAXPOOL. The best average results are marked in red color.}
\label{table:results1}
\centering
\scalebox{0.55}{
\begin{tabular}{llllllc||llllc}
\hline
 &  & \multicolumn{5}{c||}{\textbf{Test Accuracy}} & \multicolumn{5}{c}{\textbf{Robustness}} \\ 
 &  & \multicolumn{1}{c}{\textbf{Profile}} & \multicolumn{1}{c}{\textbf{Spacy}} & \multicolumn{1}{c}{\textbf{Bert}} & \multicolumn{1}{c}{\textbf{Content}} & \textbf{Average} & \multicolumn{1}{c}{\textbf{Profile}} & \multicolumn{1}{c}{\textbf{Spacy}} & \multicolumn{1}{c}{\textbf{Bert}} & \multicolumn{1}{c}{\textbf{Content}} & \textbf{Average} \\ \hline
 & \textbf{No Aug} & 90.48 ± 0.02 & 96.19 ± 0.07 & 95.73 ± 0.54 & 95.44 ± 0.72 & 94.46 & 63.19 ± 0.63 & 85.31 ± 1.73 & 89.01 ± 0.68 & 77.17 ± 0.76 & 78.67 \\
 & \textbf{MAXPOOL} & 93.17 ± 0.87 & 96.52 ± 0.09 & 96.12 ± 0.13 & 95.79 ± 0.45 & 95.40 & 66.97 ± 1.06 & 86.28 ± 1.21 & 90.51 ± 0.35 & 78.98 ± 1.24 & 80.69 \\
 & \textbf{GMT} & \cellcolor[HTML]{C0C0C0}93.58 ± 0.32 & 96.44 ± 0.07 & \cellcolor[HTML]{C0C0C0}96.25 ± 0.19 & \cellcolor[HTML]{C0C0C0}96.57 ± 0.02 & \cellcolor[HTML]{C0C0C0}95.71 & \cellcolor[HTML]{C0C0C0}75.37 ± 2.93 & 74.38 ± 0.02 & 72.39 ± 1.74 & \cellcolor[HTML]{C0C0C0}94.07 ± 1.11 & 79.05 \\
 & \textbf{Type1} & \cellcolor[HTML]{C0C0C0}95.72 ± 0.27 & \cellcolor[HTML]{C0C0C0}96.84 ± 0.11 & \cellcolor[HTML]{C0C0C0}96.71 ± 0.26 & \cellcolor[HTML]{C0C0C0}96.96 ± 0.21 & \cellcolor[HTML]{C0C0C0}{\color[HTML]{FE0000} 96.56} & \cellcolor[HTML]{C0C0C0}73.31 ± 1.22 & \cellcolor[HTML]{C0C0C0}87.58 ± 0.87 & \cellcolor[HTML]{C0C0C0}91.04 ± 1.25 & \cellcolor[HTML]{C0C0C0}93.31 ± 0.15 & \cellcolor[HTML]{C0C0C0}86.31 \\
 & \textbf{Type2} & \cellcolor[HTML]{C0C0C0}94.11 ± 0.26 & 95.73 ± 0.24 & \cellcolor[HTML]{C0C0C0}96.24 ± 0.04 & \cellcolor[HTML]{C0C0C0}97.36 ± 0.14 & \cellcolor[HTML]{C0C0C0}95.86 & \cellcolor[HTML]{C0C0C0}71.77 ± 1.43 & 72.08 ± 1.22 & 76.78 ± 1.89 & \cellcolor[HTML]{C0C0C0}93.36 ± 1.88 & 78.50 \\
 & \textbf{Type3} & \cellcolor[HTML]{C0C0C0}95.37 ± 0.33 & \cellcolor[HTML]{C0C0C0}96.64 ± 0.01 & \cellcolor[HTML]{C0C0C0}96.63 ± 0.09 & \cellcolor[HTML]{C0C0C0}96.79 ± 0.13 & \cellcolor[HTML]{C0C0C0}96.36 & \cellcolor[HTML]{C0C0C0}67.39 ± 1.65 & \cellcolor[HTML]{C0C0C0}87.72 ± 2.21 & 89.99 ± 1.88 & \cellcolor[HTML]{C0C0C0}83.86 ± 0.12 & \cellcolor[HTML]{C0C0C0}82.24 \\
 & \textbf{Type4} & \cellcolor[HTML]{C0C0C0}95.14 ± 0.25 & 96.28 ± 0.21 & 94.43 ± 0.59 & 95.45 ± 0.11 & \cellcolor[HTML]{C0C0C0}95.33 & \cellcolor[HTML]{C0C0C0}90.24 ± 0.21 & \cellcolor[HTML]{C0C0C0}87.33 ± 0.88 & 77.64 ± 0.95 & \cellcolor[HTML]{C0C0C0}94.23 ± 1.13 & \cellcolor[HTML]{C0C0C0}{\color[HTML]{FE0000} 87.36} \\
 & \textbf{Type5} & \cellcolor[HTML]{C0C0C0}93.64 ± 0.35 & 96.06 ± 0.04 & 95.52 ± 0.02 & \cellcolor[HTML]{C0C0C0}97.04 ± 0.21 & \cellcolor[HTML]{C0C0C0}95.57 & \cellcolor[HTML]{C0C0C0}82.91 ± 0.66 & 86.07 ± 1.88 & 77.74 ± 0.87 & \cellcolor[HTML]{C0C0C0}94.17 ± 0.36 & \cellcolor[HTML]{C0C0C0}85.22 \\
 & \textbf{Type6} & \cellcolor[HTML]{C0C0C0}93.83 ± 0.18 & 96.47 ± 0.06 & \cellcolor[HTML]{C0C0C0}96.58 ± 0.36 & \cellcolor[HTML]{C0C0C0}95.95 ± 0.25 & \cellcolor[HTML]{C0C0C0}95.71 & \cellcolor[HTML]{C0C0C0}80.88 ± 3.29 & \cellcolor[HTML]{C0C0C0}86.82 ± 1.31 & 72.99 ± 1.24 & 74.56 ± 2.75 & 78.81 \\
 & \textbf{Type7} & \cellcolor[HTML]{C0C0C0}93.26 ± 0.06 & 96.25 ± 0.01 & 95.48 ± 0.04 & 95.34 ± 0.21 & 95.08 & \cellcolor[HTML]{C0C0C0}88.68 ± 3.23 & 80.66 ± 0.85 & 71.95 ± 1.18 & \cellcolor[HTML]{C0C0C0}95.02 ± 0.22 & \cellcolor[HTML]{C0C0C0}84.08 \\
 & \textbf{Type8} & 92.12 ± 0.24 & 95.36 ± 0.13 & 92.45 ± 0.89 & 94.07 ± 0.88 & 93.50 & \cellcolor[HTML]{C0C0C0}83.61 ± 2.66 & 84.23 ± 0.72 & 75.09 ± 2.92 & \cellcolor[HTML]{C0C0C0}93.44 ± 1.73 & \cellcolor[HTML]{C0C0C0}84.10 \\
\multirow{-12}{*}{\textbf{Gossipcop}} & \textbf{Type9} & \cellcolor[HTML]{C0C0C0}93.19 ± 0.05 & 96.47 ± 0.11 & 96.03 ± 0.03 & \cellcolor[HTML]{C0C0C0}95.94 ± 0.13 & \cellcolor[HTML]{C0C0C0}95.41 & \cellcolor[HTML]{C0C0C0}88.83 ± 0.16 & 82.88 ± 3.62 & 77.66 ± 1.01 & \cellcolor[HTML]{C0C0C0}94.76 ± 0.47 & \cellcolor[HTML]{C0C0C0}86.03 \\ \hline
 & \textbf{No Aug} & \multicolumn{1}{c}{77.38 ± 1.81} & \multicolumn{1}{c}{79.55 ± 1.03} & \multicolumn{1}{c}{83.15 ± 0.23} & \multicolumn{1}{c}{84.89 ± 2.51} & 81.24 & 64.02 ± 1.59 & 73.45 ± 0.68 & 79.63 ± 1.91 & 81.02 ± 0.66 & 74.53 \\
 & \textbf{MAXPOOL} & \multicolumn{1}{c}{77.82 ± 1.26} & \multicolumn{1}{c}{79.63 ± 0.79} & \multicolumn{1}{c}{83.49 ± 0.26} & \multicolumn{1}{c}{85.52 ± 1.22} & 81.61 & 66.74 ± 0.32 & 75.11 ± 0.72 & 80.09 ± 1.27 & 82.57 ± 0.95 & 76.13 \\
 & \textbf{GMT} & \multicolumn{1}{c}{\cellcolor[HTML]{C0C0C0}78.74 ± 0.98} & \multicolumn{1}{c}{\cellcolor[HTML]{C0C0C0}81.46 ± 1.68} & \multicolumn{1}{c}{\cellcolor[HTML]{C0C0C0}84.25 ± 0.81} & \multicolumn{1}{c}{81.54 ± 1.28} & 81.50 & \cellcolor[HTML]{C0C0C0}73.53 ± 4.79 & 60.27 ± 1.27 & 79.03 ± 4.61 & 81.23 ± 0.32 & 73.52 \\
 & \textbf{Type1} & \multicolumn{1}{c}{\cellcolor[HTML]{C0C0C0}78.96 ± 1.96} & \multicolumn{1}{c}{\cellcolor[HTML]{C0C0C0}81.61 ± 2.49} & \multicolumn{1}{c}{\cellcolor[HTML]{C0C0C0}83.94 ± 0.45} & \multicolumn{1}{c}{\cellcolor[HTML]{C0C0C0}85.92 ± 1.35} & \cellcolor[HTML]{C0C0C0}{\color[HTML]{FE0000} 82.61} & \cellcolor[HTML]{C0C0C0}69.01 ± 0.32 & 74.21 ± 1.12 & 78.36 ± 2.43 & \cellcolor[HTML]{C0C0C0}84.84 ± 0.31 & \cellcolor[HTML]{C0C0C0}76.60 \\
 & \textbf{Type2} & \multicolumn{1}{c}{\cellcolor[HTML]{C0C0C0}77.83 ± 1.64} & \multicolumn{1}{c}{77.41 ± 2.09} & \multicolumn{1}{c}{\cellcolor[HTML]{C0C0C0}83.72 ± 0.74} & \multicolumn{1}{c}{85.07 ± 1.09} & 81.01 & \cellcolor[HTML]{C0C0C0}68.55 ± 0.96 & 64.93 ± 0.96 & 79.86 ± 0.32 & \cellcolor[HTML]{C0C0C0}82.81 ± 2.56 & 74.04 \\
 & \textbf{Type3} & \multicolumn{1}{c}{\cellcolor[HTML]{C0C0C0}78.73 ± 2.12} & \multicolumn{1}{c}{78.88 ± 3.41} & \multicolumn{1}{c}{83.03 ± 0.58} & \multicolumn{1}{c}{84.62 ± 1.52} & 81.31 & \cellcolor[HTML]{C0C0C0}72.85 ± 0.63 & 71.71 ± 0.95 & \cellcolor[HTML]{C0C0C0}80.54 ± 0.63 & 74.21 ± 1.92 & 74.83 \\
 & \textbf{Type4} & \multicolumn{1}{c}{77.22 ± 0.94} & \multicolumn{1}{c}{\cellcolor[HTML]{C0C0C0}81.22 ± 1.86} & \multicolumn{1}{c}{82.13 ± 3.72} & \multicolumn{1}{c}{83.37 ± 2.11} & 80.99 & \cellcolor[HTML]{C0C0C0}74.43 ± 1.61 & \cellcolor[HTML]{C0C0C0}75.56 ± 1.27 & 77.22 ± 3.79 & 81.67 ± 0.31 & \cellcolor[HTML]{C0C0C0}77.22 \\
 & \textbf{Type5} & 77.38 ± 0.46 & 77.61 ± 1.83 & 82.24 ± 4.21 & 82.81 ± 1.27 & 80.01 & \cellcolor[HTML]{C0C0C0}71.71 ± 2.24 & 72.39 ± 0.64 & 76.41 ± 4.97 & 81.44 ± 1.27 & 75.49 \\
 & \textbf{Type6} & \cellcolor[HTML]{C0C0C0}77.83 ± 1.19 & 79.31 ± 1.79 & \cellcolor[HTML]{C0C0C0}84.62 ± 0.98 & 82.35 ± 1.91 & 81.03 & \cellcolor[HTML]{C0C0C0}75.79 ± 0.96 & 73.75 ± 0.63 & \cellcolor[HTML]{C0C0C0}82.19 ± 1.45 & 75.56 ± 3.21 & \cellcolor[HTML]{C0C0C0}76.82 \\
 & \textbf{Type7} & 77.08 ± 0.53 & \cellcolor[HTML]{C0C0C0}80.66 ± 2.85 & \cellcolor[HTML]{C0C0C0}84.16 ± 0.97 & 83.03 ± 1.59 & 81.23 & \cellcolor[HTML]{C0C0C0}71.49 ± 0.64 & 71.03 ± 1.28 & 79.05 ± 4.44 & 78.28 ± 1.29 & 74.96 \\
 & \textbf{Type8} & \cellcolor[HTML]{C0C0C0}78.17 ± 1.49 & 77.95 ± 0.86 & 82.02 ± 1.97 & 83.57 ± 0.64 & 80.43 & \cellcolor[HTML]{C0C0C0}68.34 ± 0.87 & 66.96 ± 3.83 & 67.64 ± 1.59 & 74.88 ± 1.61 & 69.46 \\
\multirow{-12}{*}{\textbf{Politifact}} & \textbf{Type9} & \cellcolor[HTML]{C0C0C0}78.28 ± 1.36 & \cellcolor[HTML]{C0C0C0}80.91 ± 1.97 & \cellcolor[HTML]{C0C0C0}84.39 ± 0.79 & 84.96 ± 1.29 & \cellcolor[HTML]{C0C0C0}82.14 & \cellcolor[HTML]{C0C0C0}76.91 ± 1.27 & 71.94 ± 1.26 & 77.37 ± 1.92 & \cellcolor[HTML]{C0C0C0}84.16 ± 0.63 & \cellcolor[HTML]{C0C0C0}{\color[HTML]{FE0000} 77.60} \\ \hline
\end{tabular}
}
\end{table*}

\begin{table*}[!tb]
\caption{Effectiveness of nine types of hybrid pooling w.r.t. test accuracy $\uparrow$ (average $\pm$ standard deviation, \%) on original test data and robustness $\uparrow$ (average $\pm$ standard deviation, \%) on robust test data of trained GAT models on NLP datasets. \textbf{No Aug}: without Manifold-Mixup data augmentation. A gray background highlights results that are better than MAXPOOL. The best average results are marked in red color.}
\label{table:Pool_gat}
\centering
\scalebox{0.55}{
\begin{tabular}{llccccc||llllc}
\hline
 &  & \multicolumn{5}{c||}{\textbf{Test Accuracy}} & \multicolumn{5}{c}{\textbf{Robustness}} \\ 
 &  & \textbf{Profile} & \textbf{Spacy} & \textbf{Bert} & \textbf{Content} & \textbf{Average} & \multicolumn{1}{c}{\textbf{Profile}} & \multicolumn{1}{c}{\textbf{Spacy}} & \multicolumn{1}{c}{\textbf{Bert}} & \multicolumn{1}{c}{\textbf{Content}} & \textbf{Average} \\ \hline
 & \textbf{No Aug} & \multicolumn{1}{l}{\cellcolor[HTML]{FFFFFF}{\color[HTML]{000000} 91.19 ± 0.03}} & \multicolumn{1}{l}{\cellcolor[HTML]{FFFFFF}{\color[HTML]{000000} 95.54 ± 0.02}} & \multicolumn{1}{l}{\cellcolor[HTML]{FFFFFF}{\color[HTML]{000000} 96.49 ± 0.28}} & \multicolumn{1}{l}{\cellcolor[HTML]{FFFFFF}{\color[HTML]{000000} 97.28 ± 0.43}} & \cellcolor[HTML]{FFFFFF}{\color[HTML]{000000} 95.13} & \cellcolor[HTML]{FFFFFF}{\color[HTML]{000000} 90.69 ± 0.21} & \cellcolor[HTML]{FFFFFF}{\color[HTML]{000000} 94.96 ± 0.06} & \cellcolor[HTML]{FFFFFF}{\color[HTML]{000000} 94.49 ± 0.51} & \cellcolor[HTML]{FFFFFF}{\color[HTML]{000000} 96.23 ± 0.63} & \cellcolor[HTML]{FFFFFF}{\color[HTML]{000000} 94.09} \\
 & \textbf{MAXPOOL} & \multicolumn{1}{l}{\cellcolor[HTML]{FFFFFF}{\color[HTML]{000000} 93.87 ± 0.47}} & \multicolumn{1}{l}{\cellcolor[HTML]{FFFFFF}{\color[HTML]{000000} 96.00 ± 0.18}} & \multicolumn{1}{l}{\cellcolor[HTML]{FFFFFF}{\color[HTML]{000000} 96.74 ± 0.41}} & \multicolumn{1}{l}{\cellcolor[HTML]{FFFFFF}{\color[HTML]{000000} 97.39 ± 0.24}} & \cellcolor[HTML]{FFFFFF}{\color[HTML]{000000} 96.00} & \cellcolor[HTML]{FFFFFF}{\color[HTML]{000000} 90.81 ± 0.52} & \cellcolor[HTML]{FFFFFF}{\color[HTML]{000000} 95.22 ± 0.05} & \cellcolor[HTML]{FFFFFF}{\color[HTML]{000000} 94.53 ± 1.77} & \cellcolor[HTML]{FFFFFF}{\color[HTML]{000000} 96.31 ± 0.89} & \cellcolor[HTML]{FFFFFF}{\color[HTML]{000000} 94.22} \\
 & \textbf{GMT} & \multicolumn{1}{l}{\cellcolor[HTML]{FFFFFF}{\color[HTML]{000000} 92.38 ± 0.17}} & \multicolumn{1}{l}{\cellcolor[HTML]{FFFFFF}{\color[HTML]{000000} 95.96 ± 0.01}} & \multicolumn{1}{l}{\cellcolor[HTML]{FFFFFF}{\color[HTML]{000000} 95.93 ± 0.39}} & \multicolumn{1}{l}{\cellcolor[HTML]{C0C0C0}{\color[HTML]{000000} 97.46 ± 0.08}} & \cellcolor[HTML]{FFFFFF}{\color[HTML]{000000} 95.43} & \cellcolor[HTML]{FFFFFF}{\color[HTML]{000000} 86.33 ± 2.73} & \cellcolor[HTML]{FFFFFF}{\color[HTML]{000000} 77.44 ± 2.01} & \cellcolor[HTML]{FFFFFF}{\color[HTML]{000000} 79.04 ± 2.06} & \cellcolor[HTML]{FFFFFF}{\color[HTML]{000000} 90.78 ± 3.81} & \cellcolor[HTML]{FFFFFF}{\color[HTML]{000000} 83.40} \\
 & \textbf{Type1} & \multicolumn{1}{l}{\cellcolor[HTML]{C0C0C0}{\color[HTML]{000000} 94.68 ± 0.01}} & \multicolumn{1}{l}{\cellcolor[HTML]{C0C0C0}{\color[HTML]{000000} 96.56 ± 0.31}} & \multicolumn{1}{l}{\cellcolor[HTML]{C0C0C0}{\color[HTML]{000000} 97.27 ± 0.21}} & \multicolumn{1}{l}{\cellcolor[HTML]{C0C0C0}{\color[HTML]{000000} 97.54 ± 0.17}} & \cellcolor[HTML]{C0C0C0}{\color[HTML]{000000} 96.51} & \cellcolor[HTML]{C0C0C0}{\color[HTML]{000000} 91.89 ± 0.79} & \cellcolor[HTML]{C0C0C0}{\color[HTML]{000000} 95.71 ± 0.59} & \cellcolor[HTML]{C0C0C0}{\color[HTML]{000000} 95.01 ± 2.35} & \cellcolor[HTML]{C0C0C0}{\color[HTML]{000000} 96.82 ± 0.63} & \cellcolor[HTML]{C0C0C0}{\color[HTML]{FE0000} 94.86} \\
 & \textbf{Type2} & \multicolumn{1}{l}{\cellcolor[HTML]{C0C0C0}{\color[HTML]{000000} 94.76 ± 0.54}} & \multicolumn{1}{l}{\cellcolor[HTML]{C0C0C0}{\color[HTML]{000000} 96.15 ± 0.43}} & \multicolumn{1}{l}{\cellcolor[HTML]{C0C0C0}{\color[HTML]{000000} 96.99 ± 0.18}} & \multicolumn{1}{l}{\cellcolor[HTML]{C0C0C0}{\color[HTML]{000000} 97.41 ± 0.09}} & \cellcolor[HTML]{C0C0C0}{\color[HTML]{000000} 96.33} & \cellcolor[HTML]{C0C0C0}{\color[HTML]{000000} 91.02 ± 0.52} & \cellcolor[HTML]{FFFFFF}{\color[HTML]{000000} 94.44 ± 1.21} & \cellcolor[HTML]{C0C0C0}{\color[HTML]{000000} 94.95 ± 1.92} & \cellcolor[HTML]{C0C0C0}{\color[HTML]{000000} 96.80 ± 0.49} & \cellcolor[HTML]{C0C0C0}{\color[HTML]{000000} 94.30} \\
 & \textbf{Type3} & \multicolumn{1}{l}{\cellcolor[HTML]{C0C0C0}{\color[HTML]{000000} 94.81 ± 0.13}} & \multicolumn{1}{l}{\cellcolor[HTML]{C0C0C0}{\color[HTML]{000000} 97.08 ± 0.26}} & \multicolumn{1}{l}{\cellcolor[HTML]{C0C0C0}{\color[HTML]{000000} 97.22 ± 0.16}} & \multicolumn{1}{l}{\cellcolor[HTML]{C0C0C0}{\color[HTML]{000000} 97.53 ± 0.37}} & \cellcolor[HTML]{C0C0C0}{\color[HTML]{FE0000} 96.66} & \cellcolor[HTML]{C0C0C0}{\color[HTML]{000000} 92.03 ± 0.76} & \cellcolor[HTML]{C0C0C0}{\color[HTML]{000000} 96.46 ± 0.02} & \cellcolor[HTML]{FFFFFF}{\color[HTML]{000000} 93.11 ± 0.91} & \cellcolor[HTML]{FFFFFF}{\color[HTML]{000000} 96.22 ± 0.57} & \cellcolor[HTML]{C0C0C0}{\color[HTML]{000000} 94.46} \\
 & \textbf{Type4} & \multicolumn{1}{l}{\cellcolor[HTML]{FFFFFF}{\color[HTML]{000000} 92.57 ± 0.21}} & \multicolumn{1}{l}{\cellcolor[HTML]{FFFFFF}{\color[HTML]{000000} 95.63 ± 0.02}} & \multicolumn{1}{l}{\cellcolor[HTML]{FFFFFF}{\color[HTML]{000000} 94.96 ± 0.41}} & \multicolumn{1}{l}{\cellcolor[HTML]{FFFFFF}{\color[HTML]{000000} 96.71 ± 0.08}} & \cellcolor[HTML]{FFFFFF}{\color[HTML]{000000} 94.97} & \cellcolor[HTML]{FFFFFF}{\color[HTML]{000000} 90.68 ± 0.31} & \cellcolor[HTML]{FFFFFF}{\color[HTML]{000000} 94.48 ± 1.03} & \cellcolor[HTML]{FFFFFF}{\color[HTML]{000000} 93.87 ± 0.83} & \cellcolor[HTML]{FFFFFF}{\color[HTML]{000000} 95.53 ± 0.07} & \cellcolor[HTML]{FFFFFF}{\color[HTML]{000000} 93.64} \\
 & \textbf{Type5} & \multicolumn{1}{l}{\cellcolor[HTML]{FFFFFF}{\color[HTML]{000000} 92.81 ± 0.59}} & \multicolumn{1}{l}{\cellcolor[HTML]{FFFFFF}{\color[HTML]{000000} 95.98 ± 0.08}} & \multicolumn{1}{l}{\cellcolor[HTML]{FFFFFF}{\color[HTML]{000000} 95.42 ± 0.32}} & \multicolumn{1}{l}{\cellcolor[HTML]{FFFFFF}{\color[HTML]{000000} 97.05 ± 0.31}} & \cellcolor[HTML]{FFFFFF}{\color[HTML]{000000} 95.32} & \cellcolor[HTML]{FFFFFF}{\color[HTML]{000000} 90.44 ± 0.46} & \cellcolor[HTML]{FFFFFF}{\color[HTML]{000000} 94.13 ± 0.06} & \cellcolor[HTML]{FFFFFF}{\color[HTML]{000000} 88.23 ± 1.36} & \cellcolor[HTML]{FFFFFF}{\color[HTML]{000000} 95.82 ± 0.74} & \cellcolor[HTML]{FFFFFF}{\color[HTML]{000000} 92.16} \\
 & \textbf{Type6} & \multicolumn{1}{l}{\cellcolor[HTML]{C0C0C0}{\color[HTML]{000000} 95.18 ± 0.02}} & \multicolumn{1}{l}{\cellcolor[HTML]{FFFFFF}{\color[HTML]{000000} 95.94 ± 0.69}} & \multicolumn{1}{l}{\cellcolor[HTML]{FFFFFF}{\color[HTML]{000000} 96.63 ± 0.81}} & \multicolumn{1}{l}{\cellcolor[HTML]{FFFFFF}{\color[HTML]{000000} 96.97 ± 0.07}} & \cellcolor[HTML]{C0C0C0}{\color[HTML]{000000} 96.18} & \cellcolor[HTML]{C0C0C0}{\color[HTML]{000000} 91.85 ± 018} & \cellcolor[HTML]{FFFFFF}{\color[HTML]{000000} 94.73 ± 0.26} & \cellcolor[HTML]{FFFFFF}{\color[HTML]{000000} 92.75 ± 1.16} & \cellcolor[HTML]{FFFFFF}{\color[HTML]{000000} 96.11 ± 1.19} & \cellcolor[HTML]{FFFFFF}{\color[HTML]{000000} 93.86} \\
 & \textbf{Type7} & \multicolumn{1}{l}{\cellcolor[HTML]{FFFFFF}{\color[HTML]{000000} 92.24 ± 0.07}} & \multicolumn{1}{l}{\cellcolor[HTML]{FFFFFF}{\color[HTML]{000000} 95.23 ± 0.21}} & \multicolumn{1}{l}{\cellcolor[HTML]{FFFFFF}{\color[HTML]{000000} 94.59 ± 0.04}} & \multicolumn{1}{l}{\cellcolor[HTML]{FFFFFF}{\color[HTML]{000000} 96.66 ± 0.11}} & \cellcolor[HTML]{FFFFFF}{\color[HTML]{000000} 94.68} & \cellcolor[HTML]{FFFFFF}{\color[HTML]{000000} 90.48 ± 0.26} & \cellcolor[HTML]{FFFFFF}{\color[HTML]{000000} 94.07 ± 0.18} & \cellcolor[HTML]{FFFFFF}{\color[HTML]{000000} 93.73 ± 0.95} & \cellcolor[HTML]{FFFFFF}{\color[HTML]{000000} 94.83 ± 0.42} & \cellcolor[HTML]{FFFFFF}{\color[HTML]{000000} 93.28} \\
 & \textbf{Type8} & \multicolumn{1}{l}{\cellcolor[HTML]{FFFFFF}{\color[HTML]{000000} 91.92 ± 0.39}} & \multicolumn{1}{l}{\cellcolor[HTML]{FFFFFF}{\color[HTML]{000000} 95.36 ± 0.47}} & \multicolumn{1}{l}{\cellcolor[HTML]{FFFFFF}{\color[HTML]{000000} 95.46 ± 0.71}} & \multicolumn{1}{l}{\cellcolor[HTML]{FFFFFF}{\color[HTML]{000000} 96.39 ± 0.48}} & \cellcolor[HTML]{FFFFFF}{\color[HTML]{000000} 94.78} & \cellcolor[HTML]{FFFFFF}{\color[HTML]{000000} 90.24 ± 0.15} & \cellcolor[HTML]{FFFFFF}{\color[HTML]{000000} 94.89 ± 0.41} & \cellcolor[HTML]{FFFFFF}{\color[HTML]{000000} 94.41 ± 0.56} & \cellcolor[HTML]{FFFFFF}{\color[HTML]{000000} 95.87 ± 0.52} & \cellcolor[HTML]{FFFFFF}{\color[HTML]{000000} 93.85} \\
\multirow{-12}{*}{\textbf{Gossipcop}} & \textbf{Type9} & \multicolumn{1}{l}{\cellcolor[HTML]{FFFFFF}{\color[HTML]{000000} 92.41 ± 0.09}} & \multicolumn{1}{l}{\cellcolor[HTML]{FFFFFF}{\color[HTML]{000000} 95.78 ± 0.31}} & \multicolumn{1}{l}{\cellcolor[HTML]{FFFFFF}{\color[HTML]{000000} 96.04 ± 0.31}} & \multicolumn{1}{l}{\cellcolor[HTML]{FFFFFF}{\color[HTML]{000000} 97.15 ± 0.04}} & \cellcolor[HTML]{FFFFFF}{\color[HTML]{000000} 95.35} & \cellcolor[HTML]{FFFFFF}{\color[HTML]{000000} 90.59 ± 0.11} & \cellcolor[HTML]{C0C0C0}{\color[HTML]{000000} 95.25 ± 0.19} & \cellcolor[HTML]{FFFFFF}{\color[HTML]{000000} 93.98 ± 2.14} & \cellcolor[HTML]{C0C0C0}{\color[HTML]{000000} 96.54 ± 0.43} & \cellcolor[HTML]{FFFFFF}{\color[HTML]{000000} 94.09} \\ \hline
 & \textbf{No Aug} & \cellcolor[HTML]{FFFFFF}{\color[HTML]{000000} 74.84 ± 1.38} & \cellcolor[HTML]{FFFFFF}{\color[HTML]{000000} 79.19 ± 0.46} & \cellcolor[HTML]{FFFFFF}{\color[HTML]{000000} 81.61 ± 2.88} & \cellcolor[HTML]{FFFFFF}{\color[HTML]{000000} 85.29 ± 0.58} & \cellcolor[HTML]{FFFFFF}{\color[HTML]{000000} 80.23} & \cellcolor[HTML]{FFFFFF}{\color[HTML]{000000} 74.43 ± 0.32} & \cellcolor[HTML]{FFFFFF}{\color[HTML]{000000} 77.01 ± 1.15} & \cellcolor[HTML]{FFFFFF}{\color[HTML]{000000} 78.28 ± 1.92} & \cellcolor[HTML]{FFFFFF}{\color[HTML]{000000} 84.16 ± 0.64} & \cellcolor[HTML]{FFFFFF}{\color[HTML]{000000} 78.47} \\
 & \textbf{MAXPOOL} & \cellcolor[HTML]{FFFFFF}{\color[HTML]{000000} 75.57 ± 1.11} & \cellcolor[HTML]{FFFFFF}{\color[HTML]{000000} 79.41 ± 0.96} & \cellcolor[HTML]{FFFFFF}{\color[HTML]{000000} 82.58 ± 0.33} & \cellcolor[HTML]{FFFFFF}{\color[HTML]{000000} 85.67 ± 2.14} & \cellcolor[HTML]{FFFFFF}{\color[HTML]{000000} 80.81} & \cellcolor[HTML]{FFFFFF}{\color[HTML]{000000} 75.26 ± 0.69} & \cellcolor[HTML]{FFFFFF}{\color[HTML]{000000} 77.61 ± 0.77} & \cellcolor[HTML]{FFFFFF}{\color[HTML]{000000} 79.54 ± 1.59} & \cellcolor[HTML]{FFFFFF}{\color[HTML]{000000} 84.62 ± 012} & \cellcolor[HTML]{FFFFFF}{\color[HTML]{000000} 79.26} \\
 & \textbf{GMT} & \cellcolor[HTML]{C0C0C0}{\color[HTML]{000000} 76.93 ± 0.46} & \cellcolor[HTML]{C0C0C0}{\color[HTML]{000000} 80.32 ± 0.32} & \cellcolor[HTML]{FFFFFF}{\color[HTML]{000000} 80.54 ± 1.22} & \cellcolor[HTML]{FFFFFF}{\color[HTML]{000000} 83.71 ± 0.63} & \cellcolor[HTML]{FFFFFF}{\color[HTML]{000000} 80.38} & \cellcolor[HTML]{FFFFFF}{\color[HTML]{000000} 74.88 ± 2.87} & \cellcolor[HTML]{FFFFFF}{\color[HTML]{000000} 71.72 ± 0.32} & \cellcolor[HTML]{FFFFFF}{\color[HTML]{000000} 78.43 ± 1.82} & \cellcolor[HTML]{FFFFFF}{\color[HTML]{000000} 81.14 ± 1.59} & \cellcolor[HTML]{FFFFFF}{\color[HTML]{000000} 76.54} \\
 & \textbf{Type1} & \cellcolor[HTML]{FFFFFF}{\color[HTML]{000000} 75.11 ± 0.66} & \cellcolor[HTML]{C0C0C0}{\color[HTML]{000000} 80.41 ± 1.05} & \cellcolor[HTML]{C0C0C0}{\color[HTML]{000000} 83.04 ± 0.31} & \cellcolor[HTML]{C0C0C0}{\color[HTML]{000000} 86.88 ± 0.64} & \cellcolor[HTML]{C0C0C0}{\color[HTML]{000000} 81.36} & \cellcolor[HTML]{FFFFFF}{\color[HTML]{000000} 74.45 ± 0.27} & \cellcolor[HTML]{C0C0C0}{\color[HTML]{000000} 79.05 ± 1.21} & \cellcolor[HTML]{C0C0C0}{\color[HTML]{000000} 81.89 ± 0.64} & \cellcolor[HTML]{C0C0C0}{\color[HTML]{000000} 85.06 ± 1.28} & \cellcolor[HTML]{C0C0C0}{\color[HTML]{000000} {\color[HTML]{FE0000} 80.11}} \\
 & \textbf{Type2} & \cellcolor[HTML]{C0C0C0}{\color[HTML]{000000} 76.17 ± 0.69} & \cellcolor[HTML]{C0C0C0}{\color[HTML]{000000} 79.86 ± 0.32} & \cellcolor[HTML]{FFFFFF}{\color[HTML]{000000} 82.35 ± 0.95} & \cellcolor[HTML]{FFFFFF}{\color[HTML]{000000} 85.52 ± 0.78} & \cellcolor[HTML]{C0C0C0}{\color[HTML]{000000} 80.98} & \cellcolor[HTML]{FFFFFF}{\color[HTML]{000000} 74.66 ± 0.63} & \cellcolor[HTML]{C0C0C0}{\color[HTML]{000000} 78.73 ± 1.07} & \cellcolor[HTML]{FFFFFF}{\color[HTML]{000000} 74.21 ± 0.61} & \cellcolor[HTML]{FFFFFF}{\color[HTML]{000000} 82.57 ± 0.31} & \cellcolor[HTML]{FFFFFF}{\color[HTML]{000000} 77.54} \\
 & \textbf{Type3} & \cellcolor[HTML]{C0C0C0}{\color[HTML]{000000} 78.73 ± 1.19} & \cellcolor[HTML]{FFFFFF}{\color[HTML]{000000} 78.96 ± 0.33} & \cellcolor[HTML]{FFFFFF}{\color[HTML]{000000} 82.21 ± 0.39} & \cellcolor[HTML]{C0C0C0}{\color[HTML]{000000} 85.75 ± 0.98} & \cellcolor[HTML]{C0C0C0}{\color[HTML]{FE0000} 81.41} & \cellcolor[HTML]{C0C0C0}{\color[HTML]{000000} 76.02 ± 0.63} & \cellcolor[HTML]{FFFFFF}{\color[HTML]{000000} 76.78 ± 0.19} & \cellcolor[HTML]{FFFFFF}{\color[HTML]{000000} 78.27 ± 0.57} & \cellcolor[HTML]{FFFFFF}{\color[HTML]{000000} 81.67 ± 0.32} & \cellcolor[HTML]{FFFFFF}{\color[HTML]{000000} 78.19} \\
 & \textbf{Type4} & \cellcolor[HTML]{C0C0C0}{\color[HTML]{000000} 76.62 ± 2.03} & \cellcolor[HTML]{FFFFFF}{\color[HTML]{000000} 79.04 ± 0.45} & \cellcolor[HTML]{FFFFFF}{\color[HTML]{000000} 80.09 ± 0.22} & \cellcolor[HTML]{FFFFFF}{\color[HTML]{000000} 83.26 ± 1.28} & \cellcolor[HTML]{FFFFFF}{\color[HTML]{000000} 79.75} & \cellcolor[HTML]{FFFFFF}{\color[HTML]{000000} 74.43 ± 1.24} & \cellcolor[HTML]{FFFFFF}{\color[HTML]{000000} 77.59 ± 0.31} & \cellcolor[HTML]{FFFFFF}{\color[HTML]{000000} 78.05 ± 0.96} & \cellcolor[HTML]{FFFFFF}{\color[HTML]{000000} 75.34 ± 0.28} & \cellcolor[HTML]{FFFFFF}{\color[HTML]{000000} 76.35} \\
 & \textbf{Type5} & \cellcolor[HTML]{C0C0C0}{\color[HTML]{000000} 76.61 ± 1.38} & \cellcolor[HTML]{FFFFFF}{\color[HTML]{000000} 78.73 ± 0.53} & \cellcolor[HTML]{FFFFFF}{\color[HTML]{000000} 79.76 ± 0.31} & \cellcolor[HTML]{FFFFFF}{\color[HTML]{000000} 83.67 ± 1.45} & \cellcolor[HTML]{FFFFFF}{\color[HTML]{000000} 79.69} & \cellcolor[HTML]{C0C0C0}{\color[HTML]{000000} 75.46 ± 0.65} & \cellcolor[HTML]{FFFFFF}{\color[HTML]{000000} 75.56 ± 1.92} & \cellcolor[HTML]{FFFFFF}{\color[HTML]{000000} 75.29 ± 1.67} & \cellcolor[HTML]{FFFFFF}{\color[HTML]{000000} 78.95 ± 2.15} & \cellcolor[HTML]{FFFFFF}{\color[HTML]{000000} 76.32} \\
 & \textbf{Type6} & \cellcolor[HTML]{FFFFFF}{\color[HTML]{000000} 75.27 ± 0.94} & \cellcolor[HTML]{FFFFFF}{\color[HTML]{000000} 78.96 ± 0.32} & \cellcolor[HTML]{FFFFFF}{\color[HTML]{000000} 79.19 ± 0.29} & \cellcolor[HTML]{FFFFFF}{\color[HTML]{000000} 83.81 ± 1.12} & \cellcolor[HTML]{FFFFFF}{\color[HTML]{000000} 79.31} & \cellcolor[HTML]{FFFFFF}{\color[HTML]{000000} 72.95 ± 1.26} & \cellcolor[HTML]{FFFFFF}{\color[HTML]{000000} 76.02 ± 3.19} & \cellcolor[HTML]{FFFFFF}{\color[HTML]{000000} 74.88 ± 2.87} & \cellcolor[HTML]{FFFFFF}{\color[HTML]{000000} 81.21 ± 2.24} & \cellcolor[HTML]{FFFFFF}{\color[HTML]{000000} 76.27} \\
 & \textbf{Type7} & \cellcolor[HTML]{C0C0C0}{\color[HTML]{000000} 76.69 ± 1.95} & \cellcolor[HTML]{C0C0C0}{\color[HTML]{000000} 80.32 ± 0.96} & \cellcolor[HTML]{FFFFFF}{\color[HTML]{000000} 81.99 ± 0.11} & \cellcolor[HTML]{C0C0C0}{\color[HTML]{000000} 86.43 ± 0.77} & \cellcolor[HTML]{C0C0C0}{\color[HTML]{000000} 81.36} & \cellcolor[HTML]{FFFFFF}{\color[HTML]{000000} 75.33 ± 0.31} & \cellcolor[HTML]{FFFFFF}{\color[HTML]{000000} 75.56 ± 3.83} & \cellcolor[HTML]{C0C0C0}{\color[HTML]{000000} 79.64 ± 1.28} & \cellcolor[HTML]{FFFFFF}{\color[HTML]{000000} 71.72 ± 0.33} & \cellcolor[HTML]{FFFFFF}{\color[HTML]{000000} 75.56} \\
 & \textbf{Type8} & \cellcolor[HTML]{FFFFFF}{\color[HTML]{000000} 74.78 ± 2.77} & \cellcolor[HTML]{FFFFFF}{\color[HTML]{000000} 79.19 ± 0.64} & \cellcolor[HTML]{FFFFFF}{\color[HTML]{000000} 81.85 ± 0.21} & \cellcolor[HTML]{FFFFFF}{\color[HTML]{000000} 85.35 ± 0.65} & \cellcolor[HTML]{FFFFFF}{\color[HTML]{000000} 80.29} & \cellcolor[HTML]{FFFFFF}{\color[HTML]{000000} 72.62 ± 1.23} & \cellcolor[HTML]{FFFFFF}{\color[HTML]{000000} 75.06 ± 3.13} & \cellcolor[HTML]{FFFFFF}{\color[HTML]{000000} 76.02 ± 2.55} & \cellcolor[HTML]{FFFFFF}{\color[HTML]{000000} 78.95 ± 0.95} & \cellcolor[HTML]{FFFFFF}{\color[HTML]{000000} 75.66} \\
\multirow{-12}{*}{\textbf{Politifact}} & \textbf{Type9} & \cellcolor[HTML]{FFFFFF}{\color[HTML]{000000} 75.08 ± 2.29} & \cellcolor[HTML]{C0C0C0}{\color[HTML]{000000} 80.29 ± 1.08} & \cellcolor[HTML]{FFFFFF}{\color[HTML]{000000} 81.02 ± 0.32} & \cellcolor[HTML]{C0C0C0}{\color[HTML]{000000} 85.92 ± 0.81} & \cellcolor[HTML]{FFFFFF}{\color[HTML]{000000} 80.58} & \cellcolor[HTML]{FFFFFF}{\color[HTML]{000000} 74.45 ± 0.38} & \cellcolor[HTML]{FFFFFF}{\color[HTML]{000000} 76.26 ± 3.75} & \cellcolor[HTML]{C0C0C0}{\color[HTML]{000000} 79.63 ± 1.27} & \cellcolor[HTML]{FFFFFF}{\color[HTML]{000000} 81.22 ± 2.87} & \cellcolor[HTML]{FFFFFF}{\color[HTML]{000000} 77.89} \\ \hline
\end{tabular}
}
\end{table*}

\begin{table*}[!tb]
\caption{Effectiveness of nine types of hybrid pooling w.r.t. test accuracy $\uparrow$ (average $\pm$ standard deviation, \%) on original test data and robustness $\uparrow$ (average $\pm$ standard deviation, \%) on robust test data of trained GraphSAGE models on NLP datasets. \textbf{No Aug}: without Manifold-Mixup data augmentation. A gray background highlights results that are better than MAXPOOL. The best average results are marked in red color.}
\label{table:Pool_sage}
\centering
\scalebox{0.55}{
\begin{tabular}{llllllc||llllc}
\hline
 &  & \multicolumn{5}{c||}{\textbf{Test Accuracy}} & \multicolumn{5}{c}{\textbf{Robustness}} \\ 
 &  & \multicolumn{1}{c}{\textbf{Profile}} & \multicolumn{1}{c}{\textbf{Spacy}} & \multicolumn{1}{c}{\textbf{Bert}} & \multicolumn{1}{c}{\textbf{Content}} & \textbf{Average} & \multicolumn{1}{c}{\textbf{Profile}} & \multicolumn{1}{c}{\textbf{Spacy}} & \multicolumn{1}{c}{\textbf{Bert}} & \multicolumn{1}{c}{\textbf{Content}} & \textbf{Average} \\ \hline
 & \textbf{No Aug} & \cellcolor[HTML]{FFFFFF}{\color[HTML]{000000} 92.35 ± 0.02} & \cellcolor[HTML]{FFFFFF}{\color[HTML]{000000} 96.53 ± 0.11} & \cellcolor[HTML]{FFFFFF}{\color[HTML]{000000} 96.62 ± 0.02} & \cellcolor[HTML]{FFFFFF}{\color[HTML]{000000} 97.58 ± 0.06} & \cellcolor[HTML]{FFFFFF}{\color[HTML]{000000} 95.77} & \cellcolor[HTML]{FFFFFF}{\color[HTML]{000000} 90.17 ± 0.37} & \cellcolor[HTML]{FFFFFF}{\color[HTML]{000000} 94.68 ± 0.11} & \cellcolor[HTML]{FFFFFF}{\color[HTML]{000000} 95.61 ± 0.32} & \cellcolor[HTML]{FFFFFF}{\color[HTML]{000000} 96.34 ± 0.28} & \cellcolor[HTML]{FFFFFF}{\color[HTML]{000000} 94.20} \\
 & \textbf{MAXPOOL} & \cellcolor[HTML]{FFFFFF}{\color[HTML]{000000} 93.65 ± 0.33} & \cellcolor[HTML]{FFFFFF}{\color[HTML]{000000} 96.79 ± 0.09} & \cellcolor[HTML]{FFFFFF}{\color[HTML]{000000} 96.81 ± 0.01} & \cellcolor[HTML]{FFFFFF}{\color[HTML]{000000} 97.78 ± 0.02} & \cellcolor[HTML]{FFFFFF}{\color[HTML]{000000} 96.26} & \cellcolor[HTML]{FFFFFF}{\color[HTML]{000000} 90.72 ± 0.48} & \cellcolor[HTML]{FFFFFF}{\color[HTML]{000000} 95.01 ± 0.02} & \cellcolor[HTML]{FFFFFF}{\color[HTML]{000000} 95.84 ± 0.14} & \cellcolor[HTML]{FFFFFF}{\color[HTML]{000000} 96.99 ± 0.03} & \cellcolor[HTML]{FFFFFF}{\color[HTML]{000000} 94.64} \\
 & \textbf{GMT} & \cellcolor[HTML]{FFFFFF}{\color[HTML]{000000} 93.31 ± 0.02} & \cellcolor[HTML]{FFFFFF}{\color[HTML]{000000} 94.61 ± 0.02} & \cellcolor[HTML]{FFFFFF}{\color[HTML]{000000} 94.63 ± 0.09} & \cellcolor[HTML]{FFFFFF}{\color[HTML]{000000} 96.71 ± 0.26} & \cellcolor[HTML]{FFFFFF}{\color[HTML]{000000} 94.82} & \cellcolor[HTML]{FFFFFF}{\color[HTML]{000000} 82.18 ± 2.74} & \cellcolor[HTML]{FFFFFF}{\color[HTML]{000000} 84.72 ± 1.93} & \cellcolor[HTML]{FFFFFF}{\color[HTML]{000000} 85.19 ± 2.24} & \cellcolor[HTML]{FFFFFF}{\color[HTML]{000000} 93.31 ± 2.05} & \cellcolor[HTML]{FFFFFF}{\color[HTML]{000000} 86.35} \\
 & \textbf{Type1} & \cellcolor[HTML]{C0C0C0}{\color[HTML]{000000} 94.41 ± 0.11} & \cellcolor[HTML]{FFFFFF}{\color[HTML]{000000} 96.63 ± 0.15} & \cellcolor[HTML]{C0C0C0}{\color[HTML]{000000} 96.88 ± 0.03} & \cellcolor[HTML]{C0C0C0}{\color[HTML]{000000} 97.91 ± 0.03} & \cellcolor[HTML]{C0C0C0}{\color[HTML]{000000} 96.46} & \cellcolor[HTML]{C0C0C0}{\color[HTML]{000000} 94.05 ± 0.49} & \cellcolor[HTML]{C0C0C0}{\color[HTML]{000000} 95.15 ± 0.05} & \cellcolor[HTML]{FFFFFF}{\color[HTML]{000000} 88.89 ± 2.22} & \cellcolor[HTML]{C0C0C0}{\color[HTML]{000000} 97.69 ± 0.09} & \cellcolor[HTML]{FFFFFF}{\color[HTML]{000000} 93.95} \\
 & \textbf{Type2} & \cellcolor[HTML]{C0C0C0}{\color[HTML]{000000} 94.13 ± 0.12} & \cellcolor[HTML]{FFFFFF}{\color[HTML]{000000} 96.48 ± 0.28} & \cellcolor[HTML]{FFFFFF}{\color[HTML]{000000} 96.47 ± 0.18} & \cellcolor[HTML]{C0C0C0}{\color[HTML]{000000} 97.81 ± 0.04} & \cellcolor[HTML]{FFFFFF}{\color[HTML]{000000} 96.22} & \cellcolor[HTML]{C0C0C0}{\color[HTML]{000000} 92.27 ± 0.83} & \cellcolor[HTML]{FFFFFF}{\color[HTML]{000000} 90.08 ± 1.91} & \cellcolor[HTML]{FFFFFF}{\color[HTML]{000000} 91.85 ± 3.39} & \cellcolor[HTML]{C0C0C0}{\color[HTML]{000000} 97.26 ± 0.13} & \cellcolor[HTML]{FFFFFF}{\color[HTML]{000000} 92.87} \\
 & \textbf{Type3} & \cellcolor[HTML]{C0C0C0}{\color[HTML]{000000} 94.81 ± 0.24} & \cellcolor[HTML]{C0C0C0}{\color[HTML]{000000} 96.92 ± 0.07} & \cellcolor[HTML]{FFFFFF}{\color[HTML]{000000} 96.98 ± 0.24} & \cellcolor[HTML]{FFFFFF}{\color[HTML]{000000} 97.71 ± 0.09} & \cellcolor[HTML]{C0C0C0}{\color[HTML]{FE0000} 96.61} & \cellcolor[HTML]{C0C0C0}{\color[HTML]{000000} 94.24 ± 0.68} & \cellcolor[HTML]{C0C0C0}{\color[HTML]{000000} 95.91 ± 0.09} & \cellcolor[HTML]{C0C0C0}{\color[HTML]{000000} 95.94 ± 0.16} & \cellcolor[HTML]{C0C0C0}{\color[HTML]{000000} 97.47 ± 0.05} & \cellcolor[HTML]{C0C0C0}{\color[HTML]{FE0000} 95.89} \\
 & \textbf{Type4} & \cellcolor[HTML]{FFFFFF}{\color[HTML]{000000} 93.57 ± 0.55} & \cellcolor[HTML]{FFFFFF}{\color[HTML]{000000} 94.69 ± 0.11} & \cellcolor[HTML]{FFFFFF}{\color[HTML]{000000} 95.47 ± 0.68} & \cellcolor[HTML]{FFFFFF}{\color[HTML]{000000} 96.66 ± 0.14} & \cellcolor[HTML]{FFFFFF}{\color[HTML]{000000} 95.10} & \cellcolor[HTML]{C0C0C0}{\color[HTML]{000000} 92.91 ± 0.51} & \cellcolor[HTML]{FFFFFF}{\color[HTML]{000000} 94.23 ± 0.05} & \cellcolor[HTML]{FFFFFF}{\color[HTML]{000000} 91.07 ± 1.08} & \cellcolor[HTML]{FFFFFF}{\color[HTML]{000000} 92.57 ± 0.07} & \cellcolor[HTML]{FFFFFF}{\color[HTML]{000000} 92.70} \\
 & \textbf{Type5} & \cellcolor[HTML]{FFFFFF}{\color[HTML]{000000} 92.88 ± 0.17} & \cellcolor[HTML]{FFFFFF}{\color[HTML]{000000} 94.68 ± 0.49} & \cellcolor[HTML]{FFFFFF}{\color[HTML]{000000} 95.36 ± 0.46} & \cellcolor[HTML]{FFFFFF}{\color[HTML]{000000} 97.44 ± 0.22} & \cellcolor[HTML]{FFFFFF}{\color[HTML]{000000} 95.09} & \cellcolor[HTML]{C0C0C0}{\color[HTML]{000000} 92.18 ± 0.21} & \cellcolor[HTML]{FFFFFF}{\color[HTML]{000000} 92.67 ± 2.19} & \cellcolor[HTML]{FFFFFF}{\color[HTML]{000000} 93.78 ± 0.84} & \cellcolor[HTML]{FFFFFF}{\color[HTML]{000000} 96.33 ± 0.92} & \cellcolor[HTML]{FFFFFF}{\color[HTML]{000000} 93.74} \\
 & \textbf{Type6} & \cellcolor[HTML]{C0C0C0}{\color[HTML]{000000} 94.38 ± 0.23} & \cellcolor[HTML]{FFFFFF}{\color[HTML]{000000} 95.32 ± 0.41} & \cellcolor[HTML]{FFFFFF}{\color[HTML]{000000} 95.91 ± 0.06} & \cellcolor[HTML]{FFFFFF}{\color[HTML]{000000} 97.73 ± 0.12} & \cellcolor[HTML]{FFFFFF}{\color[HTML]{000000} 95.84} & \cellcolor[HTML]{C0C0C0}{\color[HTML]{000000} 93.65 ± 0.33} & \cellcolor[HTML]{FFFFFF}{\color[HTML]{000000} 94.17 ± 0.34} & \cellcolor[HTML]{FFFFFF}{\color[HTML]{000000} 93.76 ± 2.16} & \cellcolor[HTML]{FFFFFF}{\color[HTML]{000000} 96.64 ± 0.38} & \cellcolor[HTML]{FFFFFF}{\color[HTML]{000000} 94.56} \\
 & \textbf{Type7} & \cellcolor[HTML]{FFFFFF}{\color[HTML]{000000} 92.61 ± 0.08} & \cellcolor[HTML]{FFFFFF}{\color[HTML]{000000} 94.76 ± 0.35} & \cellcolor[HTML]{FFFFFF}{\color[HTML]{000000} 94.05 ± 0.02} & \cellcolor[HTML]{FFFFFF}{\color[HTML]{000000} 95.35 ± 0.08} & \cellcolor[HTML]{FFFFFF}{\color[HTML]{000000} 94.19} & \cellcolor[HTML]{C0C0C0}{\color[HTML]{000000} 91.82 ± 0.73} & \cellcolor[HTML]{FFFFFF}{\color[HTML]{000000} 94.36 ± 0.02} & \cellcolor[HTML]{FFFFFF}{\color[HTML]{000000} 90.99 ± 0.72} & \cellcolor[HTML]{FFFFFF}{\color[HTML]{000000} 92.27 ± 0.04} & \cellcolor[HTML]{FFFFFF}{\color[HTML]{000000} 92.36} \\
 & \textbf{Type8} & \cellcolor[HTML]{FFFFFF}{\color[HTML]{000000} 92.38 ± 0.42} & \cellcolor[HTML]{FFFFFF}{\color[HTML]{000000} 95.71 ± 0.37} & \cellcolor[HTML]{FFFFFF}{\color[HTML]{000000} 94.56 ± 0.56} & \cellcolor[HTML]{FFFFFF}{\color[HTML]{000000} 96.13 ± 0.22} & \cellcolor[HTML]{FFFFFF}{\color[HTML]{000000} 94.70} & \cellcolor[HTML]{C0C0C0}{\color[HTML]{000000} 91.76 ± 0.62} & \cellcolor[HTML]{FFFFFF}{\color[HTML]{000000} 94.43 ± 0.08} & \cellcolor[HTML]{FFFFFF}{\color[HTML]{000000} 93.37 ± 0.06} & \cellcolor[HTML]{FFFFFF}{\color[HTML]{000000} 94.77 ± 0.09} & \cellcolor[HTML]{FFFFFF}{\color[HTML]{000000} 93.58} \\
\multirow{-12}{*}{\textbf{Gossipcop}} & \textbf{Type9} & \cellcolor[HTML]{FFFFFF}{\color[HTML]{000000} 93.12 ± 0.16} & \cellcolor[HTML]{FFFFFF}{\color[HTML]{000000} 95.92 ± 0.18} & \cellcolor[HTML]{FFFFFF}{\color[HTML]{000000} 94.90 ± 0.68} & \cellcolor[HTML]{FFFFFF}{\color[HTML]{000000} 96.65 ± 0.32} & \cellcolor[HTML]{FFFFFF}{\color[HTML]{000000} 95.15} & \cellcolor[HTML]{C0C0C0}{\color[HTML]{000000} 92.45 ± 0.43} & \cellcolor[HTML]{FFFFFF}{\color[HTML]{000000} 94.39 ± 0.06} & \cellcolor[HTML]{FFFFFF}{\color[HTML]{000000} 92.35 ± 0.46} & \cellcolor[HTML]{FFFFFF}{\color[HTML]{000000} 94.52 ± 0.19} & \cellcolor[HTML]{FFFFFF}{\color[HTML]{000000} 93.43} \\ \hline
 & \textbf{No Aug} & \cellcolor[HTML]{FFFFFF}{\color[HTML]{000000} 77.65 ± 1.52} & \cellcolor[HTML]{FFFFFF}{\color[HTML]{000000} 80.27 ± 0.99} & \cellcolor[HTML]{FFFFFF}{\color[HTML]{000000} 82.53 ± 1.96} & \cellcolor[HTML]{FFFFFF}{\color[HTML]{000000} 83.84 ± 1.32} & \cellcolor[HTML]{FFFFFF}{\color[HTML]{000000} 81.07} & \cellcolor[HTML]{FFFFFF}{\color[HTML]{000000} 76.47 ± 0.45} & \cellcolor[HTML]{FFFFFF}{\color[HTML]{000000} 78.05 ± 0.95} & \cellcolor[HTML]{FFFFFF}{\color[HTML]{000000} 80.31 ± 1.59} & \cellcolor[HTML]{FFFFFF}{\color[HTML]{000000} 81.91 ± 1.28} & \cellcolor[HTML]{FFFFFF}{\color[HTML]{000000} 79.19} \\
 & \textbf{MAXPOOL} & \cellcolor[HTML]{FFFFFF}{\color[HTML]{000000} 77.83 ± 0.64} & \cellcolor[HTML]{FFFFFF}{\color[HTML]{000000} 80.54 ± 0.72} & \cellcolor[HTML]{FFFFFF}{\color[HTML]{000000} 83.26 ± 0.96} & \cellcolor[HTML]{FFFFFF}{\color[HTML]{000000} 85.37 ± 0.88} & \cellcolor[HTML]{FFFFFF}{\color[HTML]{000000} 81.75} & \cellcolor[HTML]{FFFFFF}{\color[HTML]{000000} 76.82 ± 0.42} & \cellcolor[HTML]{FFFFFF}{\color[HTML]{000000} 78.51 ± 0.96} & \cellcolor[HTML]{FFFFFF}{\color[HTML]{000000} 82.12 ± 0.31} & \cellcolor[HTML]{FFFFFF}{\color[HTML]{000000} 82.12 ± 0.31} & \cellcolor[HTML]{FFFFFF}{\color[HTML]{000000} 79.89} \\
 & \textbf{GMT} & \cellcolor[HTML]{FFFFFF}{\color[HTML]{000000} 77.15 ± 0.33} & \cellcolor[HTML]{FFFFFF}{\color[HTML]{000000} 79.19 ± 1.25} & \cellcolor[HTML]{C0C0C0}{\color[HTML]{000000} 84.01 ± 0.94} & \cellcolor[HTML]{FFFFFF}{\color[HTML]{000000} 84.11 ± 1.05} & \cellcolor[HTML]{FFFFFF}{\color[HTML]{000000} 81.12} & \cellcolor[HTML]{FFFFFF}{\color[HTML]{000000} 71.94 ± 0.57} & \cellcolor[HTML]{FFFFFF}{\color[HTML]{000000} 65.61 ± 3.21} & \cellcolor[HTML]{FFFFFF}{\color[HTML]{000000} 81.44 ± 0.79} & \cellcolor[HTML]{FFFFFF}{\color[HTML]{000000} 78.06 ± 0.95} & \cellcolor[HTML]{FFFFFF}{\color[HTML]{000000} 74.26} \\
 & \textbf{Type1} & \cellcolor[HTML]{C0C0C0}{\color[HTML]{000000} 78.89 ± 1.88} & \cellcolor[HTML]{C0C0C0}{\color[HTML]{000000} 80.69 ± 0.95} & \cellcolor[HTML]{C0C0C0}{\color[HTML]{000000} 83.71 ± 0.45} & \cellcolor[HTML]{C0C0C0}{\color[HTML]{000000} 86.27 ± 0.94} & \cellcolor[HTML]{C0C0C0}{\color[HTML]{FE0000} 82.39} & \cellcolor[HTML]{C0C0C0}{\color[HTML]{000000} 77.37 ± 0.63} & \cellcolor[HTML]{C0C0C0}{\color[HTML]{000000} 79.86 ± 0.33} & \cellcolor[HTML]{FFFFFF}{\color[HTML]{000000} 81.67 ± 2.23} & \cellcolor[HTML]{C0C0C0}{\color[HTML]{000000} 85.29 ± 0.91} & \cellcolor[HTML]{C0C0C0}{\color[HTML]{FE0000} 81.05} \\
 & \textbf{Type2} & \cellcolor[HTML]{FFFFFF}{\color[HTML]{000000} 77.61 ± 0.32} & \cellcolor[HTML]{C0C0C0}{\color[HTML]{000000} 80.59 ± 1.19} & \cellcolor[HTML]{FFFFFF}{\color[HTML]{000000} 82.13 ± 1.59} & \cellcolor[HTML]{FFFFFF}{\color[HTML]{000000} 84.73 ± 2.23} & \cellcolor[HTML]{FFFFFF}{\color[HTML]{000000} 81.27} & \cellcolor[HTML]{C0C0C0}{\color[HTML]{000000} 76.94 ± 1.19} & \cellcolor[HTML]{FFFFFF}{\color[HTML]{000000} 76.46 ± 0.64} & \cellcolor[HTML]{FFFFFF}{\color[HTML]{000000} 81.21 ± 0.32} & \cellcolor[HTML]{C0C0C0}{\color[HTML]{000000} 82.35 ± 2.07} & \cellcolor[HTML]{FFFFFF}{\color[HTML]{000000} 79.24} \\
 & \textbf{Type3} & \cellcolor[HTML]{C0C0C0}{\color[HTML]{000000} 78.73 ± 0.67} & \cellcolor[HTML]{FFFFFF}{\color[HTML]{000000} 80.43 ± 1.17} & \cellcolor[HTML]{FFFFFF}{\color[HTML]{000000} 82.96 ± 0.53} & \cellcolor[HTML]{FFFFFF}{\color[HTML]{000000} 84.71 ± 3.31} & \cellcolor[HTML]{FFFFFF}{\color[HTML]{000000} 81.71} & \cellcolor[HTML]{FFFFFF}{\color[HTML]{000000} 75.33 ± 2.24} & \cellcolor[HTML]{C0C0C0}{\color[HTML]{000000} 79.18 ± 0.63} & \cellcolor[HTML]{FFFFFF}{\color[HTML]{000000} 81.89 ± 0.65} & \cellcolor[HTML]{C0C0C0}{\color[HTML]{000000} 82.34 ± 1.26} & \cellcolor[HTML]{FFFFFF}{\color[HTML]{000000} 79.69} \\
 & \textbf{Type4} & \cellcolor[HTML]{FFFFFF}{\color[HTML]{000000} 77.23 ± 0.55} & \cellcolor[HTML]{C0C0C0}{\color[HTML]{000000} 80.66 ± 0.67} & \cellcolor[HTML]{C0C0C0}{\color[HTML]{000000} 84.01 ± 0.94} & \cellcolor[HTML]{FFFFFF}{\color[HTML]{000000} 83.76 ± 0.02} & \cellcolor[HTML]{FFFFFF}{\color[HTML]{000000} 81.42} & \cellcolor[HTML]{FFFFFF}{\color[HTML]{000000} 75.79 ± 0.33} & \cellcolor[HTML]{C0C0C0}{\color[HTML]{000000} 79.41 ± 0.96} & \cellcolor[HTML]{FFFFFF}{\color[HTML]{000000} 80.99 ± 2.56} & \cellcolor[HTML]{FFFFFF}{\color[HTML]{000000} 74.88 ± 0.33} & \cellcolor[HTML]{FFFFFF}{\color[HTML]{000000} 77.77} \\
 & \textbf{Type5} & \cellcolor[HTML]{FFFFFF}{\color[HTML]{000000} 76.89 ± 0.99} & \cellcolor[HTML]{FFFFFF}{\color[HTML]{000000} 80.32 ± 0.89} & \cellcolor[HTML]{FFFFFF}{\color[HTML]{000000} 82.58 ± 1.23} & \cellcolor[HTML]{FFFFFF}{\color[HTML]{000000} 81.86 ± 0.12} & \cellcolor[HTML]{FFFFFF}{\color[HTML]{000000} 80.41} & \cellcolor[HTML]{FFFFFF}{\color[HTML]{000000} 75.56 ± 1.27} & \cellcolor[HTML]{FFFFFF}{\color[HTML]{000000} 74.66 ± 1.29} & \cellcolor[HTML]{FFFFFF}{\color[HTML]{000000} 80.84 ± 1.46} & \cellcolor[HTML]{FFFFFF}{\color[HTML]{000000} 76.71 ± 3.52} & \cellcolor[HTML]{FFFFFF}{\color[HTML]{000000} 76.94} \\
 & \textbf{Type6} & \cellcolor[HTML]{FFFFFF}{\color[HTML]{000000} 76.57 ± 0.23} & \cellcolor[HTML]{FFFFFF}{\color[HTML]{000000} 80.09 ± 0.63} & \cellcolor[HTML]{C0C0C0}{\color[HTML]{000000} 83.27 ± 1.21} & \cellcolor[HTML]{FFFFFF}{\color[HTML]{000000} 83.26 ± 0.22} & \cellcolor[HTML]{FFFFFF}{\color[HTML]{000000} 80.80} & \cellcolor[HTML]{FFFFFF}{\color[HTML]{000000} 73.98 ± 1.59} & \cellcolor[HTML]{FFFFFF}{\color[HTML]{000000} 76.69 ± 2.24} & \cellcolor[HTML]{FFFFFF}{\color[HTML]{000000} 79.18 ± 1.92} & \cellcolor[HTML]{FFFFFF}{\color[HTML]{000000} 78.28 ± 2.56} & \cellcolor[HTML]{FFFFFF}{\color[HTML]{000000} 77.03} \\
 & \textbf{Type7} & \cellcolor[HTML]{FFFFFF}{\color[HTML]{000000} 76.93 ± 1.28} & \cellcolor[HTML]{FFFFFF}{\color[HTML]{000000} 79.37 ± 1.27} & \cellcolor[HTML]{FFFFFF}{\color[HTML]{000000} 82.21 ± 1.14} & \cellcolor[HTML]{FFFFFF}{\color[HTML]{000000} 82.81 ± 0.11} & \cellcolor[HTML]{FFFFFF}{\color[HTML]{000000} 80.33} & \cellcolor[HTML]{FFFFFF}{\color[HTML]{000000} 71.72 ± 0.32} & \cellcolor[HTML]{FFFFFF}{\color[HTML]{000000} 78.28 ± 0.61} & \cellcolor[HTML]{FFFFFF}{\color[HTML]{000000} 78.28 ± 0.61} & \cellcolor[HTML]{FFFFFF}{\color[HTML]{000000} 73.53 ± 0.85} & \cellcolor[HTML]{FFFFFF}{\color[HTML]{000000} 75.45} \\
 & \textbf{Type8} & \cellcolor[HTML]{FFFFFF}{\color[HTML]{000000} 76.71 ± 0.97} & \cellcolor[HTML]{C0C0C0}{\color[HTML]{000000} 80.55 ± 1.46} & \cellcolor[HTML]{FFFFFF}{\color[HTML]{000000} 81.91 ± 1.28} & \cellcolor[HTML]{FFFFFF}{\color[HTML]{000000} 84.39 ± 0.33} & \cellcolor[HTML]{FFFFFF}{\color[HTML]{000000} 80.89} & \cellcolor[HTML]{FFFFFF}{\color[HTML]{000000} 74.21 ± 1.17} & \cellcolor[HTML]{C0C0C0}{\color[HTML]{000000} 78.96 ± 0.31} & \cellcolor[HTML]{FFFFFF}{\color[HTML]{000000} 78.96 ± 0.31} & \cellcolor[HTML]{FFFFFF}{\color[HTML]{000000} 75.86 ± 2.14} & \cellcolor[HTML]{FFFFFF}{\color[HTML]{000000} 77.00} \\
\multirow{-12}{*}{\textbf{Politifact}} & \textbf{Type9} & \cellcolor[HTML]{FFFFFF}{\color[HTML]{000000} 76.69 ± 0.32} & \cellcolor[HTML]{FFFFFF}{\color[HTML]{000000} 80.47 ± 1.02} & \cellcolor[HTML]{C0C0C0}{\color[HTML]{000000} 84.16 ± 0.78} & \cellcolor[HTML]{FFFFFF}{\color[HTML]{000000} 85.31 ± 0.32} & \cellcolor[HTML]{FFFFFF}{\color[HTML]{000000} 81.66} & \cellcolor[HTML]{FFFFFF}{\color[HTML]{000000} 75.56 ± 0.66} & \cellcolor[HTML]{C0C0C0}{\color[HTML]{000000} 79.63 ± 0.67} & \cellcolor[HTML]{C0C0C0}{\color[HTML]{000000} 83.48 ± 0.28} & \cellcolor[HTML]{FFFFFF}{\color[HTML]{000000} 76.92 ± 3.19} & \cellcolor[HTML]{FFFFFF}{\color[HTML]{000000} 78.90} \\ \hline
\end{tabular}
}
\end{table*}

The left part in Table \ref{table:Pool_gin} is the results of the GIN model. Surprisingly, in all cases examined, both operator Type 1 and operator Type 3 consistently outperform MAXPOOL on both the Gossipcop and Politifact datasets. Compared with GMT, the operator Type 1 improves the accuracy by up to 2.10\% in Gossipcop-Profile and 1.58\% on the Politifact-Bert, respectively. Also, results from the column Average show that the operator Type 1 is still the best choice for both the Gossipcop and Politifact datasets. We focus on Type 1, Type 2, and Type 3 for the average comparison, as the Average column indicates that these three types of hybrid pooling operators perform the best. Compared to MAXPOOL, Type 1 exhibits the highest average accuracy improvement, at up to 0.86\%, followed by Type 2 at 0.39\% and Type 3 at 0.32\%. Moving to GMT, we reach the same conclusion as with MAXPOOL, where Type 1 achieves the highest average improvement, up to 0.97\%, followed by Type 2 with 0.56\% and Type 3 with 0.48\%.

Table \ref{table:results2} presents the results of PL datasets. We can see that the advanced graph pooling operators, including GMT and hybrid pooling operators, are more helpful than the MAXPOOL for \emph{Mainfold-Mixup} when dealing with PL data, where the best results are always from the hybrid pooling operators. Similar to the results of traditional NLP datasets, the Type 1 hybrid pooling operator outperforms MAXPOOL in most situations (6 out of 8 cases). Unlike the NLP datasets, the operator Type 3 consistently achieves better results than MAXPOOL. In PL datasets, the gap between using GMT and hybrid pooling operators can be up to 2.36\% of clean accuracy (GIN-Virtural, Python800, Type 1). When considering average improvement, we focus on Type 1 and Type 3 across all GNNs, as they demonstrate better performance compared to others in the results. Compared to MAXPOOL, Type 1 exhibits a better average accuracy improvement of 0.59\% compared to Type 3, which is 0.44\%. In the case of GMT, Type 1 also outperforms Type 3 in average accuracy improvement, with 0.92\% compared to 0.90\%.

\textbf{Statistical analysis.} To check the significance of the impact of hybrid pooling on the model accuracy, we conduct a statistical analysis. Specifically, we use \emph{Wilcoxon signed-rank test}~\cite{woolson2007wilcoxon} to analyze the significance. Since there are many comparisons, we adapt \emph{Bonferroni correction}~\cite{armstrong2014use} to adjust the \emph{P-values} before concluding. The upper component in Table~\ref{table:statistical_test_accuracy_robustness} presents the comparison between pooling operators and No Aug, MAXPOOL, and GMT. From the results, we can see that Type1 significantly outperforms the three baselines both on PL and NLP. In most cases, hybrid pooling operators such as Type 4, Type 5, Type 6, and Type 9 are significantly worse than the baselines.

Overall, based on the analysis including maximum improvement, average improvement, and statistical significance, we can conclude that hybrid pooling operators have a significant impact on the effectiveness of the Mixup data augmentation technique when dealing with graph-structured datasets. More specifically, for accuracy improvement, Type 1 ($\mathcal{M}_{sum}(\mathcal{P}_{att},\mathcal{P}_{max})$) is the best choice among our considered candidates, while Type 3 ($\mathcal{M}_{concat}(\mathcal{P}_{att},\mathcal{P}_{max})$) also demonstrates relatively strong performance. Both Type 1 and Type 3 are recommended for use in graph classification tasks, given their superior performance across all evaluated metrics, datasets (PL and NLP), and GNN architectures.

\noindent\colorbox{gray!20}{\framebox{\parbox{0.96\linewidth}{
\textbf{Answer to RQ1}: Hybrid pooling operators have a significant impact in enhancing the effectiveness of \emph{Manifold-Mixup}-based graph learning in terms of clean accuracy. Considering the analysis including maximum improvement, average improvement, and statistical significance, we recommend Type 1 as the optimal choice. Specifically, compared to GMT, the Type 1 operator~($\mathcal{M}_{sum}(\mathcal{P}_{att},\mathcal{P}_{max})$) demonstrates up to 4.38\% accuracy improvements on the NLP dataset and 2.36\% on the PL dataset, respectively.}}}

\subsection{RQ2: How effective are hybrid pooling operators for enhancing the robustness of Mixup-based graph learning? }

\textbf{Robustness analysis.} The right part in Table~\ref{table:results1} presents the results of the robustness of the GCN model on the NLP dataset. Firstly, it is observed that \emph{Manifold-Mixup} significantly enhances the generalization ability of the GNN model. In addition, although GMT is an advanced graph pooling operator, it is unable to help \emph{Manifold-Mixup} effectively enhance the robustness of the GCN model. It proves to be effective in only 3 out of 8 cases.  Then, concerning hybrid pooling operators, surprisingly, all operators obtain better performance in improving the robustness than MAXPOOL on the Gossipcop-Profile and Politifact-Profile datasets. Furthermore, in terms of robustness, the operator Type 9 outperforms GMT by a significant margin of up to 13.46\% (Gossipcop-Profile), and the operator Type 4 obtains an even higher improvement in robustness by up to 15.29\% (Politifact-Spacy) compared to GMT. When examining the average robustness improvement, we focus on three hybrid pooling operators including Type 1, Type 4, and Type 9, as they demonstrate better performance compared to others, as shown in the Average column. Compared to MAXPOOL, Type 9 exhibits the highest average robustness improvement, at 6.18\%, followed by Type 4 at 5.96\% and Type 1 at 3.38\%.  In the case of GMT, different from MAXPOOL, Type 4 achieves a better average robustness improvement of 6.23\% compared to Type 1 at 6.18\% and Type 9 at 5.74\%.

\begin{table*}[!tb]
\caption{Effectiveness of nine types of hybrid pooling w.r.t. test accuracy $\uparrow$ (average $\pm$ standard deviation, \%) on original test data and robustness $\uparrow$ (average $\pm$ standard deviation, \%) on robust test data of trained GIN models on NLP datasets. \textbf{No Aug}: without Manifold-Mixup data augmentation. A gray background highlights results that are better than MAXPOOL. The best average results are marked in red color.}

\label{table:Pool_gin}

\centering
\scalebox{0.55}{
\begin{tabular}{llllllc||llllc}
\hline
 &  & \multicolumn{5}{c||}{\textbf{Test Accuracy}} & \multicolumn{5}{c}{\textbf{Robustness}} \\ 
 &  & \multicolumn{1}{c}{\textbf{Profile}} & \multicolumn{1}{c}{\textbf{Spacy}} & \multicolumn{1}{c}{\textbf{Bert}} & \multicolumn{1}{c}{\textbf{Content}} & \textbf{Average} & \multicolumn{1}{c}{\textbf{Profile}} & \multicolumn{1}{c}{\textbf{Spacy}} & \multicolumn{1}{c}{\textbf{Bert}} & \multicolumn{1}{c}{\textbf{Content}} & \textbf{Average} \\ \hline
 & \textbf{No Aug} & \cellcolor[HTML]{FFFFFF}{\color[HTML]{000000} 93.39 ± 0.29} & \cellcolor[HTML]{FFFFFF}{\color[HTML]{000000} 95.86 ± 0.34} & \cellcolor[HTML]{FFFFFF}{\color[HTML]{000000} 94.32 ± 0.11} & \cellcolor[HTML]{FFFFFF}{\color[HTML]{000000} 95.75 ± 0.45} & \cellcolor[HTML]{FFFFFF}{\color[HTML]{000000} 94.83} & \cellcolor[HTML]{FFFFFF}{\color[HTML]{000000} 82.04 ± 1.99} & \cellcolor[HTML]{FFFFFF}{\color[HTML]{000000} 86.81 ± 3.04} & \cellcolor[HTML]{FFFFFF}{\color[HTML]{000000} 72.92 ± 1.04} & \cellcolor[HTML]{FFFFFF}{\color[HTML]{000000} 74.18 ± 0.16} & \cellcolor[HTML]{FFFFFF}{\color[HTML]{000000} 78.99} \\
 & \textbf{MAXPOOL} & \cellcolor[HTML]{FFFFFF}{\color[HTML]{000000} 93.51 ± 0.27} & \cellcolor[HTML]{FFFFFF}{\color[HTML]{000000} 96.01 ± 0.24} & \cellcolor[HTML]{FFFFFF}{\color[HTML]{000000} 94.91 ± 0.56} & \cellcolor[HTML]{FFFFFF}{\color[HTML]{000000} 96.42 ± 0.46} & \cellcolor[HTML]{FFFFFF}{\color[HTML]{000000} 95.21} & \cellcolor[HTML]{FFFFFF}{\color[HTML]{000000} 85.46 ± 0.83} & \cellcolor[HTML]{FFFFFF}{\color[HTML]{000000} 87.01 ± 3.16} & \cellcolor[HTML]{FFFFFF}{\color[HTML]{000000} 74.27 ± 2.87} & \cellcolor[HTML]{FFFFFF}{\color[HTML]{000000} 76.45 ± 0.22} & \cellcolor[HTML]{FFFFFF}{\color[HTML]{000000} 80.80} \\
 & \textbf{GMT} & \cellcolor[HTML]{FFFFFF}{\color[HTML]{000000} 92.44 ± 0.25} & \cellcolor[HTML]{FFFFFF}{\color[HTML]{000000} 95.53 ± 0.28} & \cellcolor[HTML]{FFFFFF}{\color[HTML]{000000} 94.73 ± 0.13} & \cellcolor[HTML]{FFFFFF}{\color[HTML]{000000} 96.29 ± 0.03} & \cellcolor[HTML]{FFFFFF}{\color[HTML]{000000} 94.75} & \cellcolor[HTML]{FFFFFF}{\color[HTML]{000000} 70.63 ± 1.22} & \cellcolor[HTML]{FFFFFF}{\color[HTML]{000000} 75.65 ± 3.64} & \cellcolor[HTML]{FFFFFF}{\color[HTML]{000000} 73.23 ± 0.74} & \cellcolor[HTML]{C0C0C0}{\color[HTML]{000000} 84.51 ± 0.91} & \cellcolor[HTML]{FFFFFF}{\color[HTML]{000000} 76.01} \\
 & \textbf{Type1} & \cellcolor[HTML]{C0C0C0}{\color[HTML]{000000} 94.54 ± 0.21} & \cellcolor[HTML]{C0C0C0}{\color[HTML]{000000} 96.51 ± 0.22} & \cellcolor[HTML]{C0C0C0}{\color[HTML]{000000} 95.41 ± 0.07} & \cellcolor[HTML]{C0C0C0}{\color[HTML]{000000} 97.19 ± 0.19} & \cellcolor[HTML]{C0C0C0}{\color[HTML]{FE0000} 95.91} & \cellcolor[HTML]{C0C0C0}{\color[HTML]{000000} 87.98 ± 0.48} & \cellcolor[HTML]{C0C0C0}{\color[HTML]{000000} 88.36 ± 2.35} & \cellcolor[HTML]{C0C0C0}{\color[HTML]{000000} 92.04 ± 0.15} & \cellcolor[HTML]{C0C0C0}{\color[HTML]{000000} 92.65 ± 0.27} & \cellcolor[HTML]{C0C0C0}{\color[HTML]{FE0000} 90.26} \\
 & \textbf{Type2} & \cellcolor[HTML]{C0C0C0}{\color[HTML]{000000} 94.06 ± 0.32} & \cellcolor[HTML]{FFFFFF}{\color[HTML]{000000} 95.83 ± 0.18} & \cellcolor[HTML]{C0C0C0}{\color[HTML]{000000} 95.11 ± 0.38} & \cellcolor[HTML]{C0C0C0}{\color[HTML]{000000} 96.89 ± 0.11} & \cellcolor[HTML]{C0C0C0}{\color[HTML]{000000} 95.47} & \cellcolor[HTML]{FFFFFF}{\color[HTML]{000000} 72.01 ± 0.18} & \cellcolor[HTML]{FFFFFF}{\color[HTML]{000000} 72.79 ± 0.35} & \cellcolor[HTML]{FFFFFF}{\color[HTML]{000000} 71.02 ± 0.22} & \cellcolor[HTML]{FFFFFF}{\color[HTML]{000000} 72.06 ± 2.27} & \cellcolor[HTML]{FFFFFF}{\color[HTML]{000000} 71.97} \\
 & \textbf{Type3} & \cellcolor[HTML]{C0C0C0}{\color[HTML]{000000} 93.83 ± 0.34} & \cellcolor[HTML]{C0C0C0}{\color[HTML]{000000} 96.17 ± 0.31} & \cellcolor[HTML]{C0C0C0}{\color[HTML]{000000} 94.91 ± 0.71} & \cellcolor[HTML]{C0C0C0}{\color[HTML]{000000} 96.62 ± 0.25} & \cellcolor[HTML]{C0C0C0}{\color[HTML]{000000} 95.38} & \cellcolor[HTML]{FFFFFF}{\color[HTML]{000000} 81.04 ± 0.25} & \cellcolor[HTML]{FFFFFF}{\color[HTML]{000000} 76.13 ± 5.34} & \cellcolor[HTML]{FFFFFF}{\color[HTML]{000000} 71.11 ± 0.98} & \cellcolor[HTML]{C0C0C0}{\color[HTML]{000000} 85.57 ± 2.41} & \cellcolor[HTML]{FFFFFF}{\color[HTML]{000000} 78.46} \\
 & \textbf{Type4} & \cellcolor[HTML]{FFFFFF}{\color[HTML]{000000} 93.42 ± 0.22} & \cellcolor[HTML]{FFFFFF}{\color[HTML]{000000} 95.92 ± 0.29} & \cellcolor[HTML]{FFFFFF}{\color[HTML]{000000} 93.12 ± 0.69} & \cellcolor[HTML]{C0C0C0}{\color[HTML]{000000} 96.75 ± 0.31} & \cellcolor[HTML]{FFFFFF}{\color[HTML]{000000} 94.80} & \cellcolor[HTML]{FFFFFF}{\color[HTML]{000000} 84.47 ± 0.74} & \cellcolor[HTML]{FFFFFF}{\color[HTML]{000000} 73.31 ± 1.97} & \cellcolor[HTML]{C0C0C0}{\color[HTML]{000000} 83.46 ± 2.16} & \cellcolor[HTML]{C0C0C0}{\color[HTML]{000000} 86.18 ± 2.31} & \cellcolor[HTML]{C0C0C0}{\color[HTML]{000000} 81.86} \\
 & \textbf{Type5} & \cellcolor[HTML]{FFFFFF}{\color[HTML]{000000} 93.22 ± 0.06} & \cellcolor[HTML]{FFFFFF}{\color[HTML]{000000} 95.05 ± 0.37} & \cellcolor[HTML]{FFFFFF}{\color[HTML]{000000} 91.98 ± 1.22} & \cellcolor[HTML]{C0C0C0}{\color[HTML]{000000} 96.94 ± 0.41} & \cellcolor[HTML]{FFFFFF}{\color[HTML]{000000} 94.30} & \cellcolor[HTML]{C0C0C0}{\color[HTML]{000000} 91.68 ± 0.76} & \cellcolor[HTML]{FFFFFF}{\color[HTML]{000000} 74.59 ± 1.47} & \cellcolor[HTML]{C0C0C0}{\color[HTML]{000000} 84.05 ± 4.13} & \cellcolor[HTML]{C0C0C0}{\color[HTML]{000000} 88.59 ± 1.97} & \cellcolor[HTML]{C0C0C0}{\color[HTML]{000000} 84.73} \\
 & \textbf{Type6} & \cellcolor[HTML]{C0C0C0}{\color[HTML]{000000} 94.66 ± 0.52} & \cellcolor[HTML]{C0C0C0}{\color[HTML]{000000} 96.11 ± 0.33} & \cellcolor[HTML]{C0C0C0}{\color[HTML]{000000} 94.99 ± 0.09} & \cellcolor[HTML]{C0C0C0}{\color[HTML]{000000} 96.61 ± 0.76} & \cellcolor[HTML]{C0C0C0}{\color[HTML]{000000} 95.59} & \cellcolor[HTML]{C0C0C0}{\color[HTML]{000000} 93.86 ± 0.22} & \cellcolor[HTML]{FFFFFF}{\color[HTML]{000000} 75.95 ± 0.66} & \cellcolor[HTML]{C0C0C0}{\color[HTML]{000000} 78.17 ± 3.36} & \cellcolor[HTML]{C0C0C0}{\color[HTML]{000000} 89.11 ± 2.08} & \cellcolor[HTML]{C0C0C0}{\color[HTML]{000000} 84.27} \\
 & \textbf{Type7} & \cellcolor[HTML]{FFFFFF}{\color[HTML]{000000} 93.17 ± 0.35} & \cellcolor[HTML]{FFFFFF}{\color[HTML]{000000} 95.73 ± 0.23} & \cellcolor[HTML]{FFFFFF}{\color[HTML]{000000} 94.13 ± 0.79} & \cellcolor[HTML]{C0C0C0}{\color[HTML]{000000} 96.96 ± 0.15} & \cellcolor[HTML]{FFFFFF}{\color[HTML]{000000} 95.00} & \cellcolor[HTML]{FFFFFF}{\color[HTML]{000000} 73.38 ± 2.57} & \cellcolor[HTML]{FFFFFF}{\color[HTML]{000000} 85.99 ± 2.58} & \cellcolor[HTML]{C0C0C0}{\color[HTML]{000000} 81.51 ± 2.49} & \cellcolor[HTML]{C0C0C0}{\color[HTML]{000000} 89.68 ± 0.75} & \cellcolor[HTML]{C0C0C0}{\color[HTML]{000000} 82.64} \\
 & \textbf{Type8} & \cellcolor[HTML]{FFFFFF}{\color[HTML]{000000} 92.02 ± 0.86} & \cellcolor[HTML]{FFFFFF}{\color[HTML]{000000} 95.43 ± 0.45} & \cellcolor[HTML]{FFFFFF}{\color[HTML]{000000} 92.05 ± 0.73} & \cellcolor[HTML]{FFFFFF}{\color[HTML]{000000} 95.44 ± 0.26} & \cellcolor[HTML]{FFFFFF}{\color[HTML]{000000} 93.74} & \cellcolor[HTML]{FFFFFF}{\color[HTML]{000000} 62.14 ± 1.22} & \cellcolor[HTML]{FFFFFF}{\color[HTML]{000000} 86.01 ± 0.44} & \cellcolor[HTML]{FFFFFF}{\color[HTML]{000000} 71.77 ± 0.11} & \cellcolor[HTML]{FFFFFF}{\color[HTML]{000000} 65.76 ± 4.51} & \cellcolor[HTML]{FFFFFF}{\color[HTML]{000000} 71.42} \\
\multirow{-12}{*}{\textbf{Gossipcop}} & \textbf{Type9} & \cellcolor[HTML]{FFFFFF}{\color[HTML]{000000} 93.08 ± 0.34} & \cellcolor[HTML]{FFFFFF}{\color[HTML]{000000} 95.93 ± 0.21} & \cellcolor[HTML]{FFFFFF}{\color[HTML]{000000} 93.81 ± 0.41} & \cellcolor[HTML]{C0C0C0}{\color[HTML]{000000} 97.01 ± 0.51} & \cellcolor[HTML]{FFFFFF}{\color[HTML]{000000} 94.96} & \cellcolor[HTML]{FFFFFF}{\color[HTML]{000000} 76.69 ± 1.41} & \cellcolor[HTML]{FFFFFF}{\color[HTML]{000000} 83.78 ± 0.72} & \cellcolor[HTML]{C0C0C0}{\color[HTML]{000000} 74.36 ± 2.96} & \cellcolor[HTML]{C0C0C0}{\color[HTML]{000000} 90.95 ± 1.47} & \cellcolor[HTML]{C0C0C0}{\color[HTML]{000000} 81.45} \\ \hline
 & \textbf{No Aug} & \cellcolor[HTML]{FFFFFF}{\color[HTML]{000000} 77.01 ± 1.91} & \cellcolor[HTML]{FFFFFF}{\color[HTML]{000000} 80.43 ± 0.68} & \cellcolor[HTML]{FFFFFF}{\color[HTML]{000000} 84.16 ± 0.32} & \cellcolor[HTML]{FFFFFF}{\color[HTML]{000000} 84.73 ± 1.35} & \cellcolor[HTML]{FFFFFF}{\color[HTML]{000000} 81.58} & \cellcolor[HTML]{FFFFFF}{\color[HTML]{000000} 61.99 ± 1.92} & \cellcolor[HTML]{FFFFFF}{\color[HTML]{000000} 76.24 ± 0.32} & \cellcolor[HTML]{FFFFFF}{\color[HTML]{000000} 81.89 ± 0.45} & \cellcolor[HTML]{FFFFFF}{\color[HTML]{000000} 50.75 ± 0.17} & \cellcolor[HTML]{FFFFFF}{\color[HTML]{000000} 67.72} \\
 & \textbf{MAXPOOL} & \cellcolor[HTML]{FFFFFF}{\color[HTML]{000000} 77.33 ± 0.89} & \cellcolor[HTML]{FFFFFF}{\color[HTML]{000000} 81.68 ± 1.14} & \cellcolor[HTML]{FFFFFF}{\color[HTML]{000000} 84.35 ± 0.55} & \cellcolor[HTML]{FFFFFF}{\color[HTML]{000000} 84.88 ± 1.59} & \cellcolor[HTML]{FFFFFF}{\color[HTML]{000000} 82.06} & \cellcolor[HTML]{FFFFFF}{\color[HTML]{000000} 68.78 ± 0.63} & \cellcolor[HTML]{FFFFFF}{\color[HTML]{000000} 77.17 ± 0.98} & \cellcolor[HTML]{FFFFFF}{\color[HTML]{000000} 82.12 ± 0.31} & \cellcolor[HTML]{FFFFFF}{\color[HTML]{000000} 51.12 ± 0.71} & \cellcolor[HTML]{FFFFFF}{\color[HTML]{000000} 69.80} \\
 & \textbf{GMT} & \cellcolor[HTML]{C0C0C0}{\color[HTML]{000000} 77.74 ± 1.26} & \cellcolor[HTML]{C0C0C0}{\color[HTML]{000000} 81.75 ± 1.01} & \cellcolor[HTML]{C0C0C0}{\color[HTML]{000000} 84.39 ± 1.69} & \cellcolor[HTML]{C0C0C0}{\color[HTML]{000000} 84.95 ± 1.35} & \cellcolor[HTML]{C0C0C0}{\color[HTML]{000000} 82.21} & \cellcolor[HTML]{FFFFFF}{\color[HTML]{000000} 61.68 ± 3.21} & \cellcolor[HTML]{FFFFFF}{\color[HTML]{000000} 61.43 ± 0.79} & \cellcolor[HTML]{C0C0C0}{\color[HTML]{000000} 83.48 ± 0.32} & \cellcolor[HTML]{C0C0C0}{\color[HTML]{000000} 82.81 ± 1.93} & \cellcolor[HTML]{C0C0C0}{\color[HTML]{000000} 72.35} \\
 & \textbf{Type1} & \cellcolor[HTML]{C0C0C0}{\color[HTML]{000000} 78.28 ± 0.93} & \cellcolor[HTML]{C0C0C0}{\color[HTML]{000000} 82.24 ± 0.57} & \cellcolor[HTML]{C0C0C0}{\color[HTML]{000000} 85.97 ± 0.27} & \cellcolor[HTML]{C0C0C0}{\color[HTML]{000000} 85.41 ± 1.19} & \cellcolor[HTML]{C0C0C0}{\color[HTML]{FE0000} 82.98} & \cellcolor[HTML]{C0C0C0}{\color[HTML]{000000} 73.31 ± 2.56} & \cellcolor[HTML]{FFFFFF}{\color[HTML]{000000} 76.91 ± 1.27} & \cellcolor[HTML]{C0C0C0}{\color[HTML]{000000} 83.03 ± 0.28} & \cellcolor[HTML]{FFFFFF}{\color[HTML]{000000} 50.93 ± 0.08} & \cellcolor[HTML]{C0C0C0}{\color[HTML]{000000} 71.05} \\
 & \textbf{Type2} & \cellcolor[HTML]{C0C0C0}{\color[HTML]{000000} 78.06 ± 1.61} & \cellcolor[HTML]{FFFFFF}{\color[HTML]{000000} 81.46 ± 0.28} & \cellcolor[HTML]{C0C0C0}{\color[HTML]{000000} 85.52 ± 0.35} & \cellcolor[HTML]{C0C0C0}{\color[HTML]{000000} 85.04 ± 1.77} & \cellcolor[HTML]{C0C0C0}{\color[HTML]{000000} 82.52} & \cellcolor[HTML]{C0C0C0}{\color[HTML]{000000} 74.21 ± 0.64} & \cellcolor[HTML]{FFFFFF}{\color[HTML]{000000} 74.36 ± 1.59} & \cellcolor[HTML]{FFFFFF}{\color[HTML]{000000} 81.67 ± 0.35} & \cellcolor[HTML]{FFFFFF}{\color[HTML]{000000} 50.61 ± 0.15} & \cellcolor[HTML]{C0C0C0}{\color[HTML]{000000} 70.21} \\
 & \textbf{Type3} & \cellcolor[HTML]{C0C0C0}{\color[HTML]{000000} 77.92 ± 2.85} & \cellcolor[HTML]{C0C0C0}{\color[HTML]{000000} 81.87 ± 0.94} & \cellcolor[HTML]{C0C0C0}{\color[HTML]{000000} 85.07 ± 0.31} & \cellcolor[HTML]{C0C0C0}{\color[HTML]{000000} 85.24 ± 1.36} & \cellcolor[HTML]{C0C0C0}{\color[HTML]{000000} 82.53} & \cellcolor[HTML]{C0C0C0}{\color[HTML]{000000} 75.06 ± 3.78} & \cellcolor[HTML]{FFFFFF}{\color[HTML]{000000} 76.46 ± 0.64} & \cellcolor[HTML]{FFFFFF}{\color[HTML]{000000} 80.99 ± 1.17} & \cellcolor[HTML]{FFFFFF}{\color[HTML]{000000} 50.82 ± 0.43} & \cellcolor[HTML]{C0C0C0}{\color[HTML]{000000} 70.83} \\
 & \textbf{Type4} & \cellcolor[HTML]{C0C0C0}{\color[HTML]{000000} 77.74 ± 0.74} & \cellcolor[HTML]{FFFFFF}{\color[HTML]{000000} 80.09 ± 0.55} & \cellcolor[HTML]{C0C0C0}{\color[HTML]{000000} 84.39 ± 0.96} & \cellcolor[HTML]{FFFFFF}{\color[HTML]{000000} 83.86 ± 1.88} & \cellcolor[HTML]{FFFFFF}{\color[HTML]{000000} 81.52} & \cellcolor[HTML]{C0C0C0}{\color[HTML]{000000} 75.56 ± 1.92} & \cellcolor[HTML]{FFFFFF}{\color[HTML]{000000} 76.69 ± 2.25} & \cellcolor[HTML]{C0C0C0}{\color[HTML]{000000} 82.65 ± 1.39} & \cellcolor[HTML]{C0C0C0}{\color[HTML]{000000} 79.86 ± 2.88} & \cellcolor[HTML]{C0C0C0}{\color[HTML]{000000} 78.69} \\
 & \textbf{Type5} & \cellcolor[HTML]{C0C0C0}{\color[HTML]{000000} 77.64 ± 1.72} & \cellcolor[HTML]{FFFFFF}{\color[HTML]{000000} 80.77 ± 0.98} & \cellcolor[HTML]{C0C0C0}{\color[HTML]{000000} 84.46 ± 0.53} & \cellcolor[HTML]{FFFFFF}{\color[HTML]{000000} 82.81 ± 0.64} & \cellcolor[HTML]{FFFFFF}{\color[HTML]{000000} 81.42} & \cellcolor[HTML]{C0C0C0}{\color[HTML]{000000} 76.41 ± 0.36} & \cellcolor[HTML]{C0C0C0}{\color[HTML]{000000} 78.28 ± 0.63} & \cellcolor[HTML]{C0C0C0}{\color[HTML]{000000} 83.71 ± 0.65} & \cellcolor[HTML]{C0C0C0}{\color[HTML]{000000} 78.05 ± 0.96} & \cellcolor[HTML]{C0C0C0}{\color[HTML]{000000} 79.11} \\
 & \textbf{Type6} & \cellcolor[HTML]{C0C0C0}{\color[HTML]{000000} 78.05 ± 1.08} & \cellcolor[HTML]{FFFFFF}{\color[HTML]{000000} 81.45 ± 0.64} & \cellcolor[HTML]{FFFFFF}{\color[HTML]{000000} 84.16 ± 0.22} & \cellcolor[HTML]{FFFFFF}{\color[HTML]{000000} 84.05 ± 1.86} & \cellcolor[HTML]{FFFFFF}{\color[HTML]{000000} 81.93} & \cellcolor[HTML]{C0C0C0}{\color[HTML]{000000} 74.66 ± 0.79} & \cellcolor[HTML]{C0C0C0}{\color[HTML]{000000} 77.92 ± 3.48} & \cellcolor[HTML]{C0C0C0}{\color[HTML]{000000} 82.36 ± 0.67} & \cellcolor[HTML]{C0C0C0}{\color[HTML]{000000} 77.31 ± 4.35} & \cellcolor[HTML]{C0C0C0}{\color[HTML]{000000} 78.06} \\
 & \textbf{Type7} & \cellcolor[HTML]{C0C0C0}{\color[HTML]{000000} 77.83 ± 1.61} & \cellcolor[HTML]{FFFFFF}{\color[HTML]{000000} 80.03 ± 0.49} & \cellcolor[HTML]{C0C0C0}{\color[HTML]{000000} 84.44 ± 0.68} & \cellcolor[HTML]{FFFFFF}{\color[HTML]{000000} 84.39 ± 0.94} & \cellcolor[HTML]{FFFFFF}{\color[HTML]{000000} 81.67} & \cellcolor[HTML]{C0C0C0}{\color[HTML]{000000} 75.11 ± 0.66} & \cellcolor[HTML]{FFFFFF}{\color[HTML]{000000} 76.92 ± 1.92} & \cellcolor[HTML]{FFFFFF}{\color[HTML]{000000} 81.89 ± 1.28} & \cellcolor[HTML]{C0C0C0}{\color[HTML]{000000} 76.79 ± 4.29} & \cellcolor[HTML]{C0C0C0}{\color[HTML]{000000} 77.68} \\
 & \textbf{Type8} & \cellcolor[HTML]{C0C0C0}{\color[HTML]{000000} 78.39 ± 0.78} & \cellcolor[HTML]{FFFFFF}{\color[HTML]{000000} 80.01 ± 0.51} & \cellcolor[HTML]{FFFFFF}{\color[HTML]{000000} 83.81 ± 1.32} & \cellcolor[HTML]{FFFFFF}{\color[HTML]{000000} 81.61 ± 0.52} & \cellcolor[HTML]{FFFFFF}{\color[HTML]{000000} 80.96} & \cellcolor[HTML]{FFFFFF}{\color[HTML]{000000} 57.19 ± 2.36} & \cellcolor[HTML]{FFFFFF}{\color[HTML]{000000} 67.64 ± 2.87} & \cellcolor[HTML]{FFFFFF}{\color[HTML]{000000} 77.14 ± 1.61} & \cellcolor[HTML]{C0C0C0}{\color[HTML]{000000} 61.24 ± 3.87} & \cellcolor[HTML]{FFFFFF}{\color[HTML]{000000} 65.80} \\
\multirow{-12}{*}{\textbf{Politifact}} & \textbf{Type9} & \cellcolor[HTML]{C0C0C0}{\color[HTML]{000000} 77.74 ± 1.71} & \cellcolor[HTML]{FFFFFF}{\color[HTML]{000000} 80.99 ± 0.68} & \cellcolor[HTML]{C0C0C0}{\color[HTML]{000000} 84.77 ± 1.03} & \cellcolor[HTML]{C0C0C0}{\color[HTML]{000000} 85.19 ± 1.93} & \cellcolor[HTML]{C0C0C0}{\color[HTML]{000000} 82.17} & \cellcolor[HTML]{C0C0C0}{\color[HTML]{000000} 74.36 ± 2.28} & \cellcolor[HTML]{C0C0C0}{\color[HTML]{000000} 77.61 ± 0.92} & \cellcolor[HTML]{C0C0C0}{\color[HTML]{000000} 83.48 ± 0.37} & \cellcolor[HTML]{C0C0C0}{\color[HTML]{000000} 83.03 ± 2.23} & \cellcolor[HTML]{C0C0C0}{\color[HTML]{FE0000} 79.62} \\ \hline
\end{tabular}
}
\end{table*}

\begin{table}[!tb]
\caption{Effectiveness of nine types of hybrid pooling w.r.t. test accuracy $\uparrow$ (average $\pm$ standard deviation, \%) on original test data and robustness $\uparrow$ (average $\pm$ standard deviation, \%) on robust test data on PL datasets. \textbf{No Aug}: without Manifold-Mixup data augmentation. A gray background highlights results that are better than MAXPOOL. The best average results are marked in red color.}
\label{table:results2}
\centering
\scalebox{0.6}{
\begin{tabular}{llll|lllll|ll}
\hline
 &  & \multicolumn{2}{c|}{\textbf{Test Accuracy}} & \multicolumn{2}{c}{\textbf{Robustness}} &   &\multicolumn{2}{c|}{\textbf{Test Accuracy}} & \multicolumn{2}{c}{\textbf{Robustness}}  \\
 &  & \multicolumn{1}{c}{\textbf{JAVA250}} & \multicolumn{1}{c|}{\textbf{Python800}} & \multicolumn{1}{c}{\textbf{JAVA250}} & \multicolumn{1}{c}{\textbf{Python800}} &  & \multicolumn{1}{c}{\textbf{JAVA250}} & \multicolumn{1}{c|}{\textbf{Python800}} & \multicolumn{1}{c}{\textbf{JAVA250}} & \multicolumn{1}{c}{\textbf{Python800}} \\ \hline
 
 & \textbf{No Aug} & 92.29 ± 0.17 & 93.73 ± 0.03 & 82.05 ± 2.03 & 82.37 ± 1.65 & {\color[HTML]{333333} } & 93.46 ± 0.07 & 93.89 ± 0.18 & 85.01 ± 1.34 & 85.48 ± 1.26 \\
 & \textbf{MAXPOOL} & 92.39 ± 0.08 & 93.94 ± 0.07 & 82.23 ± 2.14 & 82.63 ± 1.32 &{\color[HTML]{333333} } & 93.84 ± 0.31 & 94.27 ± 0.21 & 85.66 ± 2.01 & 86.13 ± 1.34  \\
 & \textbf{GMT} & \cellcolor[HTML]{C0C0C0}{\color[HTML]{333333} 92.66 ± 0.32} & \cellcolor[HTML]{C0C0C0}{\color[HTML]{333333} 94.36 ± 0.21} & \cellcolor[HTML]{C0C0C0}{\color[HTML]{333333} 82.69 ± 1.78} & 82.04 ± 2.34 & {\color[HTML]{333333} } & 92.49 ± 0.26 & 93.88 ± 0.33 & \cellcolor[HTML]{C0C0C0}{\color[HTML]{333333} 84.17 ± 1.87} & 82.77 ± 1.87 \\
 & \textbf{Type1} & \cellcolor[HTML]{C0C0C0}{\color[HTML]{333333} 92.89 ± 0.12} & \cellcolor[HTML]{C0C0C0}{\color[HTML]{333333} 94.29 ± 0.06} & \cellcolor[HTML]{C0C0C0}{\color[HTML]{333333} 86.85 ± 1.67} & \cellcolor[HTML]{C0C0C0}{\color[HTML]{FE0000} \textbf{92.36 ± 1.02}} & {\color[HTML]{333333} } & \cellcolor[HTML]{C0C0C0}{\color[HTML]{FE0000} \textbf{94.56 ± 0.27}} & \cellcolor[HTML]{C0C0C0}{\color[HTML]{FE0000} \textbf{95.02 ± 0.32}} & \cellcolor[HTML]{C0C0C0}{\color[HTML]{FE0000} \textbf{91.19 ± 1.53}} & \cellcolor[HTML]{C0C0C0}{\color[HTML]{FE0000} \textbf{92.42 ± 1.75}} \\
 & \textbf{Type2} & \cellcolor[HTML]{C0C0C0}{\color[HTML]{333333} 92.58 ± 0.25} & \cellcolor[HTML]{C0C0C0}{\color[HTML]{333333} 93.86 ± 0.02} & \cellcolor[HTML]{C0C0C0}{\color[HTML]{333333} 86.43 ± 2.07} & \cellcolor[HTML]{C0C0C0}{\color[HTML]{333333} 86.81 ± 2.46} & {\color[HTML]{333333} } & 93.32 ± 0.34 & 94.18 ± 0.39 & \cellcolor[HTML]{C0C0C0}{\color[HTML]{333333} 88.31 ± 2.01} & \cellcolor[HTML]{C0C0C0}{\color[HTML]{333333} 89.14 ± 2.01} \\
 & \textbf{Type3} & \cellcolor[HTML]{C0C0C0}{\color[HTML]{FE0000} \textbf{93.72 ± 0.15}} & \cellcolor[HTML]{C0C0C0}{\color[HTML]{FE0000} \textbf{94.99 ± 0.12}} & \cellcolor[HTML]{C0C0C0}{\color[HTML]{FE0000} \textbf{91.27 ± 1.43}} & \cellcolor[HTML]{C0C0C0}{\color[HTML]{333333} 90.94 ± 1.89} & {\color[HTML]{333333} }  & \cellcolor[HTML]{C0C0C0}{\color[HTML]{333333} 94.06 ± 0.28} & \cellcolor[HTML]{C0C0C0}{\color[HTML]{333333} 94.43 ± 0.08} & \cellcolor[HTML]{C0C0C0}{\color[HTML]{333333} 86.54 ± 1.26} & \cellcolor[HTML]{C0C0C0}{\color[HTML]{333333} 90.56 ± 1.09}  \\
 & \textbf{Type4} & 90.68 ± 0.21 & \cellcolor[HTML]{C0C0C0}{\color[HTML]{333333} 94.02 ± 0.09} & \cellcolor[HTML]{C0C0C0}{\color[HTML]{333333} 82.49 ± 2.11} & \cellcolor[HTML]{C0C0C0}{\color[HTML]{333333} 84.38 ± 2.11} & {\color[HTML]{333333} }  & 91.21 ± 0.23 & \cellcolor[HTML]{C0C0C0}{\color[HTML]{333333} 94.39 ± 0.28} & 85.11 ± 1.34 & \cellcolor[HTML]{C0C0C0}{\color[HTML]{333333} 88.15 ± 1.54} \\ 
 & \textbf{Type5} & 91.19 ± 0.13 & \cellcolor[HTML]{C0C0C0}{\color[HTML]{333333} 93.98 ± 0.08} & \cellcolor[HTML]{C0C0C0}{\color[HTML]{333333} 83.62 ± 2.02} & \cellcolor[HTML]{C0C0C0}{\color[HTML]{333333} 83.15 ± 1.67} & {\color[HTML]{333333} } & 91.04 ± 0.31 & 93.78 ± 0.11 & 84.38 ± 1.76 & 85.31 ± 1.76 \\
 & \textbf{Type6} & 91.69 ± 0.16 & 93.57 ± 0.04 & \cellcolor[HTML]{C0C0C0}{\color[HTML]{333333} 83.68 ± 1.25} & \cellcolor[HTML]{C0C0C0}{\color[HTML]{333333} 83.77 ± 2.01} & {\color[HTML]{333333} } & 92.17 ± 0.31 & 93.54 ± 0.13 & 84.16 ± 2.11 & 83.59 ± 2.11 \\
 & \textbf{Type7} & \cellcolor[HTML]{C0C0C0}{\color[HTML]{333333} 92.78 ± 0.08} & 93.09 ± 0.21 & 82.01 ± 1.67 & \cellcolor[HTML]{C0C0C0}{\color[HTML]{333333} 82.17 ± 1.47} & {\color[HTML]{333333} }  & 92.35 ± 0.29 & 92.89 ± 0.25 & 83.04 ± 1.98 & 81.37 ± 1.55  \\
 & \textbf{Type8} & 91.96 ± 0.19 & 92.45 ± 0.19 & 81.98 ± 1.87 & 81.89 ± 1.88 &{\color[HTML]{333333} } & 92.69 ± 0.45 & 92.43 ± 0.19 & 82.71 ± 2.31 & 81.04 ± 1.65 \\
\multirow{-12}{*}{\textbf{GCN}} & \textbf{Type9} & \cellcolor[HTML]{C0C0C0}{\color[HTML]{333333} 92.81 ± 0.05} & 93.66 ± 0.12 & \cellcolor[HTML]{C0C0C0}{\color[HTML]{333333} 84.74 ± 2.56} & \cellcolor[HTML]{C0C0C0}{\color[HTML]{333333} 84.87 ± 2.01} & \multirow{-12}{*}{{\color[HTML]{333333} \textbf{\begin{tabular}[c]{@{}l@{}}GCN-\\ Virtural\end{tabular}}}} & 93.31 ± 0.33 & 93.29 ± 0.21 & 84.35 ± 1.78 & 84.19 ± 1.34  \\ \hline
{\color[HTML]{333333} } & {\color[HTML]{333333} \textbf{No Aug}} & 92.57 ± 0.04 & 93.94 ± 0.26 & 84.07 ± 1.43 & 82.12 ± 1.86 & {\color[HTML]{333333} }  & 91.46 ± 0.66 & 94.06 ± 0.08 & 83.12 ± 1.46 & 84.37 ± 1.28 \\
{\color[HTML]{333333} } & {\color[HTML]{333333} \textbf{MAXPOOL}} & 93.35 ± 0.05 & 94.15 ± 0.03 & 85.11 ± 2.02 & 82.26 ± 1.65 & {\color[HTML]{333333} }  & 92.57 ± 0.36 & 94.46 ± 0.16 & 84.17 ± 2.11 & 85.69 ± 2.13\\
{\color[HTML]{333333} } & {\color[HTML]{333333} \textbf{GMT}} & \cellcolor[HTML]{C0C0C0}{\color[HTML]{333333} 93.41 ± 0.13} & \cellcolor[HTML]{C0C0C0}{\color[HTML]{333333} 94.27 ± 0.18} & 83.14 ± 1.35 & \cellcolor[HTML]{C0C0C0}{\color[HTML]{333333} 83.81 ± 2.09} &{\color[HTML]{333333} } &  92.36 ± 0.49 & 93.01 ± 0.22 & 83.92 ± 1.86 & 85.37 ± 1.24 \\
{\color[HTML]{333333} } & {\color[HTML]{333333} \textbf{Type1}} & 93.13 ± 0.11 & \cellcolor[HTML]{C0C0C0}{\color[HTML]{FE0000} \textbf{95.11 ± 0.08}} & \cellcolor[HTML]{C0C0C0}{\color[HTML]{333333} 85.65 ± 2.07} & \cellcolor[HTML]{C0C0C0}{\color[HTML]{333333} 87.92 ± 1.83} & {\color[HTML]{333333} }  & \cellcolor[HTML]{C0C0C0}{\color[HTML]{FE0000} \textbf{93.08 ± 0.45}} & \cellcolor[HTML]{C0C0C0}{\color[HTML]{FE0000} \textbf{95.37 ± 0.12}} & \cellcolor[HTML]{C0C0C0}{\color[HTML]{FE0000} \textbf{90.89 ± 1.22}} & \cellcolor[HTML]{C0C0C0}{\color[HTML]{333333} 91.03 ± 2.19} \\
{\color[HTML]{333333} } & {\color[HTML]{333333} \textbf{Type2}} & 93.11 ± 0.17 & \cellcolor[HTML]{C0C0C0}{\color[HTML]{333333} 94.31 ± 0.13} & 85.01 ± 1.65 & \cellcolor[HTML]{C0C0C0}{\color[HTML]{333333} 88.38 ± 1.76} &{\color[HTML]{333333} }  & 92.47 ± 0.23 & \cellcolor[HTML]{C0C0C0}{\color[HTML]{333333} 94.89 ± 0.14} & \cellcolor[HTML]{C0C0C0}{\color[HTML]{333333} 87.13 ± 2.01} & \cellcolor[HTML]{C0C0C0}{\color[HTML]{333333} 89.87 ± 1.22}\\
{\color[HTML]{333333} } & {\color[HTML]{333333} \textbf{Type3}} & \cellcolor[HTML]{C0C0C0}{\color[HTML]{333333} 93.67 ± 0.08} & \cellcolor[HTML]{C0C0C0}{\color[HTML]{333333} 94.73 ± 0.11} & \cellcolor[HTML]{C0C0C0}{\color[HTML]{333333} 85.72 ± 2.11} & \cellcolor[HTML]{C0C0C0}{\color[HTML]{FE0000} \textbf{92.47 ± 1.38}} & {\color[HTML]{333333} }  & \cellcolor[HTML]{C0C0C0}{\color[HTML]{333333} 92.98 ± 0.28} & \cellcolor[HTML]{C0C0C0}{\color[HTML]{333333} 95.02 ± 0.09} & \cellcolor[HTML]{C0C0C0}{\color[HTML]{333333} 85.92 ± 2.12} & \cellcolor[HTML]{C0C0C0}{\color[HTML]{333333} 90.11 ± 1.76}\\
{\color[HTML]{333333} } & {\color[HTML]{333333} \textbf{Type4}} & 93.21 ± 0.19 & 93.97 ± 0.09 & 84.83 ± 1.27 & \cellcolor[HTML]{C0C0C0}{\color[HTML]{333333} 86.71 ± 2.06} &{\color[HTML]{333333} }  & 92.03 ± 0.14 & 93.88 ± 0.17 & 84.32 ± 1.23 & \cellcolor[HTML]{C0C0C0}{\color[HTML]{333333} 86.37 ± 2.18}\\
{\color[HTML]{333333} } & {\color[HTML]{333333} \textbf{Type5}} & \cellcolor[HTML]{C0C0C0}{\color[HTML]{333333} 93.42 ± 0.04} & 93.12 ± 0.15 & 84.66 ± 2.01 & \cellcolor[HTML]{C0C0C0}{\color[HTML]{333333} 84.32 ± 2.11} & {\color[HTML]{333333} } &  91.78 ± 0.65 & 93.43 ± 0.21 & 83.27 ± 1.39 & 85.43 ± 1.23  \\
{\color[HTML]{333333} } & {\color[HTML]{333333} \textbf{Type6}} & \cellcolor[HTML]{C0C0C0}{\color[HTML]{FE0000} \textbf{94.81 ± 0.12}} & \cellcolor[HTML]{C0C0C0}{\color[HTML]{333333} 94.22 ± 0.12} & \cellcolor[HTML]{C0C0C0}{\color[HTML]{FE0000} \textbf{91.43 ± 1.33}} & \cellcolor[HTML]{C0C0C0}{\color[HTML]{333333} 85.49 ± 1.68} & {\color[HTML]{333333} }  & 92.47 ± 0.28 & 94.35 ± 0.11 & 83.09 ± 2.87 & 84.67 ± 2.17 \\
{\color[HTML]{333333} } & {\color[HTML]{333333} \textbf{Type7}} & 93.14 ± 0.15 & 93.73 ± 0.05 & 82.89 ± 1.32 & \cellcolor[HTML]{C0C0C0}{\color[HTML]{333333} 82.39 ± 1.88} &{\color[HTML]{333333} } & 92.51 ± 0.12 & 92.79 ± 0.19 & 82.37 ± 2.18 & 83.03 ± 1.75\\
{\color[HTML]{333333} } & {\color[HTML]{333333} \textbf{Type8}} & 92.08 ± 0.22 & 93.42 ± 0.14 & 81.31 ± 1.43 & 81.57 ± 2.02 &{\color[HTML]{333333} } & 91.98 ± 0.29 & 92.13 ± 0.21 & 82.11 ± 1.77 & 82.34 ± 1.58 \\
\multirow{-12}{*}{{\color[HTML]{333333} \textbf{GIN}}} & {\color[HTML]{333333} \textbf{Type9}} & \cellcolor[HTML]{C0C0C0}{\color[HTML]{333333} 93.63 ± 0.07} & 93.89 ± 0.06 & 83.47 ± 2.13 & \cellcolor[HTML]{C0C0C0}{\color[HTML]{333333} 83.59 ± 1.22} & \multirow{-12}{*}{{\color[HTML]{333333} \textbf{\begin{tabular}[c]{@{}l@{}} GIN\\ -Virtual\end{tabular}}}} & 92.39 ± 0.39 & 93.43 ± 0.23 & 83.47 ± 1.08 & \cellcolor[HTML]{C0C0C0}{\color[HTML]{FE0000} \textbf{92.58 ± 1.66}} \\ \hline
\end{tabular}
}
\end{table}

\begin{table*}[!tb]
\caption{Results of statistical testing on Accuracy and Robustness. \textbf{True}: indicates that the comparison is statistically significant after the  \emph{Bonferroni correction} adjustment. Statistical testing methods: \emph{Wilcoxon signed-rank test}. \textbf{GNNs(PL)}: represents the statistical testing conducted on all GNN models including GCN, GIN, GCN-Virtual, and GIN-Virtual. A gray background highlights the result marked as \textbf{True}.}
\label{table:statistical_test_accuracy_robustness}
\centering
\scalebox{0.55}{
\begin{tabular}{lccc|ccc|ccc|ccc|ccc} \hline
\textbf{} & \multicolumn{15}{c}{\textbf{Test Accuracy}} \\ \hline
      & \multicolumn{3}{c|}{GCN (NLP)} & \multicolumn{3}{c|}{GAT (NLP)}   & \multicolumn{3}{c|}{ GraphSAGE (NLP)} & \multicolumn{3}{c|}{ GIN (NLP)}  & \multicolumn{3}{c}{ GNNs (PL)}\\
      & No Aug & MAXPOOL & GMT   & No Aug & MAXPOOL & GMT & No Aug & MAXPOOL & GMT & No Aug & MAXPOOL & GMT& No Aug & MAXPOOL & GMT  \\ \hline 
Type1 & \cellcolor[HTML]{C0C0C0}{\color[HTML]{000000}True }  & \cellcolor[HTML]{C0C0C0}{\color[HTML]{000000}True }   & \cellcolor[HTML]{C0C0C0}{\color[HTML]{000000}True } & \cellcolor[HTML]{C0C0C0}{\color[HTML]{000000}True }   & \cellcolor[HTML]{C0C0C0}{\color[HTML]{000000}True }    & \cellcolor[HTML]{C0C0C0}{\color[HTML]{000000}True } & \cellcolor[HTML]{C0C0C0}{\color[HTML]{000000}True }      & \cellcolor[HTML]{C0C0C0}{\color[HTML]{000000}True }       & \cellcolor[HTML]{C0C0C0}{\color[HTML]{000000}True } & \cellcolor[HTML]{C0C0C0}{\color[HTML]{000000}True }      & \cellcolor[HTML]{C0C0C0}{\color[HTML]{000000}True }       & \cellcolor[HTML]{C0C0C0}{\color[HTML]{000000}True } &  \cellcolor[HTML]{C0C0C0}{\color[HTML]{000000}True }  & \cellcolor[HTML]{C0C0C0}{\color[HTML]{000000}True }  &  \cellcolor[HTML]{C0C0C0}{\color[HTML]{000000}True }  \\
Type2 & \cellcolor[HTML]{C0C0C0}{\color[HTML]{000000}True }  & -       & -    & \cellcolor[HTML]{C0C0C0}{\color[HTML]{000000}True }   & -       & -    & -      & -       & -          & \cellcolor[HTML]{C0C0C0}{\color[HTML]{000000}True }      & -       & \cellcolor[HTML]{C0C0C0}{\color[HTML]{000000}True } &\cellcolor[HTML]{C0C0C0}{\color[HTML]{000000}True }  & - & - \\
Type3 & -      & -       & -    & \cellcolor[HTML]{C0C0C0}{\color[HTML]{000000}True }   & -       & \cellcolor[HTML]{C0C0C0}{\color[HTML]{000000}True } & \cellcolor[HTML]{C0C0C0}{\color[HTML]{000000}True }      & -       & \cellcolor[HTML]{C0C0C0}{\color[HTML]{000000}True }    & \cellcolor[HTML]{C0C0C0}{\color[HTML]{000000}True }      & \cellcolor[HTML]{C0C0C0}{\color[HTML]{000000}True }       & \cellcolor[HTML]{C0C0C0}{\color[HTML]{000000}True } & \cellcolor[HTML]{C0C0C0}{\color[HTML]{000000}True }  &  \cellcolor[HTML]{C0C0C0}{\color[HTML]{000000}True }  \\
Type4 & -      & -       & -    & -      & \cellcolor[HTML]{C0C0C0}{\color[HTML]{000000}True }    & \cellcolor[HTML]{C0C0C0}{\color[HTML]{000000}True } & -      & -       & -          & -      & -       & - & -      & \cellcolor[HTML]{C0C0C0}{\color[HTML]{000000}True }      & - \\
Type5 & -      & -       & -    & -      & \cellcolor[HTML]{C0C0C0}{\color[HTML]{000000}True }     & -   & -      & \cellcolor[HTML]{C0C0C0}{\color[HTML]{000000}True }       & -       & -      & -       & - & -      & \cellcolor[HTML]{C0C0C0}{\color[HTML]{000000}True }        & \cellcolor[HTML]{C0C0C0}{\color[HTML]{000000}True }  \\
Type6 & -      & -       & -    & -      & -       & -    & -      & \cellcolor[HTML]{C0C0C0}{\color[HTML]{000000}True }       & -       & \cellcolor[HTML]{C0C0C0}{\color[HTML]{000000}True }      & -       & - & -       & - & -       \\
Type7 & -      & -       & -    & -      & -       & -    & \cellcolor[HTML]{C0C0C0}{\color[HTML]{000000}True }   & \cellcolor[HTML]{C0C0C0}{\color[HTML]{000000}True }      & \cellcolor[HTML]{C0C0C0}{\color[HTML]{000000}True }     & -      & -       & - & -      & \cellcolor[HTML]{C0C0C0}{\color[HTML]{000000}True }        & \cellcolor[HTML]{C0C0C0}{\color[HTML]{000000}True }  \\
Type8 & -      & \cellcolor[HTML]{C0C0C0}{\color[HTML]{000000}True }   & \cellcolor[HTML]{C0C0C0}{\color[HTML]{000000}True } & -      & \cellcolor[HTML]{C0C0C0}{\color[HTML]{000000}True }    & -    & -      & \cellcolor[HTML]{C0C0C0}{\color[HTML]{000000}True }       & -       & -      & \cellcolor[HTML]{C0C0C0}{\color[HTML]{000000}True }       & \cellcolor[HTML]{C0C0C0}{\color[HTML]{000000}True } & \cellcolor[HTML]{C0C0C0}{\color[HTML]{000000}True }  &  \cellcolor[HTML]{C0C0C0}{\color[HTML]{000000}True }  &  \cellcolor[HTML]{C0C0C0}{\color[HTML]{000000}True }  \\
Type9 & \cellcolor[HTML]{C0C0C0}{\color[HTML]{000000}True }  & -       & -   & -      & -       & -     & -      & \cellcolor[HTML]{C0C0C0}{\color[HTML]{000000}True }      & -        & -      & -       & -  & -      & -       & - \\    \hline

\textbf{} & \multicolumn{15}{c}{\textbf{Robustness}} \\ \hline
Type1 &\cellcolor[HTML]{C0C0C0}{\color[HTML]{000000}True }   & -       & -   & \cellcolor[HTML]{C0C0C0}{\color[HTML]{000000}True }       & \cellcolor[HTML]{C0C0C0}{\color[HTML]{000000}True }        & \cellcolor[HTML]{C0C0C0}{\color[HTML]{000000}True }     & -      & -     & \cellcolor[HTML]{C0C0C0}{\color[HTML]{000000}True }         & \cellcolor[HTML]{C0C0C0}{\color[HTML]{000000}True }         & \cellcolor[HTML]{C0C0C0}{\color[HTML]{000000}True }           & -  & \cellcolor[HTML]{C0C0C0}{\color[HTML]{000000}True }     & \cellcolor[HTML]{C0C0C0}{\color[HTML]{000000}True }       & \cellcolor[HTML]{C0C0C0}{\color[HTML]{000000}True }\\
Type2 & -  & -       & -   & -      & -       & \cellcolor[HTML]{C0C0C0}{\color[HTML]{000000}True }      & -      & -     & \cellcolor[HTML]{C0C0C0}{\color[HTML]{000000}True }         & -      & -       & -  & \cellcolor[HTML]{C0C0C0}{\color[HTML]{000000}True }      & \cellcolor[HTML]{C0C0C0}{\color[HTML]{000000}True }       & \cellcolor[HTML]{C0C0C0}{\color[HTML]{000000}True } \\
Type3 & -  & -       & -   & -      & -       & \cellcolor[HTML]{C0C0C0}{\color[HTML]{000000}True }     & \cellcolor[HTML]{C0C0C0}{\color[HTML]{000000}True }      & -     & \cellcolor[HTML]{C0C0C0}{\color[HTML]{000000}True }       & -      & -       & -  & \cellcolor[HTML]{C0C0C0}{\color[HTML]{000000}True }      &\cellcolor[HTML]{C0C0C0}{\color[HTML]{000000}True }     & \cellcolor[HTML]{C0C0C0}{\color[HTML]{000000}True }\\
Type4 & -  & -       & \cellcolor[HTML]{C0C0C0}{\color[HTML]{000000}True }   & \cellcolor[HTML]{C0C0C0}{\color[HTML]{000000}True }       & \cellcolor[HTML]{C0C0C0}{\color[HTML]{000000}True }       & -     & -      & -     & -        & -      & -       & -  & \cellcolor[HTML]{C0C0C0}{\color[HTML]{000000}True }      & -       & \cellcolor[HTML]{C0C0C0}{\color[HTML]{000000}True }\\
Type5 & -  & -       & -   & \cellcolor[HTML]{C0C0C0}{\color[HTML]{000000}True }       & \cellcolor[HTML]{C0C0C0}{\color[HTML]{000000}True }        & \cellcolor[HTML]{C0C0C0}{\color[HTML]{000000}True }      & -      & \cellcolor[HTML]{C0C0C0}{\color[HTML]{000000}True }     &\cellcolor[HTML]{C0C0C0}{\color[HTML]{000000}True }        & \cellcolor[HTML]{C0C0C0}{\color[HTML]{000000}True }          & -       & -  & -      & -       & \cellcolor[HTML]{C0C0C0}{\color[HTML]{000000}True } \\
Type6 & -  & -       & -   & \cellcolor[HTML]{C0C0C0}{\color[HTML]{000000}True }       & \cellcolor[HTML]{C0C0C0}{\color[HTML]{000000}True }        & \cellcolor[HTML]{C0C0C0}{\color[HTML]{000000}True }      & -      & -     & \cellcolor[HTML]{C0C0C0}{\color[HTML]{000000}True }        & \cellcolor[HTML]{C0C0C0}{\color[HTML]{000000}True }         & -       & -  & -      & -       & - \\
Type7 & -  & -       & -   & -      & \cellcolor[HTML]{C0C0C0}{\color[HTML]{000000}True }        & -     & \cellcolor[HTML]{C0C0C0}{\color[HTML]{000000}True }       & \cellcolor[HTML]{C0C0C0}{\color[HTML]{000000}True }     & -        & -      & -          & \cellcolor[HTML]{C0C0C0}{\color[HTML]{000000}True }      & \cellcolor[HTML]{C0C0C0}{\color[HTML]{000000}True }      & \cellcolor[HTML]{C0C0C0}{\color[HTML]{000000}True }       & \cellcolor[HTML]{C0C0C0}{\color[HTML]{000000}True } \\
Type8 & -  & -       & -   & \cellcolor[HTML]{C0C0C0}{\color[HTML]{000000}True }       & \cellcolor[HTML]{C0C0C0}{\color[HTML]{000000}True }        & -     & -      & \cellcolor[HTML]{C0C0C0}{\color[HTML]{000000}True }      & -        & -         & \cellcolor[HTML]{C0C0C0}{\color[HTML]{000000}True }           & -  &\cellcolor[HTML]{C0C0C0}{\color[HTML]{000000}True }     & \cellcolor[HTML]{C0C0C0}{\color[HTML]{000000}True }      & \cellcolor[HTML]{C0C0C0}{\color[HTML]{000000}True } \\
Type9 & -  & -       & \cellcolor[HTML]{C0C0C0}{\color[HTML]{000000}True }    & -      & -       & \cellcolor[HTML]{C0C0C0}{\color[HTML]{000000}True }      & -      & -     & \cellcolor[HTML]{C0C0C0}{\color[HTML]{000000}True }         & -      & -       & \cellcolor[HTML]{C0C0C0}{\color[HTML]{000000}True }     & -      & -       & - \\    \hline
\end{tabular}}
\end{table*}

The right part in Table~\ref{table:Pool_gat} shows the result of the GAT model. Surprisingly, all of the cases show that GMT can be unable to help \emph{Manifold-Mixup} improve the robustness compared to MAXPOOL. In contrast, 4 out of the 9 hybrid operators on the Gossipcop-Profile and 3 out of the 9 hybrid operators on the Politifact-Bert are able to achieve better performance than MAXPOOL. In particular, the operator Type 3 improves the performance of robustness by up to 19.02\% on the Gossipcop-Spacy compared to GMT. The operator Type 1 outperforms GMT by up to 7.33\% on the Politifact-Spacy. However, the results from the column Average show that the operator Type 1 is still the optimal choice for GAT in robustness improvement. We focus on Type 1 and Type 3 for the analysis of average robustness improvement, as these two hybrid pooling operators outperform others, as indicated by the Average column. Type 1 exhibits a higher average robustness improvement of 0.86\% compared to Type 3 with 0.41\% when compared to MAXPOOL. Compared to GMT, we can reach a similar conclusion that Type 1 exhibits the best improvement in GAT robustness, with an average improvement of 7.57\%, whereas Type 3 only shows an average improvement of 5.67\%.

The right part in Table~\ref{table:Pool_sage} presents the robustness results of the GraphSAGE model. Similar to the finding from Table~\ref{table:Pool_gat}, GMT is unable to help \emph{Manifold-Mixup} in robustness improvement. However, all hybrid operators are effective in improving the robustness compared to MAXPOOL on the Gossipcop-Profile. Compared to GMT, the operator Type 1 improves the robustness by up to 11.87\% on the Gossipcop-Profile and 14.25\% on the Politifact-Spacy.  Similar to GAT, we continue to focus on Type 1 and Type 3 as these two hybrid pooling operators demonstrate the best performance compared to others, as indicated by the Average column. When compared to MAXPOOL, Type 1 demonstrates a superior average robustness improvement of 1.16\%, while Type 3 exhibits an average improvement of 0.73\%. Different from the comparison in MAXPOOL,  Type 3 exhibits a superior average robustness improvement of 8.71\% compared to Type 1, which achieves 7.19\%, when considering GMT.

The right part in Table~\ref{table:Pool_gin} presents the results of the GIN model. Maybe surprisingly, the gap between MAXPOOL and GMT can reach 31.69\% (Politifact-Content) in terms of robustness. Moving to robustness improvement for GIN training,  the operator Type 6 outperforms GMT by up to 23.23\% on the Gossipcop-Profile. Similarly, the Type 5 operator achieves an improvement of up to 16.49\% compared to GMT on the Politifact-Spacy. We analyze the average robustness improvement of Type 1 and Type 9 as they demonstrate superior performance compared to others, based on the Average column. Type 9 exhibits a superior average robustness improvement of 6.74\% compared to MAXPOOL, while Type 1 achieves a lower average improvement of 5.41\%. However, moving to GMT, Type 1 demonstrates a better average robustness improvement of 10.52\% compared to Type 9, which only achieves 6.36\%.

Table \ref{table:results2} presents the results of JAVA250 and Python800. Hybrid pooling operators are more effective in improving the robustness of GNN models than GMT and MAXPOOL. Notably, the operator Type 1 achieves the highest performance in 4 out of 8 cases. Moving to the gap between hybrid operators and GMT, the operator Type 1 can outperform GMT by up to 10.32\% in the robustness improvement (GCN, Python800, Type 1). We analyze the average robustness improvement in the PL dataset using Type 1, Type 3, and Type 9. These three hybrid pooling operators are selected based on their superior performance compared to others, as indicated by the results from the Average column. When compared to MAXPOOL, Type 1 exhibits the highest average robustness improvement at 5.55\%, followed by Type 3 at 4.96\% and Type 9 at 1.62\%. Moving to GMT, Type 1 still shows the highest improvement at 6.30\%, followed by Type 3 at 5.70\% and Type 9 at 1.75\%.

\textbf{Statistical analysis.} We also conduct statistical analysis to check the significance of the robustness improvement by using hybrid pooling. The lower part in Table~\ref{table:statistical_test_accuracy_robustness} presents the results. We can see that similar to the results of test accuracy, Type 1 significantly outperforms No Aug, MAXPOOL, and GMT in 10 out of 15 cases. This indicates that Type 1 is a relatively stable way to enhance the robustness of GNN models. Besides, we find that most hybrid pooling operators show better robustness improvement on GAT in NLP and all GNNs in PL, compared to No Aug, MAXPOOL, and GMT.  

Overall, based on the analysis of NLP and PL datasets, we find that hybrid pooling operators can indeed improve the robustness of GNNs, whether compared to MAXPOOL or GMT, across all GNNs. We comprehensively assess the performance of all hybrid operators across these three evaluation metrics based on maximum improvement, average improvement, and statistical significance. Our analysis concludes that Type 1 exhibits the highest effectiveness, closely followed by Type 3.

\noindent\colorbox{gray!20}{\framebox{\parbox{0.96\linewidth}{
\textbf{Answer to RQ2}: Hybrid pooling operators can enhance the robustness of GNN models. Among our studied operators, based on the analysis of improvement, average enhancement, and statistical significance, we recommend Type 1 and Type 3 for improving the robustness of GNNs. Concretely, the operator Type 1~($\mathcal{M}_{sum}(\mathcal{P}_{att},\mathcal{P}_{max})$) improves the robustness by 10.23\% compared to GMT on the PL dataset. }}}

%\noindent\colorbox{gray!20}{\framebox{\parbox{0.96\linewidth}{
%\textbf{Answer to RQ2}: Hybrid pooling operators can enhance the robustness of GNN models. Among these operators, the Type 6 operator~($\mathcal{M}_{concat}(\mathcal{P}_{att},\mathcal{P}_{sum})$) is the best one for the NLP dataset and outperforms GMT by up to 23.23\% robustness. Additionally, the operator Type 1~($\mathcal{M}_{sum}(\mathcal{P}_{att},\mathcal{P}_{max})$) improves the robustness by 10.23\% compared to GMT on the PL dataset. }}}

\subsection{RQ3: How does the hyperparameter setting affect the effectiveness of Manifold-Mixup when hybrid pooling operators are applied?}
\label{sec:alpha}

From the last part, we know that hybrid pooling operators are more helpful for \emph{Manifold-Mixup} when dealing with graph-structured data. However, the potential influence of  $\alpha$ that is introduced in Section~\ref{mixup}, which is the key hyperparameter in the Mixup technique~\cite{zhang2018mixup, dong2023mixcode, verma2019manifold, yun2019cutmix}, on the effectiveness of the Mixup training is still unclear. In this part, we tend to explore such influence using the dataset Politifact with Profile node embedding and set the $\alpha$ from 0.05 to 0.5 in 0.05 intervals.

\begin{table*}[!tb]
\caption{Results (average $\pm$ standard deviation, \%) of trained GCN models with \emph{Manifold-Mixup} (dataset: Politifact) using hybrid pooling operators with different $\alpha$ settings. The best results are marked in red color.}
\label{table:results3}
\centering
\scalebox{0.55}{
\begin{tabular}{lcccccccccccc}
\hline
\textbf{} & \multicolumn{11}{c}{\textbf{Test Accuracy}} & \multicolumn{1}{l}{} \\ \hline
\textbf{Mixup Ratio Alpha} & \textbf{0.05} & \textbf{0.10} & \textbf{0.15} & \textbf{0.20} & \textbf{0.25} & \textbf{0.30} & \textbf{0.35} & \textbf{0.40} & \textbf{0.45} & \textbf{0.50} & {\color[HTML]{343434} \textbf{Average}} & \textbf{Std} \\ \hline
\textbf{Type1} & \multicolumn{1}{l}{79.07 ± 0.57} & \multicolumn{1}{l}{78.96 ± 1.96} & \multicolumn{1}{l}{77.01 ± 1.21} & \multicolumn{1}{l}{78.73 ± 0.46} & \multicolumn{1}{l}{78.28 ± 0.45} & \multicolumn{1}{l}{79.49 ± 0.94} & \multicolumn{1}{l}{77.83 ± 0.91} & \multicolumn{1}{l}{78.28 ± 1.57} & \multicolumn{1}{l}{77.68 ± 0.26} & \multicolumn{1}{l}{75.87 ± 1.45} & 78.12 & \multicolumn{1}{r}{1.08} \\
\textbf{Type2} & \multicolumn{1}{l}{75.79 ± 0.32} & \multicolumn{1}{l}{77.83 ± 1.64} & \multicolumn{1}{l}{74.36 ± 1.14} & \multicolumn{1}{l}{75.57 ± 2.39} & \multicolumn{1}{l}{76.32 ± 1.14} & \multicolumn{1}{l}{75.42 ± 2.09} & \multicolumn{1}{l}{76.17 ± 0.52} & \multicolumn{1}{l}{77.06 ± 1.31} & \multicolumn{1}{l}{78.13 ± 0.69} & \multicolumn{1}{l}{75.11 ± 0.91} & 76.18 & \multicolumn{1}{r}{1.20} \\
\textbf{Type3} & \multicolumn{1}{l}{76.02 ± 1.28} & \multicolumn{1}{l}{78.73 ± 2.12} & \multicolumn{1}{l}{77.38 ± 2.39} & \multicolumn{1}{l}{76.77 ± 1.83} & \multicolumn{1}{l}{77.22 ± 1.88} & \multicolumn{1}{l}{73.75 ± 3.59} & \multicolumn{1}{l}{78.58 ± 1.38} & \multicolumn{1}{l}{73.15 ± 2.57} & \multicolumn{1}{l}{76.47 ± 2.75} & \multicolumn{1}{l}{77.38 ± 0.92} & 76.55 & \multicolumn{1}{r}{1.84} \\
\textbf{Type4} & \multicolumn{1}{l}{77.15 ± 0.31} & \multicolumn{1}{l}{77.22 ± 0.94} & \multicolumn{1}{l}{76.92 ± 0.46} & \multicolumn{1}{l}{76.32 ± 0.94} & \multicolumn{1}{l}{77.52 ± 1.04} & \multicolumn{1}{l}{77.53 ± 0.26} & \multicolumn{1}{l}{77.98 ± 0.69} & \multicolumn{1}{l}{77.38 ± 0.91} & \multicolumn{1}{l}{76.92 ± 0.91} & \multicolumn{1}{l}{76.17 ± 1.31} & 77.11 & \multicolumn{1}{r}{{\color[HTML]{FE0000} \textbf{0.55}}} \\
\textbf{Type5} & \multicolumn{1}{l}{77.22 ± 1.38} & \multicolumn{1}{l}{77.38 ± 0.46} & \multicolumn{1}{l}{77.53 ± 1.13} & \multicolumn{1}{l}{77.32 ± 1.73} & \multicolumn{1}{l}{76.17 ± 3.77} & \multicolumn{1}{l}{73.61 ± 1.14} & \multicolumn{1}{l}{76.02 ± 2.75} & \multicolumn{1}{l}{76.47 ± 0.45} & \multicolumn{1}{l}{75.26 ± 2.65} & \multicolumn{1}{l}{77.38 ± 2.75} & 76.44 & \multicolumn{1}{r}{1.25} \\
\textbf{Type6} & \multicolumn{1}{l}{78.06 ± 0.95} & \multicolumn{1}{l}{77.83 ± 1.19} & \multicolumn{1}{l}{75.72 ± 0.26} & \multicolumn{1}{l}{76.17 ± 0.26} & \multicolumn{1}{l}{76.47 ± 0.91} & \multicolumn{1}{l}{75.87 ± 1.16} & \multicolumn{1}{l}{75.56 ± 1.19} & \multicolumn{1}{l}{75.56 ± 0.78} & \multicolumn{1}{l}{77.22 ± 0.94} & \multicolumn{1}{l}{76.92 ± 0.89} & 76.54 & \multicolumn{1}{r}{0.93} \\
\textbf{Type7} & \multicolumn{1}{l}{77.83 ± 1.28} & \multicolumn{1}{l}{77.08 ± 0.53} & \multicolumn{1}{l}{77.07 ± 1.14} & \multicolumn{1}{l}{77.98 ± 0.94} & \multicolumn{1}{l}{77.61 ± 1.41} & \multicolumn{1}{l}{76.47 ± 1.21} & \multicolumn{1}{l}{77.08 ± 0.53} & \multicolumn{1}{l}{77.98 ± 1.38} & \multicolumn{1}{l}{77.68 ± 0.26} & \multicolumn{1}{l}{75.72 ± 2.49} & 77.25 & \multicolumn{1}{r}{0.73} \\
\textbf{Type8} & \multicolumn{1}{l}{78.84 ± 0.77} & \multicolumn{1}{l}{78.17 ± 1.49} & \multicolumn{1}{l}{79.64 ± 0.64} & \multicolumn{1}{l}{77.37 ± 1.99} & \multicolumn{1}{l}{79.34 ± 1.13} & \multicolumn{1}{l}{79.78 ± 0.94} & \multicolumn{1}{l}{79.04 ± 1.45} & \multicolumn{1}{l}{79.34 ± 0.26} & \multicolumn{1}{l}{78.58 ± 0.92} & \multicolumn{1}{l}{79.79 ± 1.31} & {\color[HTML]{FE0000} \textbf{78.99}} & \multicolumn{1}{r}{0.78} \\
\textbf{Type9} & \multicolumn{1}{l}{77.38 ± 0.64} & \multicolumn{1}{l}{78.28 ± 1.36} & \multicolumn{1}{l}{77.08 ± 0.14} & \multicolumn{1}{l}{78.28 ± 0.45} & \multicolumn{1}{l}{77.98 ± 0.94} & \multicolumn{1}{l}{77.83 ± 1.21} & \multicolumn{1}{l}{78.58 ± 0.69} & \multicolumn{1}{l}{77.98 ± 0.94} & \multicolumn{1}{l}{77.53 ± 1.13} & \multicolumn{1}{l}{78.43 ± 0.69} & 77.94 & \multicolumn{1}{r}{0.49} \\
\textbf{Average} & 77.48 & {\color[HTML]{FE0000} \textbf{77.94}} & 76.97 & 77.17 & 77.43 & 76.64 & 77.43 & 77.02 & 77.27 & 76.97 & 77.23 & 0.98 \\ \hline
 & \multicolumn{11}{c}{\textbf{Robustness}} & \multicolumn{1}{l}{} \\ \hline
\textbf{Type1} & 70.78 ± 0.45 & 69.01 ± 0.32 & 68.33 ± 0.21 & 68.94 ± 0.69 & 68.63 ± 0.93 & 68.74 ± 0.45 & 67.82 ± 0.54 & 67.91 ± 0.34 & 67.58 ± 0.54 & 67.23 ± 1.55 & 68.50 & 1.00 \\
\textbf{Type2} & 68.97 ± 0.65 & 68.55 ± 0.96 & 68.21 ± 1.34 & 68.44 ± 1.29 & 69.01 ± 0.74 & 68.96 ± 1.99 & 68.97 ± 0.16 & 68.96 ± 0.91 & 68.98 ± 0.86 & 68.65 ± 0.86 & 68.77 & 0.29 \\
\textbf{Type3} & 73.04 ± 0.55 & 72.85 ± 0.63 & 71.34 ± 1.23 & 71.03 ± 0.63 & 71.28 ± 0.58 & 70.35 ± 2.39 & 71.98 ± 2.58 & 70.25 ± 1.34 & 70.64 ± 1.45 & 70.32 ± 1.23 & 71.31 & 1.02 \\
\textbf{Type4} & 74.54 ± 1.32 & 74.43 ± 1.61 & 74.12 ± 1.56 & 74.01 ± 1.45 & 74.86 ± 0.94 & 74.59 ± 1.16 & 74.62 ± 0.94 & 74.28 ± 0.51 & 74.03 ± 3.21 & 73.99 ± 2.11 & 74.35 & {\color[HTML]{FE0000} \textbf{0.31}} \\
\textbf{Type5} & 72.57 ± 0.87 & 71.71 ± 2.24 & 71.75 ± 0.98 & 71.61 ± 0.76 & 70.97 ± 2.43 & 70.01 ± 0.54 & 70.99 ± 1.45 & 70.78 ± 0.94 & 70.16 ± 1.05 & 70.19 ± 1.87 & 71.07 & 0.83 \\
\textbf{Type6} & 77.13 ± 1.03 & 75.79 ± 0.96 & 75.01 ± 0.87 & 75.33 ± 1.44 & 75.41 ± 1.92 & 75.02 ± 0.67 & 75.01 ± 0.55 & 74.98 ± 1.38 & 75.22 ± 1.24 & 75.11 ± 0.65 & 75.40 & 0.66 \\
\textbf{Type7} & 71.87 ± 0.83 & 71.49 ± 0.64 & 71.37 ± 0.56 & 71.89 ± 0.86 & 71.81 ± 1.46 & 71.45 ± 1.96 & 71.01 ± 0.69 & 71.01 ± 2.31 & 70.78 ± 0.06 & 71.56 ± 1.66 & 71.42 & 0.39 \\
\textbf{Type8} & 68.55 ± 0.87 & 68.34 ± 0.87 & 70.23 ± 1.24 & 69.87 ± 1.79 & 70.12 ± 2.08 & 70.45 ± 0.45 & 70.08 ± 3.02 & 70.11 ± 1.45 & 70.01 ± 0.91 & 70.12 ± 0.86 & 69.79 & 0.72 \\
\textbf{Type9} & 76.17 ± 0.32 & 76.91 ± 1.27 & 76.08 ± 1.44 & 76.45 ± 0.95 & 76.29 ± 1.49 & 76.13 ± 1.81 & 76.58 ± 2.09 & 76.21 ± 1.45 & 76.03 ± 1.23 & 75.31 ± 2.01 & {\color[HTML]{FE0000} \textbf{76.22}} & 0.42 \\
\textbf{Average} & {\color[HTML]{FE0000} \textbf{72.62}} & {\color[HTML]{343434} 72.12} & {\color[HTML]{343434} 71.83} & {\color[HTML]{343434} 71.95} & {\color[HTML]{343434} 72.04} & {\color[HTML]{343434} 71.74} & {\color[HTML]{343434} 71.90} & {\color[HTML]{343434} 71.61} & {\color[HTML]{343434} 71.49} & {\color[HTML]{343434} 71.39} & 71.87 & 0.63 \\ \hline
\end{tabular}
}
\end{table*}

\label{sec:apppendix_alpha}
\begin{table*}[!tb]
\caption{Results (average $\pm$ standard deviation, \%) of trained GAT models with \emph{Manifold-Mixup} (dataset: Politifact) using hybrid pooling operators with different $\alpha$ settings. The best results are marked in red color.}
\label{table:alpha_GAT}
\centering
\scalebox{0.55}{
\begin{tabular}{lcccccccccccc}
\hline
\textbf{} & \multicolumn{11}{c}{\textbf{Test Accuracy}} & \multicolumn{1}{l}{} \\ \hline
\textbf{Mixup Ratio Alpha} & \multicolumn{1}{c}{\textbf{0.05}} & \multicolumn{1}{c}{\textbf{0.10}} & \multicolumn{1}{c}{\textbf{0.15}} & \multicolumn{1}{c}{\textbf{0.20}} & \multicolumn{1}{c}{\textbf{0.25}} & \multicolumn{1}{c}{\textbf{0.30}} & \multicolumn{1}{c}{\textbf{0.35}} & \multicolumn{1}{c}{\textbf{0.40}} & \multicolumn{1}{c}{\textbf{0.45}} & \multicolumn{1}{c}{\textbf{0.50}} & {\color[HTML]{343434} \textbf{Average}} & \textbf{Std} \\ \hline
\textbf{Type1} & 76.32 ± 1.72 & 75.11 ± 0.66 & 75.27 ± 0.94 & 76.47 ± 1.36 & 76.02 ± 0.45 & 75.11 ± 0.79 & 75.26 ± 0.69 & 76.17 ± 1.05 & 74.96 ± 1.38 & 77.23 ± 1.14 & {\color[HTML]{000000} 75.79} & {\color[HTML]{000000} 0.76} \\
\textbf{Type2} & 76.77 ± 0.94 & 76.17 ± 0.69 & 76.17 ± 1.83 & 76.47 ± 1.21 & 74.96 ± 2.23 & 74.51 ± 0.94 & 76.17 ± 2.27 & 77.22 ± 1.31 & 76.47 ± 1.57 & 75.56 ± 0.79 & {\color[HTML]{000000} 76.05} & {\color[HTML]{000000} 0.82} \\
\textbf{Type3} & 77.68 ± 1.14 & 78.73 ± 1.19 & 77.68 ± 2.04 & 77.83 ± 1.19 & 77.53 ± 1.46 & 77.83 ± 0.45 & 78.73 ± 0.91 & 77.23 ± 0.69 & 77.98 ± 2.57 & 77.83 ± 0.45 & {\color[HTML]{FE0000} 77.91} & {\color[HTML]{000000} 0.48} \\
\textbf{Type4} & 76.77 ± 1.15 & 76.62 ± 2.03 & 75.57 ± 2.39 & 76.32 ± 2.23 & 77.37 ± 1.19 & 76.47 ± 3.14 & 75.87 ± 2.62 & 74.06 ± 2.57 & 75.87 ± 1.88 & 75.87 ± 2.49 & {\color[HTML]{000000} 76.08} & {\color[HTML]{000000} 0.89} \\
\textbf{Type5} & 76.17 ± 1.18 & 76.61 ± 1.38 & 76.77 ± 1.82 & 77.23 ± 1.72 & 75.87 ± 4.11 & 78.13 ± 0.69 & 76.92 ± 0.91 & 77.68 ± 1.45 & 75.11 ± 0.79 & 75.42 ± 3.86 & {\color[HTML]{000000} 76.59} & {\color[HTML]{000000} 0.96} \\
\textbf{Type6} & 75.57 ± 1.63 & 75.27 ± 0.94 & 76.02 ± 1.97 & 74.66 ± 1.57 & 76.02 ± 1.97 & 76.32 ± 0.94 & 74.21 ± 1.21 & 73.61 ± 2.73 & 74.21 ± 2.07 & 75.87 ± 1.72 & {\color[HTML]{000000} 75.18} & {\color[HTML]{000000} 0.94} \\
\textbf{Type7} & 74.81 ± 2.65 & 76.69 ± 1.95 & 76.77 ± 0.26 & 76.47 ± 1.97 & 77.98 ± 0.69 & 75.27 ± 1.38 & 77.08 ± 0.94 & 74.96 ± 1.71 & 74.66 ± 2.39 & 75.57 ± 2.72 & {\color[HTML]{000000} 76.02} & {\color[HTML]{000000} 1.12} \\
\textbf{Type8} & 76.92 ± 1.57 & 74.78 ± 2.77 & 73.76 ± 1.81 & 75.56 ± 2.76 & 74.06 ± 3.85 & 73.91 ± 4.18 & 72.41 ± 1.63 & 73.15 ± 0.69 & 74.06 ± 2.57 & 72.55 ± 1.31 & {\color[HTML]{000000} 74.11} & {\color[HTML]{000000} 1.37} \\
\textbf{Type9} & 77.38 ± 0.46 & 76.08 ± 1.29 & 75.87 ± 1.83 & 77.22 ± 1.89 & 74.96 ± 3.66 & 74.66 ± 2.07 & 76.02 ± 2.52 & 76.17 ± 1.38 & 76.32 ± 2.32 & 76.74 ± 0.79 & {\color[HTML]{000000} 76.14} & {\color[HTML]{000000} 0.87} \\
\textbf{Average} & {\color[HTML]{FE0000} 76.49} & {\color[HTML]{000000} 76.23} & {\color[HTML]{000000} 75.99} & {\color[HTML]{000000} 76.47} & {\color[HTML]{000000} 76.09} & {\color[HTML]{000000} 75.80} & {\color[HTML]{000000} 75.85} & {\color[HTML]{000000} 75.58} & {\color[HTML]{000000} 75.52} & {\color[HTML]{000000} 75.85} & {\color[HTML]{000000} 75.99} & \cellcolor[HTML]{FFFFFF}{\color[HTML]{FE0000} 0.24} \\ \hline
 & \multicolumn{11}{c}{\textbf{Robustness}} & \multicolumn{1}{l}{} \\ \hline
\textbf{Type1} & 75.28 ± 1.22 & 74.45 ± 0.27 & 74.48 ± 1.82 & 74.51 ± 0.65 & 74.21 ± 1.25 & 74.09 ± 1.25 & 74.11± 2.33 & 74.37 ± 2.11 & 74.06 ± 2.12 & 74.43 ± 0.45 & 74.40 & 0.35 \\
\textbf{Type2} & 75.47± 0.34 & 74.66 ± 0.63 & 74.61 ± 0.23 & 74.65 ± 1.51 & 74.19 ± 0.54 & 74.01 ± 2.86 & 74.23 ± 1.04 & 74.62 ± 1.45 & 74.21 ± 3.21 & 74.01 ± 1.34 & 74.47 & 0.44 \\
\textbf{Type3} & 75.99 ± 0.84 & 76.02 ± 0.63 & 75.78 ± 2.14 & 75.81 ± 1.56 & 75.33 ± 2.36 & 75.54 ± 1.22 & 75.76 ± 0.56 & 75.12 ± 0.64 & 75.32 ± 0.76 & 75.21 ± 1.45 & {\color[HTML]{FE0000} 75.59} & 0.33 \\
\textbf{Type4} & 74.55 ± 1.34 & 74.43 ± 1.24 & 74.87 ± 1.29 & 75.32 ± 1.56 & 75.53 ± 0.65 & 75.23 ± 2.17 & 75.07 ± 1.45 & 74.01 ± 1.52 & 74.86 ± 2.01 & 74.69 ± 3.21 & 74.86 & {\color[HTML]{000000} 0.46} \\
\textbf{Type5} & 75.02± 1.89 & 75.46 ± 0.65 & 75.65 ± 0.92 & 75.71 ± 0.52 & 75.01 ± 2.11 & 76.12 ± 1.65 & 75.87 ± 2.12 & 75.91 ± 2.14 & 74.85 ± 0.76 & 74.89 ± 2.43 & 75.45 & 0.47 \\
\textbf{Type6} & 73.69 ± 0.93 & 72.95 ± 1.26 & 72.99 ± 0.76 & 73.51 ± 2.37 & 73.86 ± 1.37 & 73.88 ± 0.96 & 73.02 ± 3.22 & 72.56 ± 1.86 & 72.64 ± 1.75 & 72.66 ± 1.43 & 73.18 & 0.51 \\
\textbf{Type7} & 74.81 ± 1.35 & 75.33 ± 0.31 & 75.37 ± 0.96 & 75.37 ± 2.56 & 76.01 ± 0.96 & 75.01 ± 2.11 & 75.78 ± 0.43 & 74.92 ± 3.72 & 74.21 ± 2.01 & 74.26 ± 0.34 & 75.11 & 0.59 \\
\textbf{Type8} & 73.98 ± 0.97 & 72.62 ± 1.23 & 72.58 ± 1.21 & 73.52 ± 0.86 & 73.26 ± 2.15 & 73.06 ± 2.19 & 72.21 ± 0.54 & 72.65 ± 3.87 & 72.83 ± 1.43 & 72.02 ± 2.31 & 72.87 & 0.60 \\
\textbf{Type9} & 75.51 ± 1.01 & 74.45 ± 0.38 & 74.42 ± 1.33 & 75.62 ± 1.09 & 73.66 ± 1.96 & 73.58 ± 1.27 & 73.87 ± 1.44 & 73.89 ± 0.49 & 73.91 ± 1.05 & 73.86 ± 2.14 & {\color[HTML]{000000} 74.28} & 0.74 \\
\textbf{Average} & \multicolumn{1}{r}{{\color[HTML]{FE0000} 74.92}} & \multicolumn{1}{r}{{\color[HTML]{343434} 74.49}} & \multicolumn{1}{r}{{\color[HTML]{343434} 74.53}} & \multicolumn{1}{r}{{\color[HTML]{343434} 74.89}} & \multicolumn{1}{r}{{\color[HTML]{343434} 74.56}} & \multicolumn{1}{r}{{\color[HTML]{343434} 74.50}} & \multicolumn{1}{r}{{\color[HTML]{343434} 74.43}} & \multicolumn{1}{r}{{\color[HTML]{343434} 74.23}} & \multicolumn{1}{r}{{\color[HTML]{343434} 74.10}} & \multicolumn{1}{r}{{\color[HTML]{343434} 74.00}} & 74.47 & {\color[HTML]{FE0000} 0.13} \\ \hline
\end{tabular}
}
\end{table*}

\begin{table*}[!tb]
\caption{Results (average $\pm$ standard deviation, \%) of trained GraphSAGE models with \emph{Manifold-Mixup} (dataset: Politifact) using hybrid pooling operators with different $\alpha$ settings. The best results are marked in red color.}
\label{table:alpha_GraphSAGE}
\centering
\scalebox{0.55}{
\begin{tabular}{lcccccccccccc}
\hline
\textbf{} & \multicolumn{11}{c}{\textbf{Test Accuracy}} & \multicolumn{1}{l}{} \\ \hline
\textbf{Mixup Ratio Alpha} & \textbf{0.05} & \textbf{0.10} & \textbf{0.15} & \textbf{0.20} & \textbf{0.25} & \textbf{0.30} & \textbf{0.35} & \textbf{0.40} & \textbf{0.45} & \textbf{0.50} & {\color[HTML]{343434} \textbf{Average}} & \textbf{Std} \\ \hline
\textbf{Type1} & 78.88 ± 1.14 & 78.89 ± 1.88 & 78.58 ± 0.26 & 77.22 ± 1.89 & 76.32± 0.68 & 78.28 ± 0.79 & 77.98 ± 1.59 & 77.83 ± 0.91 & 78.28 ± 1.18 & 78.43 ± 0.68 & {\color[HTML]{000000} 78.07} & {\color[HTML]{000000} 0.79} \\
\textbf{Type2} & 76.62 ± 1.16 & 77.61 ± 0.32 & 76.62 ± 0.94 & 77.38 ± 0.79 & 79.19 ± 0.91 & 76.92 ± 0.91 & 77.07 ± 1.14 & 77.38 ± 0.92 & 77.83 ± 1.19 & 77.07 ± 0.66 & {\color[HTML]{000000} 77.37} & {\color[HTML]{000000} 0.75} \\
\textbf{Type3} & 79.18 ± 0.79 & 78.73 ± 0.67 & 77.68 ± 0.93 & 79.03 ± 0.53 & 78.58 ± 2.04 & 78.58 ± 1.14 & 79.33 ± 1.05 & 77.83 ± 0.79 & 78.73 ± 1.63 & 75.87 ± 1.72 & {\color[HTML]{EA4335} \textbf{78.35}} & {\color[HTML]{000000} 1.02} \\
\textbf{Type4} & 76.32 ± 0.69 & 77.23 ± 0.55 & 76.77 ± 1.46 & 76.02 ± 1.19 & 77.38 ± 2.27 & 75.72 ± 1.31 & 77.37 ± 1.97 & 76.17 ± 0.69 & 77.38 ± 0.46 & 76.02 ± 0.45 & {\color[HTML]{000000} 76.64} & {\color[HTML]{000000} 0.66} \\
\textbf{Type5} & 77.53 ± 1.31 & 76.89 ± 0.99 & 77.22 ± 3.01 & 76.77 ± 2.91 & 77.38 ± 2.08 & 74.51 ± 1.83 & 74.66 ± 4.15 & 77.23 ± 2.04 & 73.61 ± 2.91 & 75.57 ± 1.19 & {\color[HTML]{000000} 76.14} & {\color[HTML]{000000} 1.43} \\
\textbf{Type6} & 74.66 ± 2.07 & 76.57 ± 0.23 & 75.31 ± 2.21 & 76.77 ± 3.02 & 74.96 ± 2.49 & 74.66 ± 4.07 & 75.42 ± 2.09 & 74.81 ± 1.72 & 74.81 ± 0.94 & 74.51 ± 1.88 & {\color[HTML]{000000} 75.25} & {\color[HTML]{000000} 0.80} \\
\textbf{Type7} & 76.63 ± 1.38 & 76.93 ± 1.28 & 74.51 ± 1.38 & 77.83 ± 0.45 & 76.93 ± 1.19 & 75.26 ± 1.14 & 76.77 ± 0.94 & 76.77 ± 0.94 & 75.87 ± 0.69 & 76.62 ± 1.83 & {\color[HTML]{000000} 76.41} & {\color[HTML]{000000} 0.95} \\
\textbf{Type8} & 76.47 ± 1.36 & 76.71 ± 0.97 & 75.41 ± 1.31 & 75.26 ± 2.32 & 74.96 ± 3.21 & 76.32 ± 1.83 & 75.26 ± 1.14 & 74.36 ± 2.76 & 76.77 ± 2.32 & 74.81 ± 1.38 & {\color[HTML]{000000} 75.63} & {\color[HTML]{000000} 0.86} \\
\textbf{Type9} & 75.41 ± 0.53 & 76.69 ± 0.32 & 75.11 ± 1.63 & 75.87 ± 1.14 & 74.51 ± 1.14 & 76.62 ± 1.86 & 72.71 ± 1.16 & 75.72 ± 0.69 & 74.06 ± 2.94 & 74.06 ± 1.45 & {\color[HTML]{000000} 75.08} & {\color[HTML]{000000} 1.25} \\
\textbf{Average} & {\color[HTML]{000000} 76.86} & {\color[HTML]{EA4335} \textbf{77.36}} & {\color[HTML]{000000} 76.36} & {\color[HTML]{000000} 76.91} & {\color[HTML]{000000} 76.69} & {\color[HTML]{000000} 76.32} & {\color[HTML]{000000} 76.29} & {\color[HTML]{000000} 76.46} & {\color[HTML]{000000} 76.37} & {\color[HTML]{000000} 75.88} & {\color[HTML]{000000} 76.55} & \cellcolor[HTML]{FFFFFF}{\color[HTML]{FE0000} 0.25} \\ \hline
 & \multicolumn{11}{c}{\textbf{Robustness}} & \multicolumn{1}{l}{} \\ \hline
\textbf{Type1} & 77.26 ± 0.34 & 77.37 ± 0.63 & 77.12 ± 1.23 & 76.78 ± 2.31 & 76.01 ± 1.23 & 76.67 ± 1.34 & 76.56 ± 2.31 & 76.45 ± 1.45 & 76.54 ± 2.12 & 76.53 ± 1.23 & {\color[HTML]{FE0000} 76.73} & 0.42 \\
\textbf{Type2} & 76.25 ± 2.34 & 76.94 ± 1.19 & 76.23 ± 2.12 & 76.84 ± 1.23 & 77.21 ± 0.45 & 76.01 ± 1.36 & 76.34 ± 2.32 & 76.57 ± 2.56 & 76.87 ± 2.15 & 76.54 ± 2.45 & 76.58 & {\color[HTML]{FE0000} 0.38} \\
\textbf{Type3} & 75.96 ± 1.94 & 75.33 ± 2.24 & 74.84 ± 1.45 & 75.65 ± 1.12 & 75.12 ± 1.34 & 75.08 ± 3.21 & 75.35 ± 3.21 & 75.21 ± 3.45 & 75.32 ± 0.56 & 74.21 ± 2.19 & {\color[HTML]{000000} 75.21} & 0.47 \\
\textbf{Type4} & 75.54 ± 0.65 & 75.79 ± 0.33 & 75.34 ± 2.46 & 75.21 ± 0.53 & 75.87 ± 1.56 & 74.78 ± 3.56 & 75.98 ± 1.34 & 75.02 ± 2.11 & 75.09 ± 0.76 & 74.65 ± 1.94 & 75.33 & {\color[HTML]{000000} 0.46} \\
\textbf{Type5} & 75.89 ± 2.12 & 75.56 ± 1.27 & 75.65 ± 1.83 & 75.02 ± 1.34 & 75.98 ± 2.01 & 74.87 ± 0.45 & 74.45 ± 2.11 & 75.78 ± 0.56 & 75.02 ± 1.34 & 75.43 ± 1.45 & 75.37 & 0.50 \\
\textbf{Type6} & 73.21 ± 0.67 & 73.98 ± 1.59 & 73.01 ± 2.76 & 74.21 ± 2.94 & 73.03 ± 1.34 & 73.01 ± 2.12 & 73.88 ± 0.45 & 73.23 ± 0.54 & 73.12 ± 0.45 & 73.01 ± 0.54 & 73.37 & 0.47 \\
\textbf{Type7} & 71.71 ± 2.54 & 71.72 ± 0.32 & 71.12 ± 0.65 & 72.65 ± 2.61 & 72.45 ± 0.34 & 72.01 ± 0.45 & 71.98 ± 2.12 & 71.78 ± 1.23 & 71.45 ± 2.56 & 71.98 ± 0.76 & 71.89 & 0.44 \\
\textbf{Type8} & 74.02 ± 2.12 & 74.21 ± 1.17 & 73.87 ± 0.75 & 73.68 ± 3.21 & 73.54 ± 2.14 & 74.32 ± 0.46 & 74.12 ± 2.14 & 73.87 ± 1.65 & 73.92 ± 2.54 & 72.99 ± 0.87 & 73.85 & {\color[HTML]{FE0000} 0.38} \\
\textbf{Type9} & 75.12 ± 1.32 & 75.56 ± 0.66 & 75.02 ± 1.45 & 75.21 ± 0.87 & 75.01 ± 0.45 & 76.02 ± 1.23 & 71.98 ± 1.09 & 72.56 ± 0.87 & 71.88 ± 1.07 & 71.78 ± 1.47 & {\color[HTML]{000000} 74.01} & 1.73 \\
\textbf{Average} & {\color[HTML]{000000} 75.00} & {\color[HTML]{FE0000} 75.16} & {\color[HTML]{000000} 74.69} & {\color[HTML]{000000} 75.03} & {\color[HTML]{000000} 74.91} & {\color[HTML]{000000} 74.75} & {\color[HTML]{000000} 74.52} & {\color[HTML]{000000} 74.50} & {\color[HTML]{000000} 74.36} & {\color[HTML]{000000} 74.12} & 74.70 & {\color[HTML]{000000} 0.43} \\ \hline
\end{tabular}
}
\end{table*}

\begin{table*}[!tb]
\caption{Results (average $\pm$ standard deviation, \%) of trained GIN models with \emph{Manifold-Mixup} (dataset: Politifact) using hybrid pooling operators with different $\alpha$ settings. The best results are marked in red color.}
\label{table:alpha_GIN}
\centering
\scalebox{0.55}{
\begin{tabular}{lcccccccccccc}
\hline
\textbf{} & \multicolumn{11}{c}{\textbf{Test Accuracy}} & \multicolumn{1}{l}{} \\ \hline
\textbf{Mixup Ratio Alpha} & \textbf{0.05} & \textbf{0.10} & \textbf{0.15} & \textbf{0.20} & \textbf{0.25} & \textbf{0.30} & \textbf{0.35} & \textbf{0.40} & \textbf{0.45} & \textbf{0.50} & {\color[HTML]{343434} \textbf{Average}} & \textbf{Std} \\ \hline
 \textbf{Type1} & {\color[HTML]{000000} 75.87 ± 2.14} & {\color[HTML]{000000} 78.28 ± 0.93} & {\color[HTML]{000000} 75.72 ± 1.83} & {\color[HTML]{000000} 74.96 ± 1.14} & {\color[HTML]{000000} 78.43 ± 1.45} & {\color[HTML]{000000} 77.98 ± 0.26} & {\color[HTML]{000000} 77.52 ± 0.46} & {\color[HTML]{000000} 76.77 ± 0.69} & {\color[HTML]{000000} 78.13 ± 1.59} & {\color[HTML]{000000} 76.47 ± 0.45} & {\color[HTML]{000000} 77.01} & {\color[HTML]{000000} 1.23} \\
\textbf{Type2} & {\color[HTML]{000000} 78.28 ± 0.91} & {\color[HTML]{000000} 78.06 ± 1.61} & {\color[HTML]{000000} 78.73 ± 1.26} & {\color[HTML]{000000} 78.28 ± 0.78} & {\color[HTML]{000000} 76.77 ± 0.69} & {\color[HTML]{000000} 78.28 ± 1.56} & {\color[HTML]{000000} 78.43 ± 2.14} & {\color[HTML]{000000} 78.13 ± 1.14} & {\color[HTML]{000000} 77.53 ± 0.26} & {\color[HTML]{000000} 77.38 ± 1.63} & {\color[HTML]{000000} 77.99} & {\color[HTML]{000000} 0.59} \\
\textbf{Type3} & {\color[HTML]{000000} 76.47 ± 0.45} & {\color[HTML]{000000} 77.92 ± 2.85} & {\color[HTML]{000000} 76.37 ± 2.95} & {\color[HTML]{000000} 75.57 ± 2.75} & {\color[HTML]{000000} 76.32 ± 2.72} & {\color[HTML]{000000} 75.72 ± 0.26} & {\color[HTML]{000000} 74.95 ± 1.13} & {\color[HTML]{000000} 76.62 ± 0.94} & {\color[HTML]{000000} 76.69 ± 0.32} & {\color[HTML]{000000} 76.62 ± 0.94} & {\color[HTML]{000000} 76.33} & {\color[HTML]{000000} 0.80} \\
\textbf{Type4} & {\color[HTML]{000000} 79.34 ± 0.26} & {\color[HTML]{000000} 77.74 ± 0.74} & {\color[HTML]{000000} 78.58 ± 2.23} & {\color[HTML]{000000} 77.83 ± 0.45} & {\color[HTML]{000000} 77.08 ± 1.38} & {\color[HTML]{000000} 77.08 ± 1.38} & {\color[HTML]{000000} 76.47 ± 0.78} & {\color[HTML]{000000} 77.98 ± 0.26} & {\color[HTML]{000000} 77.68 ± 0.25} & {\color[HTML]{000000} 77.58 ± 0.69} & {\color[HTML]{000000} 77.74} & {\color[HTML]{000000} 0.81} \\
\textbf{Type5} & {\color[HTML]{000000} 78.28 ± 0.92} & {\color[HTML]{000000} 77.64 ± 1.72} & {\color[HTML]{000000} 76.92 ± 0.45} & {\color[HTML]{000000} 78.43 ± 0.26} & {\color[HTML]{000000} 77.53 ± 1.71} & {\color[HTML]{000000} 78.43 ± 1.14} & {\color[HTML]{000000} 77.68 ± 0.52} & {\color[HTML]{000000} 77.98 ± 0.68} & {\color[HTML]{000000} 78.43 ± 1.14} & {\color[HTML]{000000} 77.68 ± 1.88} & {\color[HTML]{000000} 77.90} & {\color[HTML]{000000} 0.50} \\
\textbf{Type6} & {\color[HTML]{000000} 77.68 ± 1.14} & {\color[HTML]{000000} 78.05 ± 1.08} & {\color[HTML]{000000} 77.98 ± 1.38} & {\color[HTML]{000000} 78.88 ± 1.05} & {\color[HTML]{000000} 76.47 ± 1.63} & {\color[HTML]{000000} 77.23 ± 1.05} & {\color[HTML]{000000} 77.37 ± 0.79} & {\color[HTML]{000000} 77.53 ± 1.14} & {\color[HTML]{000000} 77.23 ± 1.45} & {\color[HTML]{000000} 77.98 ± 1.71} & {\color[HTML]{000000} 77.64} & {\color[HTML]{000000} 0.64} \\
\textbf{Type7} & {\color[HTML]{000000} 78.59 ± 1.05} & {\color[HTML]{000000} 77.83 ± 1.61} & {\color[HTML]{000000} 79.64 ± 0.45} & {\color[HTML]{000000} 79.34 ± 0.26} & {\color[HTML]{000000} 79.34 ± 0.69} & {\color[HTML]{000000} 79.04 ± 1.14} & {\color[HTML]{000000} 78.58 ± 0.26} & {\color[HTML]{000000} 77.83 ± 2.39} & {\color[HTML]{000000} 79.04 ± 0.69} & {\color[HTML]{000000} 79.94 ± 1.31} & {\color[HTML]{EA4335} 78.92} & {\color[HTML]{000000} 0.71} \\
\textbf{Type8} & {\color[HTML]{000000} 78.58 ± 0.26} & {\color[HTML]{000000} 78.39 ± 0.78} & {\color[HTML]{000000} 76.77 ± 2.61} & {\color[HTML]{000000} 78.43 ± 3.27} & {\color[HTML]{000000} 77.68 ± 1.83} & {\color[HTML]{000000} 78.28 ± 1.21} & {\color[HTML]{000000} 77.38 ± 1.63} & {\color[HTML]{000000} 78.13 ± 1.31} & {\color[HTML]{000000} 77.11 ± 1.86} & {\color[HTML]{000000} 76.98 ± 1.26} & {\color[HTML]{000000} 77.77} & {\color[HTML]{000000} 0.67} \\
\textbf{Type9} & {\color[HTML]{000000} 79.19 ± 0.91} & {\color[HTML]{000000} 77.74 ± 1.71} & {\color[HTML]{000000} 79.49 ± 1.14} & {\color[HTML]{000000} 79.49 ± 1.14} & {\color[HTML]{000000} 78.73 ± 0.46} & {\color[HTML]{000000} 78.73 ± 0.46} & {\color[HTML]{000000} 79.33 ± 1.75} & {\color[HTML]{000000} 78.58 ± 1.59} & {\color[HTML]{000000} 76.47 ± 1.97} & {\color[HTML]{000000} 76.92 ± 1.71} & {\color[HTML]{000000} 78.47} & {\color[HTML]{000000} 1.08} \\
\textbf{Average} & {\color[HTML]{EA4335} \textbf{78.03}} & {\color[HTML]{000000} \textbf{77.96}} & {\color[HTML]{000000} 77.80} & {\color[HTML]{000000} 77.91} & {\color[HTML]{000000} 77.59} & {\color[HTML]{000000} 77.86} & {\color[HTML]{000000} 77.52} & {\color[HTML]{000000} 77.73} & {\color[HTML]{000000} 77.59} & {\color[HTML]{000000} 77.51} & {\color[HTML]{000000} 77.75} & \cellcolor[HTML]{FFFFFF}{\color[HTML]{FE0000} 0.23} \\ \hline
 & \multicolumn{11}{c}{\textbf{Robustness}} & \multicolumn{1}{l}{} \\ \hline
\textbf{Type1} & {\color[HTML]{000000} 73.22 ± 1.34} & {\color[HTML]{000000} 73.31 ± 2.56} & {\color[HTML]{000000} 72.68 ± 2.34} & {\color[HTML]{000000} 71.87 ± 3.21} & {\color[HTML]{000000} 73.66 ± 2.34} & {\color[HTML]{000000} 73.01 ± 1.23} & {\color[HTML]{000000} 72.67 ± 1.23} & {\color[HTML]{000000} 72.21 ± 1.34} & {\color[HTML]{000000} 72.98 ± 2.23} & {\color[HTML]{000000} 71.67 ± 2.21} & {\color[HTML]{000000} 72.73} & {\color[HTML]{000000} 0.64} \\
\textbf{Type2} & {\color[HTML]{000000} 74.42 ± 2.45} & {\color[HTML]{000000} 74.21 ± 0.64} & {\color[HTML]{000000} 74.76 ± 0.45} & {\color[HTML]{000000} 74.54 ± 1.34} & {\color[HTML]{000000} 73.03 ± 2.12} & {\color[HTML]{000000} 74.21 ± 0.34} & {\color[HTML]{000000} 74.31 ± 1.45} & {\color[HTML]{000000} 74.23 ± 2.34} & {\color[HTML]{000000} 73.89 ± 0.45} & {\color[HTML]{000000} 73.55 ± 1.23} & {\color[HTML]{000000} 74.12} & {\color[HTML]{000000} 0.51} \\
\textbf{Type3} & {\color[HTML]{000000} 75.01 ± 3.21} & {\color[HTML]{000000} 75.06 ± 3.78} & {\color[HTML]{000000} 75.01 ± 0.65} & {\color[HTML]{000000} 74.76 ± 1.87} & {\color[HTML]{000000} 75.04 ± 1.34} & {\color[HTML]{000000} 74.87 ± 0.87} & {\color[HTML]{000000} 74.01 ± 0.54} & {\color[HTML]{000000} 74.87 ± 1.34} & {\color[HTML]{000000} 74.88 ± 1.23} & {\color[HTML]{000000} 74.76 ± 0.34} & {\color[HTML]{EA4335} 74.83} & {\color[HTML]{000000} 0.31} \\
\textbf{Type4} & {\color[HTML]{000000} 76.89 ± 0.65} & {\color[HTML]{000000} 75.56 ± 1.92} & {\color[HTML]{000000} 76.03 ± 1.45} & {\color[HTML]{000000} 75.87 ± 1.34} & {\color[HTML]{000000} 75.32 ± 2.32} & {\color[HTML]{000000} 75.21 ± 1.98} & {\color[HTML]{000000} 74.54 ± 1.45} & {\color[HTML]{000000} 75.02 ± 2.34} & {\color[HTML]{000000} 74.99 ± 2.12} & {\color[HTML]{000000} 74.77 ± 0.56} & {\color[HTML]{000000} 75.42} & {\color[HTML]{000000} 0.70} \\
\textbf{Type5} & {\color[HTML]{000000} 77.23 ± 1.45} & {\color[HTML]{000000} 76.41 ± 0.36} & {\color[HTML]{000000} 75.97 ± 3.21} & {\color[HTML]{000000} 76.76 ± 2.31} & {\color[HTML]{000000} 76.03 ± 2.98} & {\color[HTML]{000000} 76.76 ± 2.76} & {\color[HTML]{000000} 76.01 ± 2.12} & {\color[HTML]{000000} 76.32 ± 3.21} & {\color[HTML]{000000} 76.97 ± 1.23} & {\color[HTML]{000000} 76.74 ± 0.45} & {\color[HTML]{000000} 76.52} & {\color[HTML]{000000} 0.44} \\
\textbf{Type6} & {\color[HTML]{000000} 74.01 ± 0.45} & {\color[HTML]{000000} 74.66 ± 0.79} & {\color[HTML]{000000} 74.02 ± 0.45} & {\color[HTML]{000000} 74.98 ± 2.91} & {\color[HTML]{000000} 73.54 ± 1.65} & {\color[HTML]{000000} 73.98 ± 1.45} & {\color[HTML]{000000} 73.99 ± 2.43} & {\color[HTML]{000000} 74.04 ± 0.98} & {\color[HTML]{000000} 74.01 ± 2.14} & {\color[HTML]{000000} 74.04 ± 2.12} & {\color[HTML]{000000} 74.13} & {\color[HTML]{000000} 0.40} \\
\textbf{Type7} & {\color[HTML]{000000} 75.98 ± 0.65} & {\color[HTML]{000000} 75.11 ± 0.66} & {\color[HTML]{000000} 76.67 ± 2.12} & {\color[HTML]{000000} 76.45 ± 1.45} & {\color[HTML]{000000} 76.37 ± 0.98} & {\color[HTML]{000000} 76.12 ± 0.76} & {\color[HTML]{000000} 76.01 ± 2.54} & {\color[HTML]{000000} 75.67 ± 0.56} & {\color[HTML]{000000} 75.96 ± 1.45} & {\color[HTML]{000000} 75.99 ± 1.22} & {\color[HTML]{000000} 76.03} & {\color[HTML]{000000} 0.43} \\
\textbf{Type8} & {\color[HTML]{000000} 57.65 ± 1.34} & {\color[HTML]{000000} 57.19 ± 2.36} & {\color[HTML]{000000} 56.57 ± 0.54} & {\color[HTML]{000000} 57.03 ± 2.45} & {\color[HTML]{000000} 56.94 ± 1.23} & {\color[HTML]{000000} 57.01 ± 0.34} & {\color[HTML]{000000} 56.34 ± 0.65} & {\color[HTML]{000000} 56.54 ± 1.03} & {\color[HTML]{000000} 55.67 ± 2.34} & {\color[HTML]{000000} 55.23 ± 0.46} & {\color[HTML]{000000} 56.62} & {\color[HTML]{000000} 0.72} \\
\textbf{Type9} & {\color[HTML]{000000} 75.65 ± 2.34} & {\color[HTML]{000000} 74.36 ± 2.28} & {\color[HTML]{000000} 75.21 ± 2.01} & {\color[HTML]{000000} 75.19 ± 2.11} & {\color[HTML]{000000} 74.89 ± 2.41} & {\color[HTML]{000000} 74.78 ± 1.02} & {\color[HTML]{000000} 75.21 ± 0.54} & {\color[HTML]{000000} 74.98 ± 1.12} & {\color[HTML]{000000} 73.78 ± 0.45} & {\color[HTML]{000000} 73.84 ± 0.66} & {\color[HTML]{000000} 74.79} & {\color[HTML]{000000} 0.62} \\
\textbf{Average} & {\color[HTML]{EA4335} \textbf{73.34}} & {\color[HTML]{000000} 72.87} & {\color[HTML]{000000} 72.99} & {\color[HTML]{000000} 73.05} & {\color[HTML]{000000} 72.76} & {\color[HTML]{000000} 72.88} & {\color[HTML]{000000} 72.57} & {\color[HTML]{000000} 72.65} & {\color[HTML]{000000} 72.57} & {\color[HTML]{000000} 72.29} & {\color[HTML]{000000} 72.80} & {\color[HTML]{FE0000} 0.15} \\ \hline
\end{tabular}
}
\end{table*}

\begin{table*}[!tb]
\caption{Results of statistical testing on Accuracy \& Robustness. \textbf{True}: indicates that the comparison is statistically significant after the  \emph{Bonferroni correction} adjustment. Statistical testing methods: \emph{Wilcoxon signed-rank test}. A gray background highlights the result marked as \textbf{True}.}
\label{table:alpha_statistical_testing}
\centering
\scalebox{0.8}{
\begin{tabular}{lccccccccc}
\hline
& & \multicolumn{8}{c}{\textbf{Test Accuracy}}  \\ \hline
                    & Mixup Ration Alpha & 0.15 & 0.20 & 0.25 & 0.30 & 0.35 & 0.40 & 0.45 & 0.50 \\ \hline
\multirow{2}{*}{GCN} & 0.05               &    -  &    -  &   -   &   -   &   -   &  -    &    -  &  -    \\
                    & 0.10               &    \cellcolor[HTML]{C0C0C0}{\color[HTML]{000000}True }     &    \cellcolor[HTML]{C0C0C0}{\color[HTML]{000000}True }     &   -   &    -  &    -  &    -  &  -    &  -    \\ \hline
\multirow{2}{*}{GAT}&  0.05                  &   -   &    -  &    -  &   -   &  -    &    -  &    \cellcolor[HTML]{C0C0C0}{\color[HTML]{000000}True }   &   -   \\
                    &  0.10                  &   -   &  -    &   -   &   -   &  -    &   -   &   \cellcolor[HTML]{C0C0C0}{\color[HTML]{000000}True }    &    -  \\ \hline
\multirow{2}{*}{ GraphSAGE}& 0.05                   &  -    &   -   &  -    &  -    &  -    &   -   &   -   &  \cellcolor[HTML]{C0C0C0}{\color[HTML]{000000}True }     \\
                    &   0.10                 &    \cellcolor[HTML]{C0C0C0}{\color[HTML]{000000}True }   &   -   &  -    &   \cellcolor[HTML]{C0C0C0}{\color[HTML]{000000}True }    &    \cellcolor[HTML]{C0C0C0}{\color[HTML]{000000}True }   &   \cellcolor[HTML]{C0C0C0}{\color[HTML]{000000}True }    &    \cellcolor[HTML]{C0C0C0}{\color[HTML]{000000}True }   &  \cellcolor[HTML]{C0C0C0}{\color[HTML]{000000}True }     \\\hline
 \multirow{2}{*}{ GIN}&  0.05                  &  -    &    -  &    -  &  -    &   -   &  -    &   -   &    -  \\
                    &    0.10                &   -   &   -   &    -  & -     &   -   & -    & -     & -     \\\hline                  
& &  \multicolumn{8}{c}{\textbf{Robustness}}  \\ \hline
\multirow{2}{*}{GCN} & 0.05               &     \cellcolor[HTML]{C0C0C0}{\color[HTML]{000000}True }   &   -   &   -   &    \cellcolor[HTML]{C0C0C0}{\color[HTML]{000000}True }    &    -  &    \cellcolor[HTML]{C0C0C0}{\color[HTML]{000000}True }    &    \cellcolor[HTML]{C0C0C0}{\color[HTML]{000000}True }    &     \cellcolor[HTML]{C0C0C0}{\color[HTML]{000000}True }   \\
                    & 0.10               &   -   &   -   &  -    &   -   &  -    &   -   &   -   &  -    \\ \hline
\multirow{2}{*}{GAT}&  0.05                  & -    &   -   &  -    &   -   &   -   &    \cellcolor[HTML]{C0C0C0}{\color[HTML]{000000}True }    &   \cellcolor[HTML]{C0C0C0}{\color[HTML]{000000}True }     &     \cellcolor[HTML]{C0C0C0}{\color[HTML]{000000}True }   \\
                    &  0.10                  &   -   &   \cellcolor[HTML]{C0C0C0}{\color[HTML]{000000}True }     &  -    &   -   &  -    &   \cellcolor[HTML]{C0C0C0}{\color[HTML]{000000}True }     &    \cellcolor[HTML]{C0C0C0}{\color[HTML]{000000}True }    &      \cellcolor[HTML]{C0C0C0}{\color[HTML]{000000}True }  \\ \hline
\multirow{2}{*}{ GraphSAGE}& 0.05                   &   \cellcolor[HTML]{C0C0C0}{\color[HTML]{000000}True }    &     - &   -   &  -    &  -     &   \cellcolor[HTML]{C0C0C0}{\color[HTML]{000000}True }    &   \cellcolor[HTML]{C0C0C0}{\color[HTML]{000000}True }    &  \cellcolor[HTML]{C0C0C0}{\color[HTML]{000000}True }     \\
                    &   0.10                 &   \cellcolor[HTML]{C0C0C0}{\color[HTML]{000000}True }    &    -  &   -   &  \cellcolor[HTML]{C0C0C0}{\color[HTML]{000000}True }     &   -   &   \cellcolor[HTML]{C0C0C0}{\color[HTML]{000000}True }    &  \cellcolor[HTML]{C0C0C0}{\color[HTML]{000000}True }     &   \cellcolor[HTML]{C0C0C0}{\color[HTML]{000000}True }    \\\hline
 \multirow{2}{*}{ GIN}&  0.05                  &  -    &    -  &     \cellcolor[HTML]{C0C0C0}{\color[HTML]{000000}True }  &    -  &   -   &   \cellcolor[HTML]{C0C0C0}{\color[HTML]{000000}True }    &    -  &    \cellcolor[HTML]{C0C0C0}{\color[HTML]{000000}True }   \\
                    &    0.10                &  -    &   -   &  -    &    -  &    -  &   -   &    -  &  -    \\\hline
                    
\end{tabular}
}
\end{table*}

\begin{figure}[!tb]
	\centering       
	\subfigure[Effectiveness of trained GCN]{
    \includegraphics[scale=0.45]{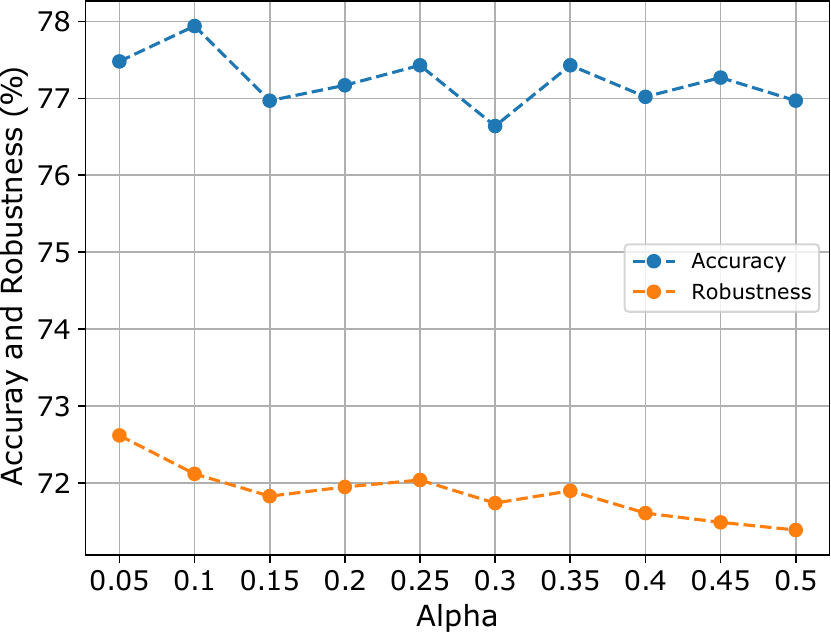}%
    } 
    \subfigure[Effectiveness of trained GAT]{
    \includegraphics[scale=0.45]{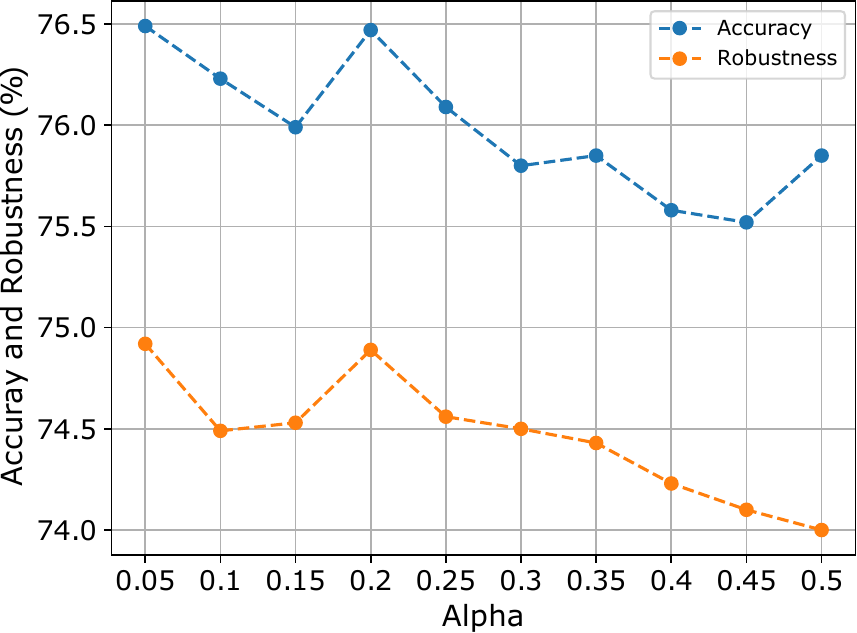}%
    }
    
    \subfigure[Effectiveness of trained GIN]{
    \includegraphics[scale=0.45]{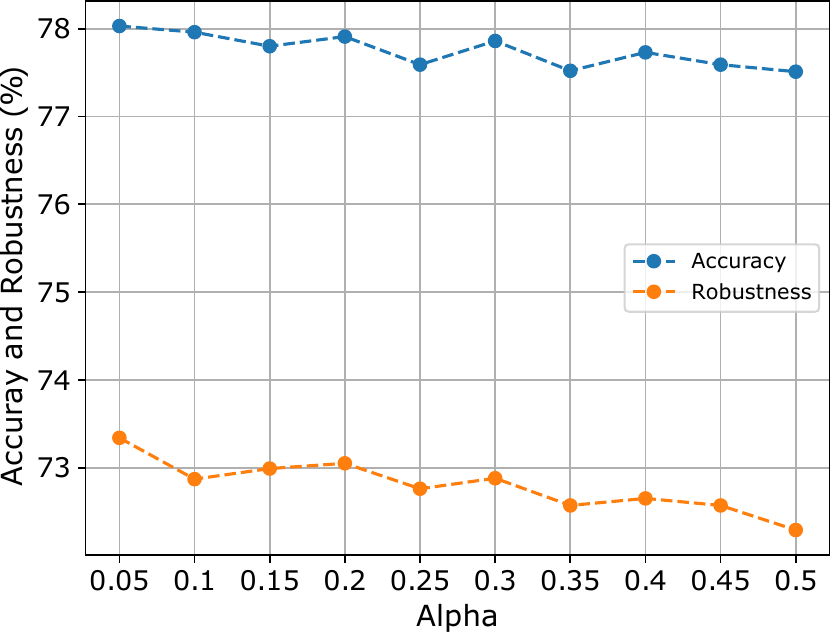}%
    }
    \subfigure[Effectiveness of trained GraphSAGE]{
    \includegraphics[scale=0.45]{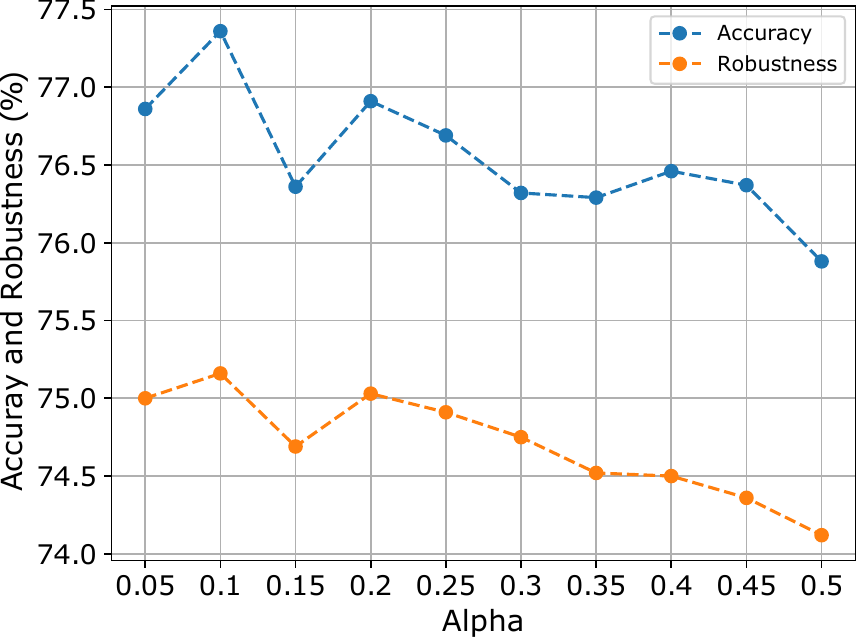}%
    }
    
	\caption{Accuracy and Robustness trendings with different $\alpha$.}
	\label{fig:trend}
\end{figure}

Table~\ref{table:results3}, Table~\ref{table:alpha_GAT}, Table~\ref{table:alpha_GraphSAGE}, and Table~\ref{table:alpha_GIN} present the detailed results of GNN models including GCN, GAT, GraphSAGE, and GIN on the Politifact dataset when using the Mixup training strategy with different $\alpha$ settings. First, we analyze how $\alpha$ affects each type of graph pooling method. Column \emph{Std} shows the standard deviation of the results of each pooling method. We can see that, in the type level, $\alpha$ also does not greatly impact the Mixup training. The maximum standard deviations are only 1.84\% (Accuracy of Type 3, Table~\ref{table:results3}) and 1.73\% (Robustness of Type 9, Table~\ref{table:alpha_GraphSAGE}). Maybe interestingly, we can find the two most stable and least stable types, i.e., The operator Type 4 ($\mathcal{M}_{sum}(\mathcal{P}_{att},\mathcal{P}_{sum})$) from Table~\ref{table:results3} and Type 2 ($\mathcal{M}_{mul}(\mathcal{P}_{att},\mathcal{P}_{max})$) from Table~\ref{table:alpha_GraphSAGE} always have the smallest standard deviation values for both accuracy and robustness. In the Average row, across all hybrid pooling operators, the greatest average accuracy improvement is observed at $\alpha$ values of 0.05 and 0.10, analyzed across all GNNs. When considering robustness, except for GraphSAGE, which demonstrates the best average robustness improvement when $\alpha$ equals 0.10, the other GNNs show that setting $\alpha$ to 0.05 yields the best average robustness improvement compared to other $\alpha$ value settings. 

Figure \ref{fig:trend} depicts the trending of average (e.g., row \textit{Average} in Table~\ref{table:results3}) results of all GNN models. Then, we can see that although both the accuracy and robustness difference between using different $\alpha$ is not that big, i.e., the gap between accuracy (robustness) is 1.30\% (1.23\%). The smaller $\alpha$ can produce models with better performance, which is consistent with the conclusion of the original Mixup work~\cite{zhang2018mixup}. Especially for the robustness, there is a clear decreasing trend when $\alpha$ becomes bigger. 

\textbf{Statistical analysis.} We conduct statistical testing to evaluate the significance of both accuracy and robustness improvement achieved by using hybrid operators at different Mixup ratios $\alpha$. As depicted in Table~\ref{table:alpha_statistical_testing}, overall, the comparison becomes more statistically significant as the value of $\alpha$ increases, in terms of accuracy and robustness. However, there are a few cases, such as Test Accuracy-GIN, where the comparisons are consistently not statistically significant regardless of the increase in the value of $\alpha$. We recommend setting $\alpha$ = 0.10 for GNNs training to improve accuracy, as in 2 out of 3 cases, it has shown to be statistically significant compared to others. Moving to robustness, we recommend setting $\alpha$ = 0.05 for GNN training, as it consistently shows statistical significance compared to other $\alpha$ settings across all GNN models. Additionally, $\alpha$ = 0.10 is statistically significant compared to other $\alpha$ settings only in GraphSAGE and GAT.

In conclusion, \emph{Manifold-Mixup} for graph-structured data with hybrid graph pooling is resilient to the $\alpha$ setting. Moreover, small $\alpha$ is recommended in the practical usage of Mixup. Although the operator Type 1 is the best for producing high-performance GNN models, Type 4 for GCN and Type 2 for GraphSAGE are the most stable methods that are least affected by $\alpha$.

\noindent\colorbox{gray!20}{\framebox{\parbox{0.96\linewidth}{
\textbf{Answer to RQ3}: The performance~(both clean accuracy and robustness) of GNN models is not sensitive (with a less than 2\% difference) to the change of parameter $\alpha$. A small $\alpha$ (e.g., 0.05 and 0.10) is recommended for GNNs training with \emph{Manifold-Mixup}. }}}

%\section{Discussion}
\section{Threats to Validity}
The internal threat to validity comes from implementing the GNNs, each pooling method, and the Mixup for graph-structured data. The implementation of GNNs is based on \cite{Fey/Lenssen/2019}, and the implementation of Mixup for graph data is based on the officially released project of Mixup~\cite{zhang2018mixup}.

External validity threats lie in the selected NLP and PL tasks, datasets, and GNNs. We consider both the traditional NLP tasks (text level) and PL tasks (source code level) in the study and include two datasets for each task. Particularly, for the NLP task, we consider two well-studied datasets and four types of node embedding. For the PL task, we include two popular programming languages (Java and Python). For the GNN models, we consider six famous graph neural networks: GCN, GCN-Virtual, GIN, GIN-Virtual, GAT, and GraphSAGE.

The construct threats to validity mainly come from the parameters of Mixup, randomness, and evaluation measures. Mixup only contains the parameter $\alpha$ that controls the weight of mixing two input instances. We follow the recommendation of the original Mixup algorithm and investigate the impact of this parameter. Moreover, further, we explore the impact of $\alpha$ in our study. To reduce the impact of randomness, we repeat each experiment five times with different random seeds and report the average and standard deviation results. Finally, concerning the evaluation measures, we consider both the test accuracy of the original test data and the robustness of corrupted test data. The latter one is specific for evaluating the generalization ability of GNNs.

\section{Conclusions}

In this paper, we comprehensively investigated how the graph pooling methods impact the effectiveness of Mixups when dealing with graph-structured data. We considered the Max-pooling operator, the state-of-the-art GMT pooling operator, and nine different hybrid pooling methods defined ourselves. In the empirical analysis part, we conducted experiments on both traditional NLP tasks (fake news detection) and PL tasks (problem classification) using six types of GNN architecture. The experimental results demonstrated that the pooling operator significantly impacts the effectiveness of Mixup, where hybrid pooling operators outperform both the Max-pooling and GMT operators in terms of producing accurate and robust GNN models. The hyperparameter $\alpha$ has a limited impact on Mixup in augmenting graph-structured data. This study gave the lesson that, when using Mixup in GNNs, carefully choosing the pooling operators could help produce better models.

\section{Acknowledgement}
This research is supported in part by JSPS KAKENHI Grant No. JP23H03372, Japan. 

\bibliographystyle{elsarticle-num}
\bibliography{reference}

\end{document}